\documentclass[10pt,twocolumn,letterpaper]{article}

\usepackage[pagenumbers]{cvpr} %

\input{preamble}

\definecolor{cvprblue}{rgb}{0.21,0.49,0.74}
\usepackage[pagebackref,breaklinks,colorlinks,allcolors=cvprblue]{hyperref}

\def\confName{CVPR}
\def\confYear{2027}

\title{LiveRelight: Sub-Second Global Illumination Editing for Gaussian Interiors}

\author{Noah Rothenberger$^{1}$ \quad Dor Verbin$^{2}$ \quad Serge Belongie$^{1}$ \quad Bernhard Kerbl$^{1}$\\
$^{1}$University of Copenhagen \qquad $^{2}$Google DeepMind\\
}

\begin{document}
\maketitle
\begin{abstract}
Relighting reconstructed 3D scenes remains challenging due to the complex nature of light transport: slow or approximate visual feedback restricts the editor's ability to fine-tune the results to satisfaction.
We present a hybrid relighting approach that enables interactive, high-fidelity updates of direct and indirect indoor illumination without per-edit optimization.
Our approach permits the editor to make a wide range of relighting edits at room-scale, with near-instant feedback. 
Our key insight is to precompute geometric light exchange rather than illumination responses, retaining an inexpensive transport solve that accommodates changes in emission and surface reflectance.
We evaluate direct illumination on reconstructed 2D Gaussians using visibility-aware sampling of one-sided emitters, while propagating diffuse interreflection through sparse radiosity on a triangle proxy.
A sparse projection fitted through Gaussian alpha compositing transfers the proxy solution to the reconstruction.
Our formulation further supports fast fixture transformations and emitter insertion by recomputing light source injection. 
Across eight relighting conditions of our synthetic test suite, our sub-second edits improve relighting fidelity over prior path-traced editing by an average of 5\,dB PSNR, while reducing editing latency by up to three orders of magnitude. 
Qualitative results further demonstrate realistic, interactive relighting of real-world captures.\\
Project page: \url{https://liverelight.github.io}

\vspace*{-\baselineskip}
\end{abstract}    
\section{Introduction}
\label{sec:intro}
Image-based rendering and novel-view synthesis are fundamental tools for immersive exploration of photorealistic 3D environments. 
Many use cases additionally demand post-hoc manipulation on top of the captured scene: a director might seek to improve poor lighting on a film set; a broker may want to show an apartment at golden hour.
Radiance field reconstructions, \eg, 3D Gaussian Splatting~\cite{kerbl2023gaussians}, can render indoor scenes photorealistically, but their appearance is restricted to the illumination observed at capture.

Physically-based inverse rendering methods enable principled relighting of neural fields~\cite{nerv2021,jin2023tensoir} and Gaussian splats~\cite{gao2024r3dg,gu2025irgs,yang2026eagpt}, but evaluating multi-bounce light transport from scratch for each edit remains prohibitively slow.
Precomputing per-light responses makes relighting comparatively inexpensive by linearly recombining stored solutions~\cite{debevec2000acquiring, liang2026luxremix}. However, these solutions encode both the scene's reflectance and the spatial configuration of its sources. %
We propose to instead precompute the \emph{transport operator}, keeping a live solver in the editing loop. 
We exploit the fact that relighting edits incur negligible changes in geometry and visibility, allowing sparse updates: %
a fresh solve updates interreflection while avoiding a vast portion of expensive geometric computation and scene exploration.

Drawing from classical two-pass rendering principles, our hybrid method splits the image by transport character: direct light---sharp and view-anchored---is efficiently sampled via next-event estimation (NEE) on one-sided reconstructed Gaussian emitters, while indirect light---typically smooth and diffuse-dominated---is solved by sparse radiosity updates on a proxy mesh. 
We further propose an optimization-based coupling of the Gaussian--mesh duality and adopt a coherent one-sided emitter scheme to achieve high-quality results for challenging indoor scenarios.

Beyond light source color and intensity, our formulation also extends to geometric manipulation of fixtures: translation, rotation, and scaling act directly on the reconstructed Gaussian assemblies as similarity transformations, without refitting. %
The resulting approach updates direct and indirect illumination after emission, albedo, and fixture edits in under a second, without additional reconstruction or per-edit optimization. 
Thus, our contributions are:
\begin{itemize}
    \item Sub-second global illumination editing of emission and reflectance in captured interiors, coupling Gaussian direct light to precomputed mesh exchange in a live solve. %
    \item A sparse projection fitted to radiance probes, coupling radiosity on a proxy mesh with radiance on 2D Gaussians. %
    \item Editable light placement: fixtures are moved, rescaled, or inserted via Sim(3) transformations with updated visibility and source injection. %
    \item Evaluation on synthetic and real scenes, including two new benchmarks with transport-decomposed ground truth for direct light, higher orders, and the composite.
\end{itemize}
The source code and data will be made publicly available. %

\section{Related Work}
\label{sec:related works}
\paragraph{Inverse rendering.}
\looseness=-1
Recovering editable physical factors from multi-view imagery requires disentangling geometry, material, and illumination~\cite{jin2023tensoir, zhang2021nerfactor}. 
For complex indoor environments, recent methods constrain this ambiguous decomposition through emissive on/off observations~\cite{jeong2024esrnerf} or joint estimation of HDR illumination and camera responses~\cite{lin2025iris}. 
Interreflection is incorporated into material recovery through inverse path tracing~\cite{wu2023fipt,yu2023milo}, neural incident-radiance fields~\cite{zhu2023i2sdf}, or mesh-based lighting and irradiance estimates~\cite{li2023texir}.
While these formulations recover editable scene factors, path-traced relighting still incurs the cost of re-evaluating visibility and multi-bounce interreflection for each new condition.

\paragraph{Gaussian light transport and relighting.}
\looseness=-1
Physically-based extensions of 3D and 2D Gaussian Splatting~\cite{kerbl2023gaussians,huang2024surfsplatting} augment splats with BRDF and visibility  models~\cite{gao2024r3dg,liang2024gsir}, or learn relightable appearance from multi-illumination captures~\cite{bi2024gs3}.
The cost of subsequent edits depends on how indirect illumination updates:
Gaussian ray queries~\cite{gu2025irgs} support interreflection, recursive path tracing evaluates repeated scattering explicitly~\cite{yang2026eagpt,zhu2026ptirgs}, and screen-space and cubemap tracing~\cite{chen2025gigs} or limited-bounce area-emitter transport~\cite{sabae2026aegir} approximate it.
LuxRemix~\cite{liang2026luxremix} decomposes captured indoor illumination into reusable per-source contributions for real-time edits, but does not recompute transport under reflectance or source geometry changes. 
RadiosityGS~\cite{jiang2025radiositygs} solves radiosity directly on semi-opaque Gaussian surfels for joint geometry and material recovery: stochastic transport sampling enables responsive relighting of object-centric scenes, but the authors identify important open challenges in scaling their approach to room-scale environments. %
Hu \etal~\cite{hu2026dynamicgi} achieve global illumination for dynamic Gaussian scenes through stochastic ray tracing and spatiotemporal caching, but their reliance on screen-space reuse restricts higher-order transport, leaving certain multi-bounce interactions through off-screen surfaces unresolved.
We instead exploit the predominantly static geometry of indoor reconstructions: 
direct illumination is evaluated on the Gaussians, while a sparse exchange operator estimated on an opaque proxy is reused for multi-bounce diffuse updates without temporal accumulation or per-edit optimization; fixture transformations enter via updated source injection. %

\paragraph{Two-pass transport and precomputed global illumination.}
\looseness=-1
Classical two-pass rendering combines a view-independent diffuse solution with view-dependent rendering~\cite{wallace1987twopass,sillion1989twopass}.
Radiosity~\cite{goral1984radiosity,cohen1988progressive}, virtual point lights~\cite{keller1997instant}, and neural rendering-equation solvers~\cite{hadadan2021neuralradiosity} provide different representations of indirect transport.
Direct-to-indirect transfer~\cite{hasan2006directindirect} maps newly evaluated direct illumination through precomputed multi-bounce transport, while meshless hierarchical bases extend this principle to free viewpoints~\cite{lehtinen2008meshless}.
Precomputed radiance transfer (PRT)~\cite{sloan2002prt} and its Gaussian variants~\cite{guo2024prtgs,zhang2024prtgaussian} precompute transfer for fixed geometry and reflectance, while precomputed polynomial representations support material edits under fixed lighting and viewpoint~\cite{benartzi2008polynomial}.
Our formulation retains geometric exchange separately from emission and reflectance, allowing both to enter a fresh radiosity solve. %
Coupling surface transport to unstructured, semi-transparent Gaussians presents a domain gap: surface solvers associate radiance with geometric elements, whereas Gaussian image formation blends overlapping contributions with view-dependent compositing weights.
Our sparse projection bridges this gap by fitting the transfer through alpha-composited image formation.

\paragraph{Generative relighting.}
\looseness=-1
To circumvent explicit transport modeling/intrinsic decomposition, recent works use diffusion priors to enable environment-lit object relighting~\cite{jin2024neuralgaffer}, local light controls~\cite{magar2025lightlab}, and illumination harmonization for reconstruction~\cite{alzayer2025generative}.
Distilling generated images into a 3D representation~\cite{zhao2024illuminerf,xing2026gr3en} provides free-viewpoint rendering but requires a new fit for each target illumination.
Lighting-conditioned object representations~\cite{tang2025rogr} and per-light scene decompositions~\cite{liang2026luxremix} amortize this cost, although the latter restrict subsequent edits to their stored source configurations.
SyncLight~\cite{serrano2026synclight} propagates a reference-view lighting edit across views, while LightBridge~\cite{cao2026lightbridge} predicts updated Gaussian appearance without per-condition fitting.
These approaches provide strong appearance priors, but do not support prescribed changes to reconstructed fixture geometry and surface albedo with corresponding updates to cast shadows and color bleeding, which our explicit transport formulation provides at sub-second latency.

\section{Method}
\label{sec:method}
Starting from a Gaussian-based scene representation, our method introduces consistent one-sided emission for reconstructed light sources and a hybrid transport formulation that evaluates direct illumination on Gaussians, while deferring diffuse interreflection to a triangle proxy. A sparse projection, fitted through differentiable image formation, transfers the proxy solution back to the Gaussians.%

\subsection{Scene Representation and Formulation}
We reconstruct a scene with $N_G$ oriented 2D Gaussians~\cite{huang2024surfsplatting} from posed, radiometrically calibrated linear-radiance images captured under a single lighting condition, using EAG-PT~\cite{yang2026eagpt} with the emission model described in \cref{sec:onesided}.
Gaussian $i$ carries center $\boldsymbol\mu_i$, tangent scales $(s_{i1},s_{i2})$, normal $\mathbf n_i$, opacity $o_i$, capture radiance $\mathbf c_i$, albedo $\boldsymbol\rho_i$, and emissive weight $e_i$. 
Selected emitting Gaussians carry RGB source radiance $\boldsymbol\ell_i\in\mathbb R^3$.
In this work, we assume Lambertian reflectance and nonreflecting emitters.

\subsection{Direct Light: NEE on 2D Gaussians}
\label{sec:direct}
Unlike indirect interreflections, which vary smoothly across surfaces, direct illumination usually produces sharp shadow boundaries and exhibits high-frequency sensitivity to emitter extent and occlusions. We therefore evaluate direct diffuse lighting directly on the reconstructed Gaussians themselves.
For each pixel, we alpha-composite depth, diffuse albedo, and camera-facing shading normals along the primary ray and normalize them by the accumulated opacity, terminating accumulation before it exceeds $0.95$. This forms a per-pixel receiver $\mathbf{x}$ with shading normal $\mathbf{n}_x$ and albedo $\boldsymbol{\rho}(\mathbf{x})$ (offset slightly along the normal to avoid self-intersection). From each receiver $\mathbf{x}$, our explicit emitter representation then enables direct shading using next-event estimation (NEE)~\cite{shirley1996direct, veach1997robust}.
Specifically, we draw $M$ samples $\{(i_m, \mathbf{y}_m)\}_{m=1}^M$, each selecting an emitting Gaussian $i_m$ with probability $p_{i_m}$, and a point $\mathbf{y}_m$ on its tangent plane with area density $q_{i_m}(\mathbf{y}_m)$ (we describe our choice of distributions later in this subsection). 
With these samples we define the Monte Carlo estimate:
\begin{equation}
 \widehat{\mathbf L}_d(\mathbf x)=\frac{\boldsymbol\rho(\mathbf x)}{\pi M}\odot
 \sum_{m=1}^{M}\frac{\boldsymbol\ell_{i_m}\alpha_{i_m}(\mathbf y_m)
 V(\mathbf x,\mathbf y_m)g_{i_m}(\mathbf x,\mathbf y_m)}
 {p_{i_m}q_{i_m}(\mathbf y_m)},
 \label{eq:nee}
\end{equation}
where $\boldsymbol{\ell}_i$ is the radiance of emitter $i$, $\alpha_i(\mathbf{y})$ is its spatially varying opacity, and $g_i$ is the geometric factor:
\begin{equation}
 g_i(\mathbf x,\mathbf y)=
 \frac{[\mathbf n_x^\top\boldsymbol\omega]_+
 [\mathbf n_i^\top(-\boldsymbol\omega)]_+}{\|\mathbf y-\mathbf x\|^2},
 \quad \boldsymbol\omega=\frac{\mathbf y-\mathbf x}{\|\mathbf y-\mathbf x\|},
 \label{eq:nee-geometry}
\end{equation}
with clamping denoted by $[\cdot]_+=\max(\cdot, 0)$, and emitter normal $\mathbf{n}_i$. %
The shadow transmittance $V(\mathbf{x}, \mathbf{y})$ accumulates $\prod_i (1-\alpha_i)$ across all Gaussians intersected along the segment between $\mathbf{x}$ and $\mathbf{y}$. Only the sampled Gaussian $i_m$ is excluded, since $\alpha_{i_m}(\mathbf y_m)$ already weights its emission.

To reduce variance in~\cref{eq:nee} we match $p_i$ and $q_i$ to each emitter's radiance and Gaussian falloff. For an emitter $i$ with tangent scales $(s_{i1}, s_{i2})$ and peak opacity $o_i$, its opacity profile is $\alpha_i(\mathbf{y}) = o_i\cdot \gamma_i(\mathbf{y})$, where $\gamma_i(\mathbf{y}) \in (0, 1]$ is the unnormalized 2D Gaussian kernel, and its integral over the plane is $A_i = 2\pi o_i s_{i1} s_{i2}$. We therefore select emitter $i$ proportionally to $A_i$ and its radiance strength, and we sample $\mathbf{y}$ directly from the 2D Gaussian distribution:
\begin{equation}
 p_i\propto\bar\ell_i A_i,\quad
 q_i(\mathbf y)=\frac{\gamma_i(\mathbf y)}{2\pi s_{i1}s_{i2}},
 \label{eq:source-sampling}
\end{equation}
where $\bar\ell_i$ is the channel mean of the Gaussian's radiance $\boldsymbol \ell_i$.
Under this choice, $\alpha_i(\mathbf{y})/q_i(\mathbf{y}) = A_i$, analytically canceling the Gaussian falloff.
We use $M=8$ samples per pixel and denoise direct reflection with the OptiX HDR denoiser, guided by reconstructed albedo and normals. Approximations from finite Gaussian support, unrenormalized sampling weights, and ray offsets are detailed in \cref{sec:supp-bounded-nee}.

\subsection{Indirect Light: Radiosity and Projection}
\label{sec:indirect}
\paragraph{Radiosity.} Diffuse interreflection typically varies more smoothly than direct illumination, making a coarse proxy, like a triangle mesh, a useful approximation. 
On this proxy, radiosity expresses a balance of diffuse power while separating geometric exchange from emission and reflectance. 
Precomputing that exchange into a \emph{form factor matrix} $F$ turns each edit into sparse transport propagation, with a residual for assessing reflection-order truncation (\cref{sec:supp-indirect-illumination}). %
EAG-PT~\cite{yang2026eagpt} demonstrates the benefit of a unified representation that keeps all transport on the reconstructed Gaussians, and we preserve this property for emission and first reflection, where the geometric inaccuracies of a proxy would be directly visible (Tab.~\ref{tab:transport_components}).
Confining the proxy to the remaining orders makes the relit results robust to its discretization (\cref{sec:exp-ablations}).
We extract a proxy mesh once per scene by TSDF fusion~\cite{curless1996volumetric} of rendered expected depth followed by attribute-aware simplification that preserves material boundaries and limits triangle extent.
Its $N_T$ triangles receive reflectances $\boldsymbol\rho^M$ from depth-consistent projections of rendered albedo at their centroids. Pairwise geometric exchange between primitives is computed by hemicube integration~\cite{cohen1985hemicube} at triangle centroids (resolution $64$), resulting in a sparse $F$.
Entry $F_{jk}$ integrates the visible solid angle of triangle $k$ at centroid $j$, weighted by the receiving cosine and divided by $\pi$. 
The first intersection occludes, and back-facing hits transfer no radiance.
With $T_\rho Z=\boldsymbol\rho^M\odot(FZ)$ for $Z\in\mathbb R^{N_T\times3}$ and emitted triangle radiance $E$, outgoing diffuse radiance satisfies:
\begin{equation}
 B=E+T_\rho B,\quad
 B_{\rm ind}=\sum_{b=2}^{K}T_\rho^bE.
 \label{eq:radiosity}
\end{equation} %
With $K=8$, we retain orders two through eight, as the Gaussian leg supplies emission and first reflection.

\paragraph{Projection.} Solving indirect illumination on a proxy decouples its cost from the scene's geometric detail, but the resulting surface radiance must still be transferred to the overlapping, semi-transparent Gaussians of the high-fidelity reconstruction.
This transfer must account for alpha compositing in Gaussian-based image formation: each Gaussian affects many pixels, and each pixel blends several Gaussians with view-dependent weights.

We fit this projection through the differentiable Gaussian renderer, using image contributions rather than geometric correspondence alone.
A scene-specific operator $P\in\mathbb R^{N_G\times N_T}$ maps triangle radiance $z$ to Gaussian radiance $Pz$, with weights shared across views and RGB channels. 
Geometric compatibility and training-view associations define the candidates and prior $P^0$. 
Gaussians without a compatible candidate have zero rows and are omitted from the indirect render. %
For pixel $r$ in training view $v$, let $\mathcal H_{vr}$ be its ordered Gaussian hits. 
Hit $h$ has Gaussian index $i(h)$, position $\mathbf y_h$, and effective blending coefficient $\alpha_h$ as evaluated by the renderer, including any filtering or clamping.
We cache its compositing weight $w_h=\alpha_h\prod_{h'<h}(1-\alpha_{h'})$ directly from the renderer and use the same cached weights in both operators below. The product includes all preceding hits, including those without valid triangle associations.
We associate $\mathbf y_h$ with its nearest triangle $t(h)$, subject to the distance, normal, and connected-component criteria defined in \cref{sec:supp-projection}, and denote the valid hits by $\mathcal H_{vr}^{\rm valid}$.
For Gaussian radiances $\mathbf f$ and triangle radiances $z$, we define:
\begin{equation}
  [W_v\mathbf f]_r=\smashoperator[r]{\sum_{h\in\mathcal H_{vr}}}w_h\mathbf f_{i(h)},\quad
  [S_vz]_r=\smashoperator[r]{\sum_{h\in\mathcal H_{vr}^{\rm valid}}}w_hz_{t(h)}.
 \label{eq:image-operators}
\end{equation}
At fixed geometry and opacity, $W_v$ is linear in radiance, although its weights vary with view. 
Evaluating the target $S_vz$ with those same weights makes the fit account for overlapping footprints and partial coverage, without requiring the proxy to reproduce the reconstruction's image formation.

To constrain transfer beyond the captured illumination without additional multi-illumination observations, we supervise $P$ across training views using analytic spatial probes and full radiosity fields under random emission:
\begin{equation}
 \begin{aligned}
 \min_P\;&\mathbb E_{v,z}d_{m_v}(W_vPz,S_vz)\\
 &+\lambda_{\rm cap}\mathbb E_v d_1(W_vPB_{\rm cap},W_v\mathbf c)\\
 &+\lambda_{\rm cyc}\mathcal L_{\rm cyc}
 +\lambda_{\rm prior}\mathcal L_{\rm prior}.
 \end{aligned}
 \label{eq:projection}
\end{equation}
Here $B_{\rm cap}$ includes capture emission and eight reflections.
The Charbonnier loss $d_{m_v}$ compares linear and log-compressed radiance, weighted per pixel by the fraction $m_v$ of compositing mass with valid associations. 
A fixed reverse operator $Q$ averages Gaussian radiance onto triangles using the same hit associations. 
$\mathcal L_{\rm cyc}$ measures relative squared error in $QPz-z$, weighted by triangle observation reliability. 
The prior loss $\mathcal L_{\rm prior}$ averages $D_{\rm KL}(P_i\|P_i^0)$, emphasizing poorly observed Gaussians. 
Supported rows of $P$ are softmax-normalized, which preserves constant radiance.
Fitted once, $P$ requires no per-edit optimization.

\subsection{Image Composition}
\label{sec:composition}
Each emitting Gaussian $i$ carries its fixture's RGB radiance $\boldsymbol\ell_i$, identical across the fixture. 
Proxy emission is initialized from the same value and calibrated as described in \cref{sec:emission-albedo}.
The emissive weight $e_i\in[0,1]$ only sets how much of a Gaussian's camera-visible color is emitted rather than reflected light; it does not change the light a fixture casts, but lets visible emission fade smoothly at fixture boundaries.
Let $\mathcal H^e_{vr}\subseteq\mathcal H_{vr}$ contain front-facing emitter hits. We compute camera-visible emission and emissive coverage as:
\begin{equation}
 [S_v^e]_r = \smashoperator[r]{\sum_{h\in\mathcal H^e_{vr}}}
 w_h e_{i(h)}\boldsymbol\ell_{i(h)}, \quad
 [w_v^e]_r = \smashoperator[r]{\sum_{h\in\mathcal H^e_{vr}}}
 w_h e_{i(h)}. 
 \label{eq:visible-emission}
\end{equation}
The final image combines emission and reflection:
\begin{equation}
 I_v=S_v^e+(1-w_v^e)\odot\left[
 \mathcal D(\widehat{\mathbf L}_{d,v})+W_v^{\rm ind}PB_{\rm ind}\right],
 \label{eq:hybrid-image}
\end{equation}
where $\mathcal D$ denotes direct-light denoising and $W_v^{\rm ind}$ renders the projected field with unsupported Gaussians removed.

\subsection{One-Sided Reconstructed Emitters}
\label{sec:onesided}
\begin{figure}[t]
  \centering
  \paperfigcaptionstyle
  \edef\osw{\the\dimexpr(\linewidth-\paperfiglabelwidth-4\paperfiggap)/3\relax}%
  \edef\osh{\the\dimexpr\osw*2/3\relax}%
  \def\oslab#1{\paperfigrowlabel{\osh}{#1}}%
  \def\oshdr#1{\paperfigheader{\osw}{#1}}%
  \paperfigsetup
  \begin{tabular}{c@{\hspace{\paperfiggap}}c@{\hspace{\paperfiggap}}c@{\hspace{\paperfiggap}}c}
    & \oshdr{EAG-PT (2-sided)} & \oshdr{Ours (1-sided)} & \oshdr{GT} \\[\paperfighdrgap]
    \oslab{Capture} &
    \includegraphics[width=\osw]{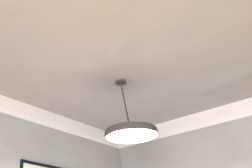} &
    \includegraphics[width=\osw]{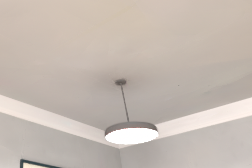} &
    \includegraphics[width=\osw]{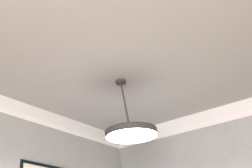} \\[\paperfiggap]
    \oslab{Albedo} &
    \includegraphics[width=\osw]{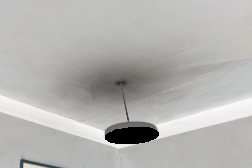} &
    \includegraphics[width=\osw]{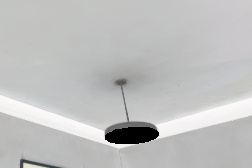} &
    \includegraphics[width=\osw]{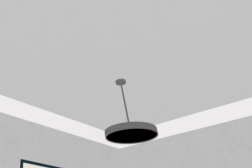} \\[\paperfiggap]
    \oslab{Relit} &
    \includegraphics[width=\osw]{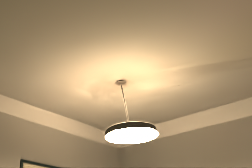} &
    \includegraphics[width=\osw]{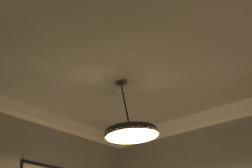} &
    \includegraphics[width=\osw]{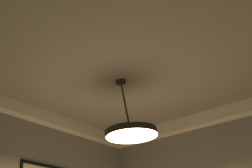} \\
  \end{tabular}
  \caption{\textbf{One-sided emitters.} They avoid albedo overcorrection of EAG-PT~\cite{yang2026eagpt}, enabling accurate relighting of free-hanging fixtures. 
  Two-sided emitters (left) illuminate the ceiling through the pendant's non-emitting back side. Material recovery compensates with darker albedo, and the error persists after relighting. %
  Enforcing oriented one-sided emission (middle) corrects both.
  }
  \label{fig:onesided}
\end{figure}

The clamped emitter cosine in \cref{eq:nee-geometry} lets each Gaussian emit only into the hemisphere its normal faces, whereas EAG-PT~\cite{yang2026eagpt} treats emitters as two-sided.
Real-world indoor light fixtures are rarely omnidirectional: they provide a clearly defined emissive surface. 
Within the EAG-PT framework, reconstructed Gaussians do not identify a physical emitting side, so light transport accumulates radiance whenever a Gaussian light source is encountered, from either side.
As a consequence, material recovery tends to compensate for this inaccuracy by darkening nearby albedos, preserving capture appearance while corrupting the recovered reflectance (Fig.~\ref{fig:onesided}).  
We propose resolving this ambiguity before recovering albedo.
Assuming that most training-camera centers $\mathbf x_v$ lie on the emitting side, we orient each Gaussian emitter by an unweighted majority vote of $\operatorname{sign}[\mathbf n_i^\top(\mathbf x_v-\boldsymbol\mu_i)]$ with the vote evaluated independently for each Gaussian.
When the vote opposes the current normal, we reverse it together with the second tangent, preserving the Gaussian footprint and recovered appearance. 
For an outgoing direction $\boldsymbol\omega$, the oriented emitter has radiance $\boldsymbol\ell_i\mathbf 1[\mathbf n_i^\top\boldsymbol\omega>0]$.
This directional support is shared by material recovery and forward rendering, including the emitter cosine of \cref{eq:nee-geometry}: back faces retain opacity but emit no light.

\section{Interactive Scene Relighting}
\label{sec:editing}
Beyond emission and albedo edits, we propose efficient update procedures for geometric manipulation of light fixtures at sub-second latency. 

\subsection{Editing Emission and Albedo}
\label{sec:emission-albedo}
Emission edits act on fixtures, the groups of Gaussians that form a light source. Identifying fixtures is not specific to our method and could be done in 2D or 3D, manually or automatically;
we mark each fixture's emitting surface with a mask in the capture views (\cref{sec:supp-editing-inputs}).
Once per fixture, we transfer this selection to both representations: Gaussians whose rendered contribution falls mostly inside the mask become the fixture's emitters, and depth-consistent first-hit proxy triangles receive the radiance of their nearest selected Gaussian, with the fixture's emitting-face area capping their total emission (\cref{sec:supp-area-calibration}).
An emission edit then only assigns new fixture radiances $\boldsymbol\ell_i$, and an albedo edit only changes Gaussian albedos and proxy reflectances $\boldsymbol\rho^M$.
Both refresh the direct estimate and re-solve the indirect transport with the same $F$ and $P$, combined via Eq.~\eqref{eq:hybrid-image}.

\subsection{Editing Fixture Geometry}
\label{sec:geometry}
Our precompute-once, relight-instantly doctrine extends to interactive edits of light fixture geometry---\eg, rescaling a ceiling light or re-hanging a pendant---while the proxy mesh, $F$, and $P$ stay frozen. 
\emph{Emitter re-injection} updates indirect illumination by recomputing how the transformed fixture injects light into the room.%

\paragraph{Geometric transformation of Gaussians.}
\looseness=-1
For a selected fixture assembly $\mathcal G_f$, including its emitting surface and associated non-emitting Gaussians, geometric edits about pivot $\mathbf p_f$ result in a similarity $(a,R,\mathbf t)\in\operatorname{Sim}(3)$:
\begin{equation}
\begin{aligned}
\boldsymbol\mu_i'&=\mathbf p_f+aR(\boldsymbol\mu_i-\mathbf p_f)+\mathbf t,\\
(s_{i1}',s_{i2}')&=a(s_{i1},s_{i2}),\quad U_i'=RU_i,
\end{aligned}
\label{eq:fixture-sim3}
\end{equation}
where $a>0$, $R\in\mathrm{SO}(3)$, and $U_i$ contains the tangent axes and oriented normal.
Similarities preserve orthogonal tangent frames and Gaussian falloff, enabling exact parameter updates without refitting.
Opacity, albedo, and emissive weights remain fixed, while the emitting-side orientation rotates with the fixture.
After each edit, the NEE renderer samples the transformed emitters and traces shadow rays through the updated Gaussian scene.

\paragraph{Emitter re-injection.}
\looseness=-1
\label{sec:injection}
To account for changes in fixture placement and extent, we recompute the edited fixture's first reflected radiance on the receivers.
We treat nonreflecting emitting surfaces as sources and retain the room's receiver-to-receiver transport. Within this approximation, the updated first-reflection field determines all retained higher orders, which we regenerate without recomputing $F$.
When an edit is initiated, we transform the fixture's emitting proxy triangles $\mathcal E_f$ and its surrounding object region in a temporary tracing scene that replaces the original object with its transformed copy: the vacated location no longer occludes, the moved body shadows its own emission, and receiver areas and the transport discretization remain unchanged (\cref{sec:supp-fixture-scene}).
We then shoot approximately $10^6$ light rays with origins uniform on each emitting triangle and cosine-distributed directions over its front hemisphere, allocating rays in proportion to triangle area with at least one per triangle.
Selected fixture-body hits and receiver back faces absorb.
Let $\mathcal A_j$ be the original area of receiver $j$, $\mathcal A_k'$ the edited area of source triangle $k$, and $E_k$ its unchanged calibrated radiance, held fixed during the edit.
With $n_k$ rays from triangle $k$, the first reflected radiance is estimated as:
\begin{equation}
\widehat d_{f,j}'=
\frac{\boldsymbol\rho_j^M}{\mathcal A_j}\odot
\sum_{k\in\mathcal E_f}\frac{\mathcal A_k'E_k}{n_k}
\sum_{r=1}^{n_k}\chi_{jkr},
\label{eq:fixture-injection}
\end{equation}
where $\chi_{jkr}=1$ only if the ray first reaches receiver $j$ on its front face, and is zero otherwise.
We replace only the edited fixture's injection while retaining all other emitters: 
with $E_{\neg f}$ the proxy emission field excluding fixture $f$, the retained first-reflection radiance is $d_{\neg f}=T_\rho E_{\neg f}$, giving:
\begin{equation}
B_{\rm ind}'=\sum_{b=1}^{K-1}T_\rho^{b}
\bigl(d_{\neg f}+\widehat d_f'\bigr).
\label{eq:fixture-propagation}
\end{equation}
The same $P$ projects $B_{\rm ind}'$, and Eq.~\eqref{eq:hybrid-image} composes the image.

\begin{figure*}[t]
  \centering
  \paperfigcaptionstyle
  \edef\relw{\the\dimexpr(\linewidth-5\paperfiggap-\paperfiggroupgap-0.5pt)/7\relax}%
  \def\relhdr#1{\paperfigheader{\relw}{#1}}%
  \paperfigsetup
  \begin{subfigure}[t]{\dimexpr\relw*4+3\paperfiggap\relax}
    \centering
    \begin{tabular}{c@{\hspace{\paperfiggap}}c@{\hspace{\paperfiggap}}c@{\hspace{\paperfiggap}}c}
      \relhdr{Capture} & \relhdr{Relit 1} & \relhdr{Relit 2} & \relhdr{Relit 3} \\[\paperfighdrgap]
    \includegraphics[width=\relw]{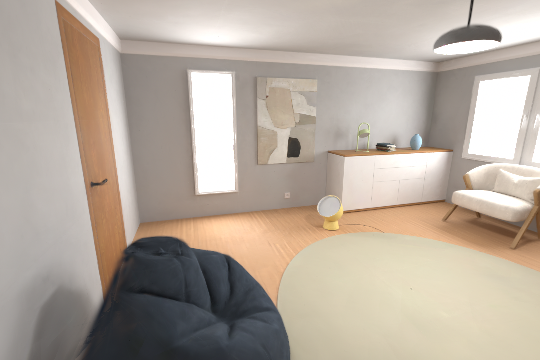} &
    \includegraphics[width=\relw]{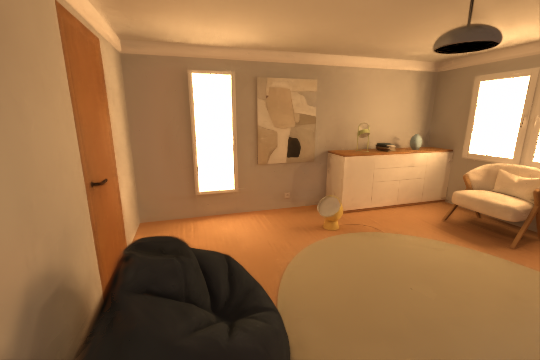} &
    \includegraphics[width=\relw]{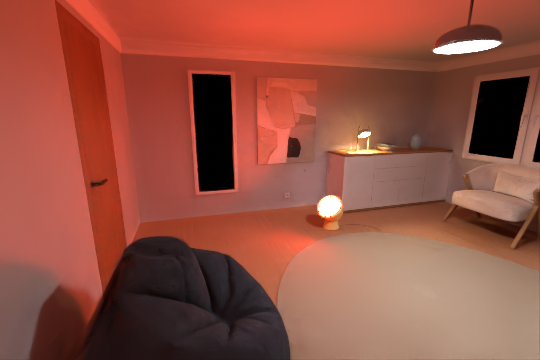} &
    \includegraphics[width=\relw]{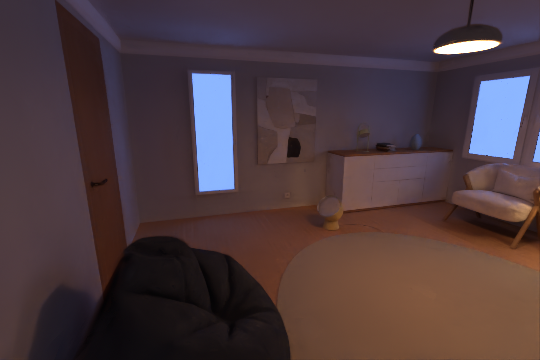} \\
      [\paperfiggap]
    \includegraphics[width=\relw]{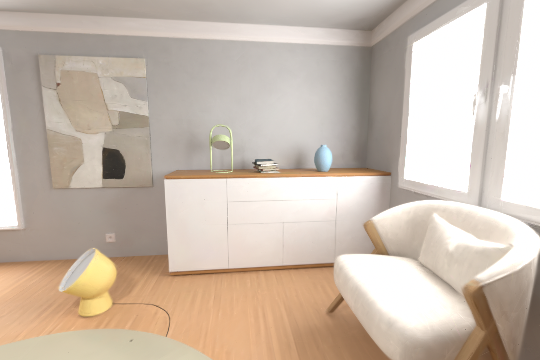} &
    \includegraphics[width=\relw]{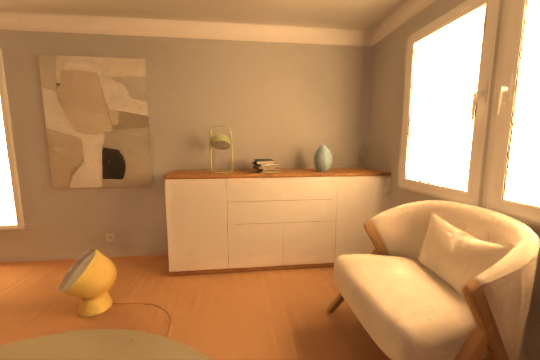} &
    \includegraphics[width=\relw]{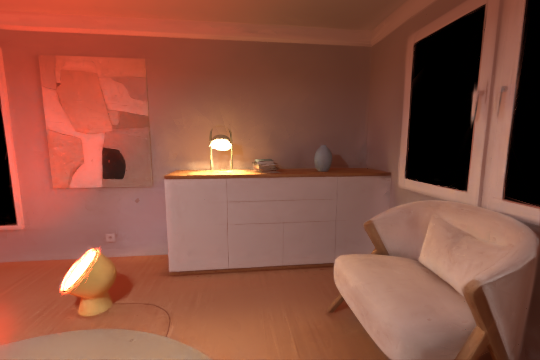} &
    \includegraphics[width=\relw]{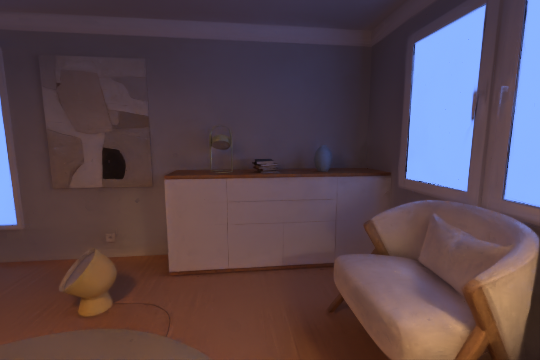} \\
      [\paperfiggap]
      \includegraphics[width=\relw]{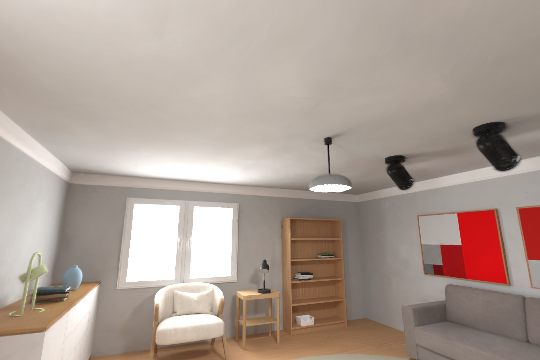} &
      \includegraphics[width=\relw]{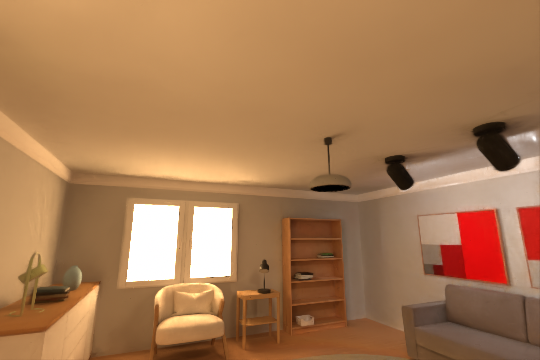} &
      \includegraphics[width=\relw]{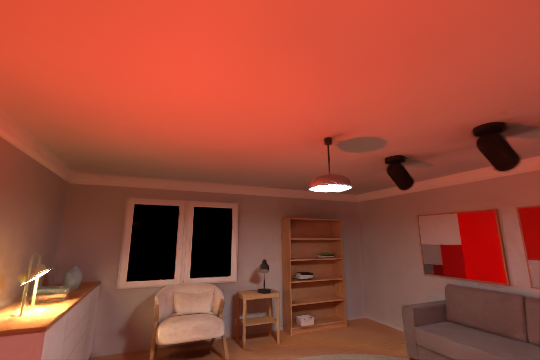} &
      \includegraphics[width=\relw]{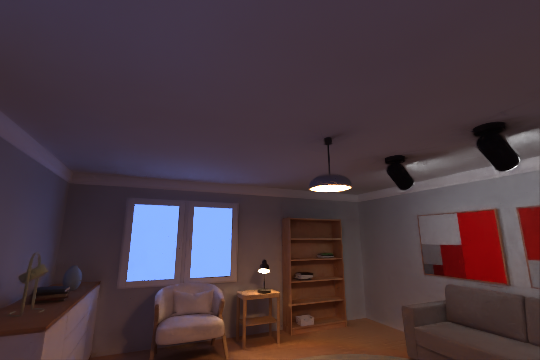} \\
    \end{tabular}
    \caption{\emph{Living room}}
    \label{fig:relighting-livingroom}
  \end{subfigure}%
  \hspace{\paperfiggroupgap}%
  \begin{subfigure}[t]{\dimexpr\relw*3+2\paperfiggap\relax}
    \centering
    \begin{tabular}{c@{\hspace{\paperfiggap}}c@{\hspace{\paperfiggap}}c}
      \relhdr{Capture} & \relhdr{Relit} & \relhdr{GT} \\[\paperfighdrgap]
      \includegraphics[width=\relw]{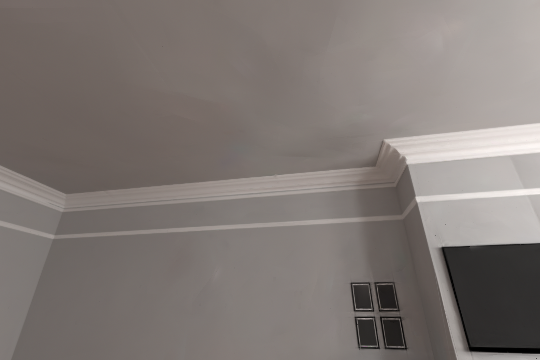} &
      \includegraphics[width=\relw]{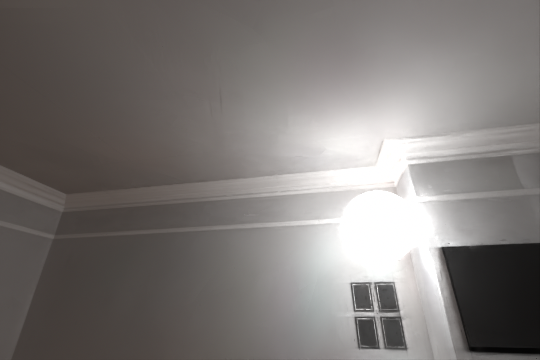} &
      \includegraphics[width=\relw]{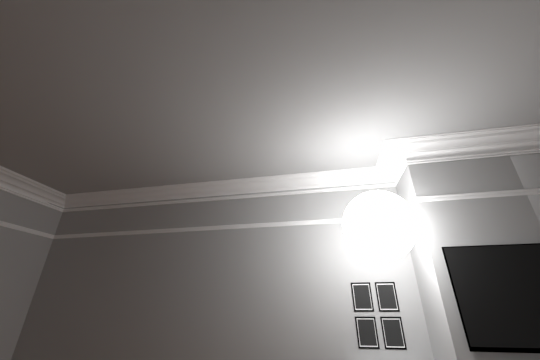} \\[\paperfiggap]
      \includegraphics[width=\relw]{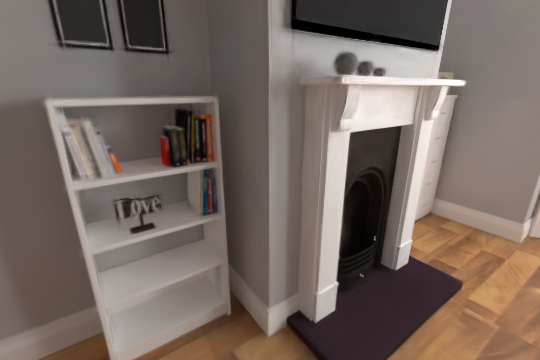} &
      \includegraphics[width=\relw]{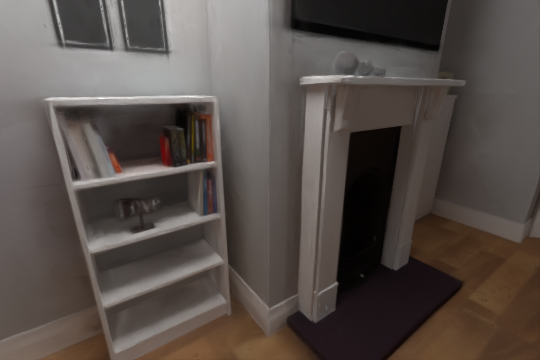} &
      \includegraphics[width=\relw]{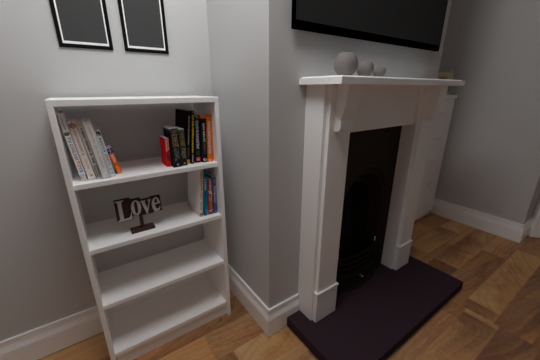} \\[\paperfiggap]
      \includegraphics[width=\relw]{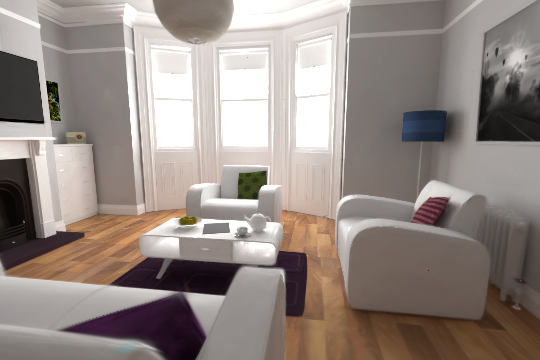} &
      \includegraphics[width=\relw]{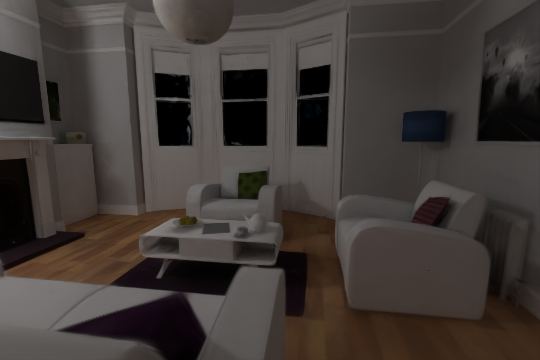} &
      \includegraphics[width=\relw]{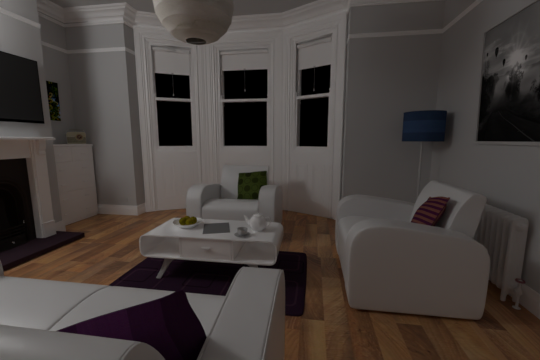} \\
    \end{tabular}
    \caption{\emph{Bitterli-Livingroom}~\cite{resources16}}
    \label{fig:relighting-blivingroom}
  \end{subfigure}
  \caption{
  \textbf{Relighting with our method.} One test view per row: the reconstruction under capture lighting (Capture) and new conditions (Relit 1--3), for which we re-solve light transport including (off-screen) interreflection in the scene.
  (a)~\emph{Living room}: three conditions change emission strength and color, switch lights off, and turn on fixtures that never emit at capture (\eg, the red floor lamp in Relit~2).
  (b)~\emph{Bitterli-Livingroom}~\cite{resources16}: the original lights are switched off and an emissive sphere is inserted (top row).
  }
  \label{fig:relighting}
\end{figure*}

\section{Evaluation}
\label{sec:evaluation}
\paragraph{Scenes and protocol.}
\looseness=-1
We evaluate relighting fidelity and speed on synthetic and real input scenes.
For the synthetic scenes, we report error metrics against path-traced ground truth for each target condition (Blender Cycles, 1024\,spp, denoised).
In addition to the established benchmarks \emph{Bitterli-Kitchen} and \emph{Bitterli-Livingroom}---two rendering-resource scenes used in EAG-PT~\cite{resources16, yang2026eagpt}---we introduce two synthetic indoor scenes: \emph{Kitchen} and \emph{Living room} contain challenging fixtures, such as free-hanging pendants, spotlights, and cabinet-mounted emitters, and provide 450 posed views each with radiance, albedo, depth, and emission reference passes. 
Three target lighting conditions per scene vary the active sources and their intensities and colors (\cref{fig:relighting}).
Each relit condition is specified through the per-fixture editing inputs of \cref{sec:emission-albedo}, with ground truth renders serving only as evaluation references on held-out test views. 
Synthetic scenes use 200k 2D Gaussians and $\sim$200k proxy triangles, real captures 500k and $\sim$400k. %
We report mean test-view PSNR, SSIM, and FLIP~\cite{andersson2020flip}.
PSNR and SSIM use sRGB images clipped to $[0,1]$, and FLIP follows EAG-PT's evaluation code in HDR mode on linear radiance.
We additionally evaluate two real-world indoor captures qualitatively only, \emph{FIPT-Classroom} and \emph{FIPT-Conference}~\cite{wu2023fipt}, since reference images of our edited conditions are unavailable.
All experiments run on a single RTX~4090 GPU.

\paragraph{Baseline.}
\looseness=-1
Our main baseline for relighting reconstructed room-scale scenes is EAG-PT~\cite{yang2026eagpt}. While thematically relevant, neither RadiosityGS~\cite{jiang2025radiositygs} (object-scale scenes under point and environment lighting) nor Hu \etal~\cite{hu2026dynamicgi} (forward rendering with given materials and lights) target relighting room-scale reconstructions with emissive fixtures from a single captured lighting condition. %
LuxRemix~\cite{liang2026luxremix} and AEGIR~\cite{sabae2026aegir} address emission edits but have not released their relighting code.
We follow EAG-PT's published conventions to evaluate its quality and performance, path tracing each edited condition at 1024\,spp with bounce limit~7.
Its optional light-baking stage distills path-traced renders into splat radiance for fast subsequent viewing; however, doing so does not reduce the time required for an edit to first become visible to the user.

\subsection{Relighting Synthetic and Real-World Scenes}
\label{sec:exp-relighting}
\paragraph{Synthetic environments.}
\looseness=-1
Tab.~\ref{tab:relighting} evaluates three conditions each for \emph{Kitchen} and \emph{Living room}, including turning on lights that never emit at capture, large power and color changes of active sources, and dim evening conditions.
Our composites exceed the path-traced baseline in PSNR (+1.4 to +10.7\,dB) and SSIM in all conditions, with lower FLIP in all but one.
The largest margin is the \emph{Living room} turn-on (Relit~2): EAG-PT assigns uniform radiance to selected splats whose footprints exceed the small fixtures' emitting areas, whereas our contribution-gated emitter selection and area-calibrated proxy emission stay close to the reference.
The baseline is strongest where captured sources dominate (\emph{Living room} Relit~1, within 1.5\,dB).
For the two Bitterli scenes, we recreate EAG-PT's relighting experiment by removing all reconstructed emitters and inserting an emissive sphere:
its separately reconstructed Gaussians are merged into the scene and an analytic sphere is appended to the proxy mesh.
Against reproduced EAG-PT, our method improves PSNR by 1.9 and 3.3\,dB, respectively.%

\begin{table}[t]
  \centering
  \small
  \renewcommand{\arraystretch}{0.95}%
  \edef\padTsep{\the\dimexpr\belowrulesep/2\relax}%
  \edef\padBsep{\the\dimexpr\aboverulesep/2\relax}%
  \setlength{\aboverulesep}{0pt}\setlength{\belowrulesep}{0pt}%
  \def\padT{\rule{0pt}{\dimexpr\arraystretch\ht\strutbox+\padTsep\relax}}%
  \def\padB{\rule[-\dimexpr\arraystretch\dp\strutbox+\padBsep\relax]{0pt}{0pt}}%
  \setlength{\tabcolsep}{2pt}%
  \begin{tabular*}{\columnwidth}{@{\extracolsep{\fill}}llcccc@{}}
    \toprule
    \padT\padB & Method & PSNR$\uparrow$ & SSIM$\uparrow$ & FLIP$\downarrow$ & Time (s)$\downarrow$ \\ %
    \midrule
    \multicolumn{2}{@{}l}{\padT\emph{Kitchen} (capture)}
      & 39.47 & 0.988 & 0.058 & -- \padB \\ %
    \padT Relit~1 & EAG-PT & 28.08 & 0.955 & 0.304 & 147.2 \\ %
      & Ours & \textbf{35.95} & \textbf{0.983} & \textbf{0.139} & \textbf{0.244} \\ %
    Relit~2 & EAG-PT & 33.83 & 0.978 & 0.130 & 142.7 \\ %
      & Ours & \textbf{36.99} & \textbf{0.983} & \textbf{0.125} & \textbf{0.307} \\ %
    Relit~3 & EAG-PT & 32.14 & 0.975 & 0.173 & 134.5 \\ %
      & Ours & \textbf{36.96} & \textbf{0.984} & \textbf{0.117} & \textbf{0.287} \padB \\ %
    \midrule
    \multicolumn{2}{@{}l}{\padT\emph{Living room} (capture)}
      & 37.55 & 0.983 & 0.059 & -- \padB \\ %
    \padT Relit~1 & EAG-PT & 30.49 & 0.969 & \textbf{0.160} & 167.9 \\ %
      & Ours & \textbf{31.90} & \textbf{0.972} & 0.165 & \textbf{0.250} \\ %
    Relit~2 & EAG-PT & 22.48 & 0.931 & 0.377 & 184.4 \\ %
      & Ours & \textbf{33.22} & \textbf{0.972} & \textbf{0.163} & \textbf{0.211} \\ %
    Relit~3 & EAG-PT & 28.78 & 0.962 & 0.284 & 166.5 \\ %
      & Ours & \textbf{35.57} & \textbf{0.981} & \textbf{0.149} & \textbf{0.248} \padB \\ %
    \midrule
    \multicolumn{2}{@{}l}{\padT\emph{Bitterli-Kitchen} (capture)}
      & 37.39 & 0.974 & 0.073 & -- \padB \\ %
    \padT Relit & EAG-PT & 28.89 & 0.955 & 0.180 & 183.5 \\ %
      & Ours & \textbf{30.78} & \textbf{0.956} & \textbf{0.156} & \textbf{0.153} \padB \\ %
    \midrule
    \multicolumn{2}{@{}l}{\padT\emph{Bitterli-Livingroom} (capture)}
      & 37.56 & 0.973 & 0.066 & -- \padB \\ %
    \padT Relit & EAG-PT & 29.23 & 0.943 & 0.210 & 160.9 \\ %
      & Ours & \textbf{32.51} & \textbf{0.957} & \textbf{0.158} & \textbf{0.149} \padB \\ %
    \bottomrule
  \end{tabular*}
  \caption{
  \textbf{Relighting accuracy against ground truth and time on synthetic scenes.}
  \emph{Living room} Relit~1--3 and \emph{Bitterli-Livingroom} edits are shown in Fig.~\ref{fig:relighting}; \emph{Kitchen} in \cref{sec:relighting-additional-results}.
  }
  \label{tab:relighting}
\end{table}

\begin{figure}[t]
  \centering
  \paperfigcaptionstyle
  \edef\editw{\the\dimexpr0.325\linewidth\relax}%
  \paperfigsetup
  \begin{subfigure}[t]{\editw}
    \captionsetup{position=bottom,skip=2pt}
    
    \label{fig:disk_ladder-a}
    \includegraphics[width=\linewidth]{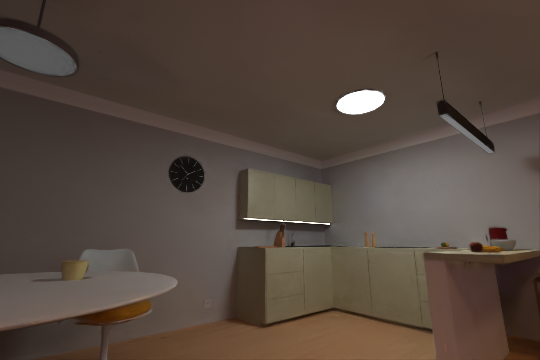}
    \caption{Unedited}
  \end{subfigure}%
  \hspace{\paperfiggap}%
  \begin{subfigure}[t]{\editw}
    \captionsetup{position=bottom,skip=2pt}
    
    \label{fig:disk_ladder-b}
    \includegraphics[width=\linewidth]{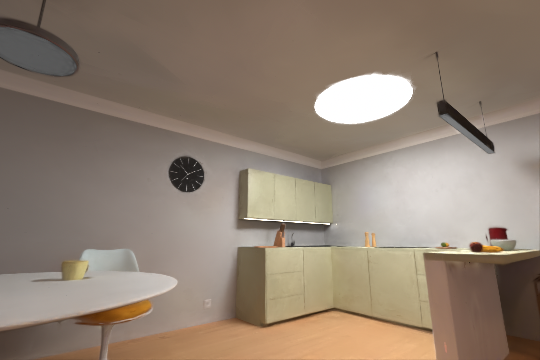}
    \caption{$2\times$}
  \end{subfigure}%
  \hspace{\paperfiggap}%
  \begin{subfigure}[t]{\editw}
    \captionsetup{position=bottom,skip=2pt}
    
    \label{fig:disk_ladder-c}
    \includegraphics[width=\linewidth]{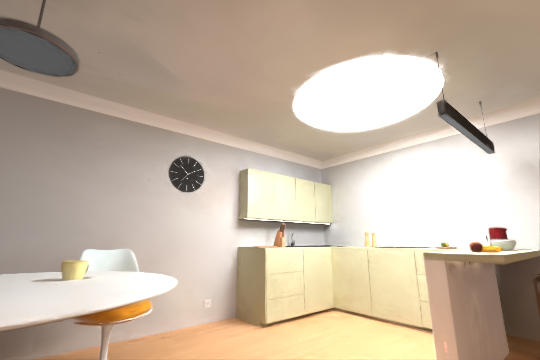}
    \caption{$3\times$}
  \end{subfigure}
  \par\vspace{\paperfigrowgroupgap}\nointerlineskip
  \begin{subfigure}[t]{\editw}
    \captionsetup{position=bottom,skip=2pt}
    
    \label{fig:pendant_moves-a}
    \includegraphics[width=\linewidth]{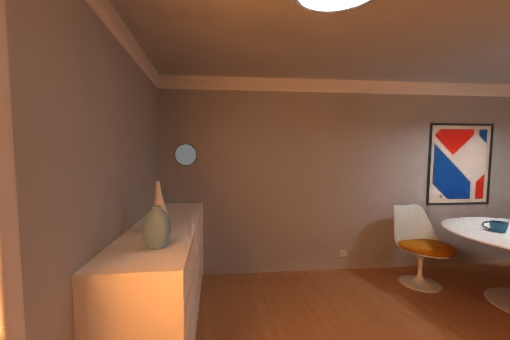}
    \caption{Unedited}
  \end{subfigure}%
  \hspace{\paperfiggap}%
  \begin{subfigure}[t]{\editw}
    \captionsetup{position=bottom,skip=2pt}
    
    \label{fig:pendant_moves-b}
    \includegraphics[width=\linewidth]{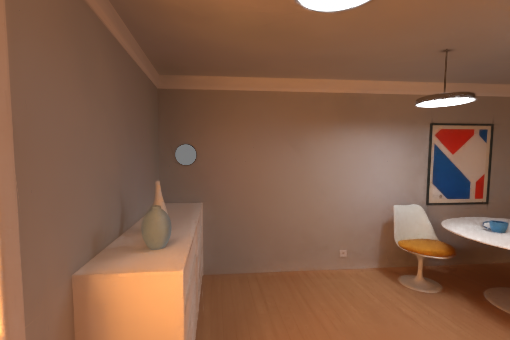}
    \caption{Moved 1\,m}
  \end{subfigure}%
  \hspace{\paperfiggap}%
  \begin{subfigure}[t]{\editw}
    \captionsetup{position=bottom,skip=2pt}
    
    \label{fig:pendant_moves-c}
    \includegraphics[width=\linewidth]{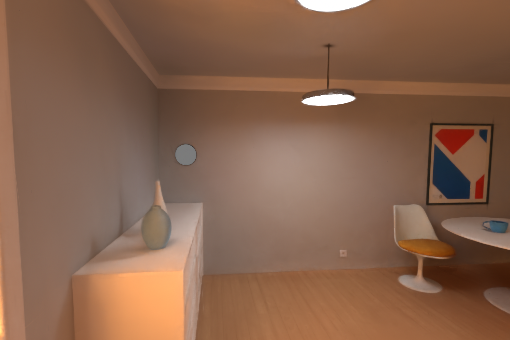}
    \caption{Moved 2\,m}
  \end{subfigure}
  \caption{
  \textbf{Fixture geometry editing in \emph{Kitchen} at fixed source radiance.}
  (a--c)~Scaling the ceiling disk about its center by $a$, which multiplies emitted power by $a^2$.
  (d--f)~Moving the (off-screen) dining pendant under Relit~3 toward the room center.
  }
  \label{fig:fixture_edits}\label{fig:fixture_scaling}
\end{figure}

We also evaluate geometric fixture edits in \emph{Kitchen}, at fixed source radiance against ground truth of each edited scene.
PSNR remains high:
ceiling-disk scalings of up to $2\times$, alone or combined with 10--20\,cm translations, stay above 33.3\,dB under Relit~2, and pendant translations of up to 2\,m stay above 36.3\,dB under Relit~3 (\cref{fig:fixture_edits,tab:fixture_edits}).
Our method provides visual feedback for these edits within about half a second;
EAG-PT must path-trace each edit and degrades with translation distance (\cref{sec:fixture-eagpt}).

\begin{figure}[t]
  \centering
  \paperfigcaptionstyle
  \edef\editw{\the\dimexpr0.325\linewidth\relax}%
  \paperfigsetup

  \begin{subfigure}[t]{\editw}
    \captionsetup{position=bottom,skip=2pt}
    
    \label{fig:real_edits-a}
    \includegraphics[width=\linewidth]{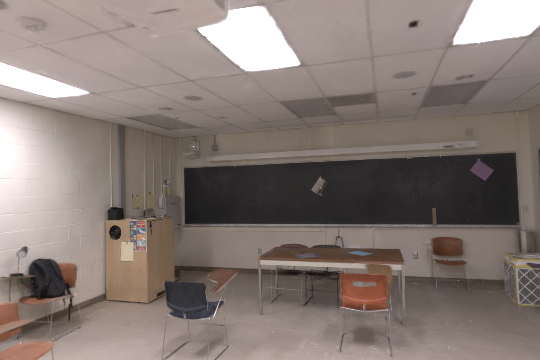}
    \caption{Capture}
  \end{subfigure}%
  \hspace{\paperfiggap}%
  \begin{subfigure}[t]{\editw}
    \captionsetup{position=bottom,skip=2pt}
    
    \label{fig:real_edits-b}
    \includegraphics[width=\linewidth]{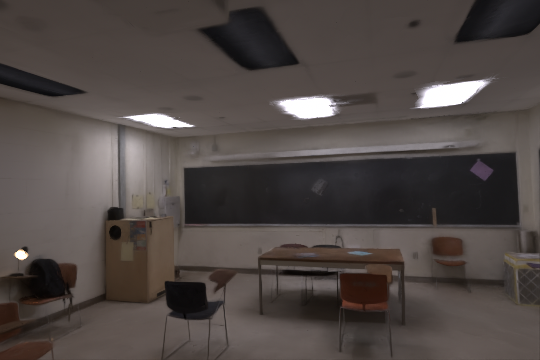}
    \caption{Relit 1}
  \end{subfigure}%
  \hspace{\paperfiggap}%
  \begin{subfigure}[t]{\editw}
    \captionsetup{position=bottom,skip=2pt}
    
    \label{fig:real_edits-c}
    \includegraphics[width=\linewidth]{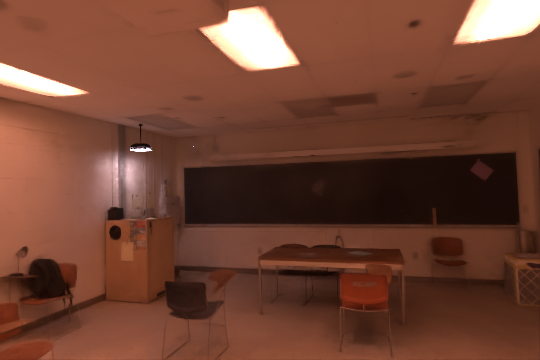}
    \caption{Relit 2}
  \end{subfigure}

  \par\vspace{\paperfigrowgroupgap}\nointerlineskip
  \begin{subfigure}[t]{\editw}
    \captionsetup{position=bottom,skip=2pt}
    
    \label{fig:real_edits-d}
    \includegraphics[width=\linewidth]{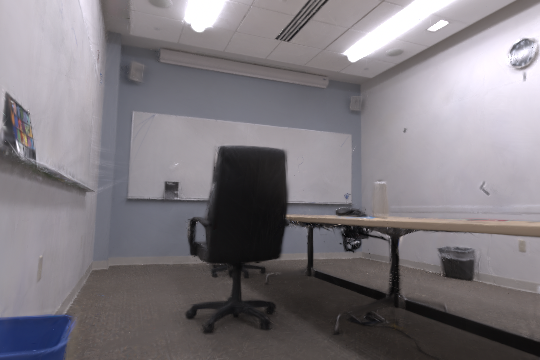}
    \caption{Capture}
  \end{subfigure}%
  \hspace{\paperfiggap}%
  \begin{subfigure}[t]{\editw}
    \captionsetup{position=bottom,skip=2pt}
    
    \label{fig:real_edits-e}
    \includegraphics[width=\linewidth]{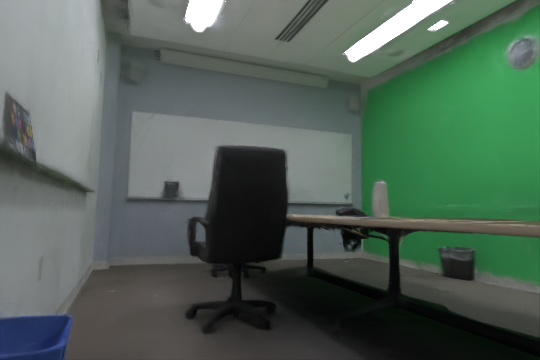}
    \caption{Relit 1}
  \end{subfigure}%
  \hspace{\paperfiggap}%
  \begin{subfigure}[t]{\editw}
    \captionsetup{position=bottom,skip=2pt}
   
    \label{fig:real_edits-f}
    \includegraphics[width=\linewidth]{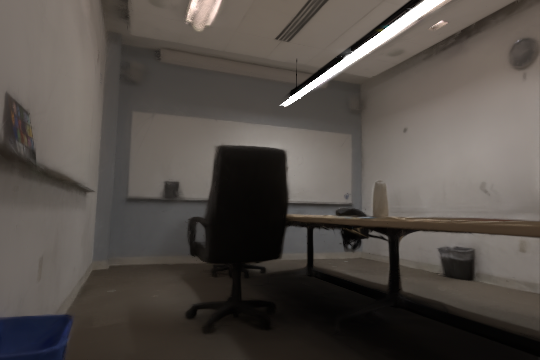}
     \caption{Relit 2}
  \end{subfigure}
  \caption{
  \textbf{Editing real indoor scenes.}
  (a,d)~Novel view under capture lighting.
  Top, \emph{FIPT-Classroom}~\cite{wu2023fipt}:
  (b)~captured ceiling panels switched off, and previously unlit desk lamp and panels switched on;
  (c)~an inserted pendant, with the ceiling panels dimmed and tinted.
  Bottom, \emph{FIPT-Conference}~\cite{wu2023fipt}:
  (e)~one wall's albedo edited, tinting its reflected light;
  (f)~an inserted panel light, with captured ceiling lights dimmed.
  }
  \label{fig:real_edits}
\end{figure}

\paragraph{Real-world environments.} 
\looseness=-1
For the reconstructions obtained from real-world captures, we apply emission, albedo, and emitter-insertion edits, including the activation of fixtures that were inactive during capture (Fig.~\ref{fig:real_edits}).
Inserted lights use emitter re-injection while retaining the room operator.
An edit reaches its first composited frame in 0.40--0.47\,s on \emph{FIPT-Classroom} and 0.36--0.52\,s on \emph{FIPT-Conference}.
The \emph{FIPT-Classroom} operators exceed the GPU's memory and are pruned, keeping $98\%$ of each row's form-factor sum (\cref{sec:supp-pruning}). 
These experiments demonstrate the ability to relight real indoor environments efficiently, resulting in natural appearance.

\begin{table*}[t]
  \centering
  \small
  \renewcommand{\arraystretch}{0.95}%
  \edef\padTsep{\the\dimexpr\belowrulesep/2\relax}%
  \edef\padBsep{\the\dimexpr\aboverulesep/2\relax}%
  \setlength{\aboverulesep}{0pt}\setlength{\belowrulesep}{0pt}%
  \def\padT{\rule{0pt}{\dimexpr\arraystretch\ht\strutbox+\padTsep\relax}}%
  \def\padB{\rule[-\dimexpr\arraystretch\dp\strutbox+\padBsep\relax]{0pt}{0pt}}%
  \setlength{\tabcolsep}{2pt}%
  \begin{tabular*}{\textwidth}{@{\extracolsep{\fill}}l*{12}{c}@{}}
    \toprule
    \padT & \multicolumn{4}{c}{Relit~1} & \multicolumn{4}{c}{Relit~2} & \multicolumn{4}{c@{}}{Relit~3} \\
    \cmidrule(r){2-5}\cmidrule(lr){6-9}\cmidrule(l){10-13}
    \padB Variant
      & PSNR$\uparrow$ & SSIM$\uparrow$ & FLIP$\downarrow$ & Time (s)$\downarrow$
      & PSNR$\uparrow$ & SSIM$\uparrow$ & FLIP$\downarrow$ & Time (s)$\downarrow$
      & PSNR$\uparrow$ & SSIM$\uparrow$ & FLIP$\downarrow$ & Time (s)$\downarrow$ \\
    \midrule
    \padT Ours (full, 200k proxy mesh) 
      & \textbf{35.95} & \textbf{0.983} & \textbf{0.139} & 0.244
      & \textbf{36.99} & \textbf{0.983} & \textbf{0.125} & 0.307
      & \textbf{36.96} & \textbf{0.984} & \textbf{0.117} & 0.287 \\
    Emission + NEE only
      & 22.02 & 0.659 & 0.681 & 0.111
      & 21.72 & 0.733 & 0.601 & 0.163
      & 20.26 & 0.819 & 0.643 & 0.148 \\
    Radiosity only 
    & 32.84 & 0.973 & 0.179 & 0.189
    & 35.52 & 0.979 & 0.132 & 0.210
    & 35.96 & 0.983 & \textbf{0.117} & 0.214 \\
    Proxy mesh with 100k triangles
      & 35.59 & 0.982 & 0.153 & 0.170
      & 36.87 & \textbf{0.983} & 0.129 & 0.227
      & 36.52 & 0.983 & 0.127 & 0.212 \\
    Nearest-centroid projection
      & 35.58 & 0.981 & 0.147 & 0.244
      & 36.46 & 0.981 & 0.135 & 0.306
      & 36.17 & 0.981 & 0.126 & 0.288 \\
    Unfitted projection $P^0$
      & 35.75 & 0.981 & 0.144 & 0.245
      & 36.62 & 0.981 & 0.132 & 0.309
      & 36.31 & 0.982 & 0.125 & 0.290 \\ 
    EAG-PT one-sided
      & 35.74 & \textbf{0.983} & 0.153 & 145.5
      & 34.82 & 0.980 & 0.167 & 142.0
      & 33.76 & 0.982 & 0.162 & 135.1 \padB \\
    \bottomrule
  \end{tabular*}
  \caption{
  \textbf{Ablation on \emph{Kitchen} against ground truth.}
  The top row is our full method, and each following row removes or replaces one component:
  the transport rows drop interreflection or replace the Gaussian leg with the projected first two radiosity orders, 
  the projection rows replace the fitted $P$, 
  the proxy mesh row halves the triangle budget,
  and the last row equips EAG-PT with our one-sided emission.
  }
  \label{tab:main_ablation}
\end{table*}

\subsection{Ablations}
\label{sec:exp-ablations}
\cref{tab:main_ablation} ablates the components of our method on \emph{Kitchen}. 

\paragraph{Transport decomposition.}
\looseness=-1
Replacing the Gaussian leg with the projected first two radiosity orders costs 1.00--3.11\,dB in the composite, while saving 55--97\,ms per edit, and dropping interreflection entirely costs 14--17\,dB.

\paragraph{Proxy-mesh budget.}
\looseness=-1
Halving the proxy mesh to 100k triangles lowers PSNR by 0.12--0.44\,dB and worsens FLIP in every condition, while reducing edit-to-first-image time to 0.17--0.23\,s.
Doubling it to 400k does not improve quality but raises the edit time to 0.41--0.49\,s  (\cref{tab:supp-proxy-budget}).

\paragraph{Projection fitting.}
\looseness=-1
With the reconstruction, radiosity solution, and direct light fixed, the fitted $P$ is best on every metric in every condition, improving composite PSNR by 0.37--0.79\,dB over nearest-centroid assignment and 0.20--0.65\,dB over the unfitted initialization $P^0$.
The choice of $P$ does not affect runtime: applying any of the three operators takes at most 0.5\,ms, negligible against the 0.24--0.31\,s edit. %
Indirect-only evaluation also favors the fitted $P$ (\cref{tab:supp-projection}).

\paragraph{EAG-PT one-sided.}
\looseness=-1
Equipping EAG-PT with our one-sided emission and reconstruction/albedo estimation removes the phantom light its two-sided emitters leak through the pendant's back.
Its rendering time stays essentially unchanged and its PSNR remains 0.21--3.20\,dB below ours.

\subsection{Proxy Mesh and Rendering Performance}
\label{sec:exp-runtime}
Reconstruction, proxy extraction, $F$, and $P$ are precomputed per scene. 
On the 200k proxies, $F$ takes 9--11\,min, 3.4--4.1\,GB compressed, and 6.5--8.1\,GiB of GPU memory, and loads in about 20\,s; extracting the proxy and fitting $P$ take 26--31\,min.
The 400k proxies of the real captures require 8.8--12.0\,GB and 16.4--23.4\,GiB, and load in 33--49\,s. 
Fixture selections (\cref{sec:emission-albedo}) are computed once and reused. %

The above experiments follow EAG-PT's convention of rendering at $540{\times}360$ pixels. %
In this setting, the first frame after editing finishes in less than 0.5\,s with our method; subsequent views render in 0.05--0.12\,s.
However, we also investigate our method's capability for realistic, higher-resolution rendering: 
at $1080{\times}720$ and $1620{\times}1080$, \emph{Living room} edit-to-first-image latency is 0.38--0.53\,s and 0.65--0.99\,s, respectively, %
while EAG-PT's per-view path tracing grows to 654--723\,s and 1445--1607\,s (\cref{sec:resolution-scaling}).

\section{Limitations and Future Work}
\label{sec:limitations}

Our formulation assumes Lambertian reflection and directly observed emitting surfaces; %
concealed sources require a richer representation, while inserted sources need specified geometry and emission.
Source assignment also relies on fixture masks and face areas (\cref{sec:emission-albedo}).
Proxy geometry limits indirect accuracy, while reconstructed emitter extent and ambiguous normal orientation affect direct illumination and material recovery.
The fitted projection cannot correct missing geometry or transport errors, and Eq.~\eqref{eq:hybrid-image} is approximate at fractional emitter coverage (\cref{sec:supp-coverage-blend}). %

Precomputation and storage grow with the number of surface interactions, even when sparse propagation is fast.
Edits to receiver geometry or opacity invalidate the transport matrix. %
Emitter re-injection updates visibility from the edited fixture to receivers but retains visibility from other emitters and between receivers; new occlusions along these paths, and indirect light on the moved housing, are not updated.
Evaluated geometry edits cover fixture scaling and translation.
Albedo edits preserve geometric exchange, but higher reflectance can require more orders; eight orders retain ${\sim}99\%$ of the indirect power in our synthetic conditions.
Improving capture-time material recovery and updating receiver visibility locally are promising directions for accuracy and edit scope. 

Our method enables fast editing and preview: a lighting edit updates the composite in 0.15--0.31\,s on synthetic and 0.36--0.52\,s on real scenes, and subsequent frames render at 8--22\,FPS ($540{\times}360$, RTX~4090), bounded by the NEE pass. %
Once a lighting condition is fixed, baking it into the Gaussians as in EAG-PT~\cite{yang2026eagpt} would reduce rendering to standard splatting (13--16\,ms per frame), left to future work.

Finally, relighting fidelity depends on the reconstructed geometry and materials.
We retain EAG-PT's albedo recovery~\cite{yang2026eagpt}, whose single-bounce estimate of incident illumination makes direct shading and interreflection sensitive to the capture lighting condition.
We consider this challenge complementary to interactive relighting: given the fast advances of radiance fields, we expect future inverse rendering improvements to integrate directly with our method.

\section{Conclusion}
We present a method for relighting radiance fields at interactive speeds that couples a surface transport solve with reconstructed Gaussians through a projection fitted under alpha-composited image formation.
Precomputed geometric exchange supports emission and albedo updates, while light source injection extends the formulation to fixture edits.
Our solution relights with sub-second latency and without per-edit optimization, faithfully simulates light transport for challenging free-hanging fixtures, scales to high-resolution output, and surpasses its path-traced baseline in fidelity and speed, paving the way for realistic illumination editing in radiance fields as an iterative authoring process.

\section*{Acknowledgments}
This work is supported by the Pioneer Centre for AI, DNRF grant number P1. We are grateful to Ioannis Siglidis and Lucia Gordon for their helpful input.

{
    \small
    \bibliographystyle{ieeenat_fullname}
    \bibliography{main}
}
\makeatletter\renewcommand\Hy@writebookmark[5]{}\makeatother %
\clearpage
\setcounter{page}{1}\renewcommand{\thepage}{S\arabic{page}}
\maketitlesupplementary
\setcounter{section}{0}\renewcommand{\thesection}{\Alph{section}}%
\setcounter{table}{0}\renewcommand{\thetable}{S\arabic{table}}%
\setcounter{figure}{0}\renewcommand{\thefigure}{S\arabic{figure}}%
\setcounter{equation}{0}\renewcommand{\theequation}{S\arabic{equation}}%
\renewcommand{\dblfloatpagefraction}{0.8}\renewcommand{\floatpagefraction}{0.8}\setlength{\floatsep}{20pt plus 2pt minus 2pt}\setlength{\dblfloatsep}{20pt plus 2pt minus 2pt}\makeatletter\setlength{\@fptop}{0pt}\setlength{\@dblfptop}{0pt}\makeatother %
\AtBeginEnvironment{table}{\setlength{\tabcolsep}{3pt}}\AtBeginEnvironment{table*}{\setlength{\tabcolsep}{3pt}}%
\renewcommand{\theHsection}{supp.\Alph{section}}\renewcommand{\theHtable}{supp.\arabic{table}}\renewcommand{\theHfigure}{supp.\arabic{figure}}\renewcommand{\theHequation}{supp.\arabic{equation}}%

\section{Direct Illumination Details}
\subsection{Bounded Gaussian Emitter Sampling}
\label{sec:supp-bounded-nee}
The density $q_i$ in Eq.~\eqref{eq:source-sampling} is normalized over the whole splat plane.
For ray intersection, the OptiX direct-light renderer represents each Gaussian's bounded support by two triangles forming the rectangle $D_i=\{\mathbf y:|u_i(\mathbf y)|\leq3,\ |v_i(\mathbf y)|\leq3\}$, with tangent coordinates $u_i,v_i$ in units of the Gaussian scales.
Opacity within this rectangle retains the Gaussian falloff, and emitter samples are restricted to the same domain.
This rectangle contains the fraction:
\begin{equation}
 c=\int_{D_i}q_i(\mathbf y)\,\mathrm dA
   =\operatorname{erf}(3/\sqrt{2})^2
   \approx0.9946
 \label{eq:nee-support-mass}
\end{equation}
of the density.
With ideal rejection sampling, accepted points have density $q_i/c$ inside $D_i$.
Our importance weights instead retain $q_i$.
With $\mathbf L_D$ denoting the direct light integral over these bounded footprints for the retained emitters, with the same receiver and visibility evaluation, the estimator has the expectation:
\begin{equation}
 \mathbb E[\widehat{\mathbf L}_d]=\frac{\mathbf L_D}{c}.
 \label{eq:nee-support-normalization}
\end{equation}
The omitted normalization therefore results in an overestimation of $\mathbf L_D$ by a factor $1/c\approx1.0054$, relative to this bounded-footprint integral.
This factor only accounts for the PDF normalization and does not bound the error of the complete renderer.
The implementation also clamps out-of-bounds coordinates, so Eq.~\eqref{eq:nee-support-normalization} is an idealized relation.
Shadow-ray origins are offset by 0.01 scene units along the receiver normal, and shadow rays stop $10^{-4}$ scene units short of the sampled point.

\section{Indirect Illumination Details}
\label{sec:supp-indirect-illumination}
\subsection{Proxy Mesh}
\begin{figure*}[t]
  \centering
  \paperfigcaptionstyle
  \setlength{\paperfigmeshbase}{0.495\linewidth}%
  \edef\meshw{\the\dimexpr0.495\paperfigmeshbase\relax}%
  \edef\meshgroupw{\the\dimexpr\meshw*2+\paperfiggap\relax}%
  \begin{subfigure}[t]{\meshgroupw}
    \centering
    \includegraphics[width=\meshw]{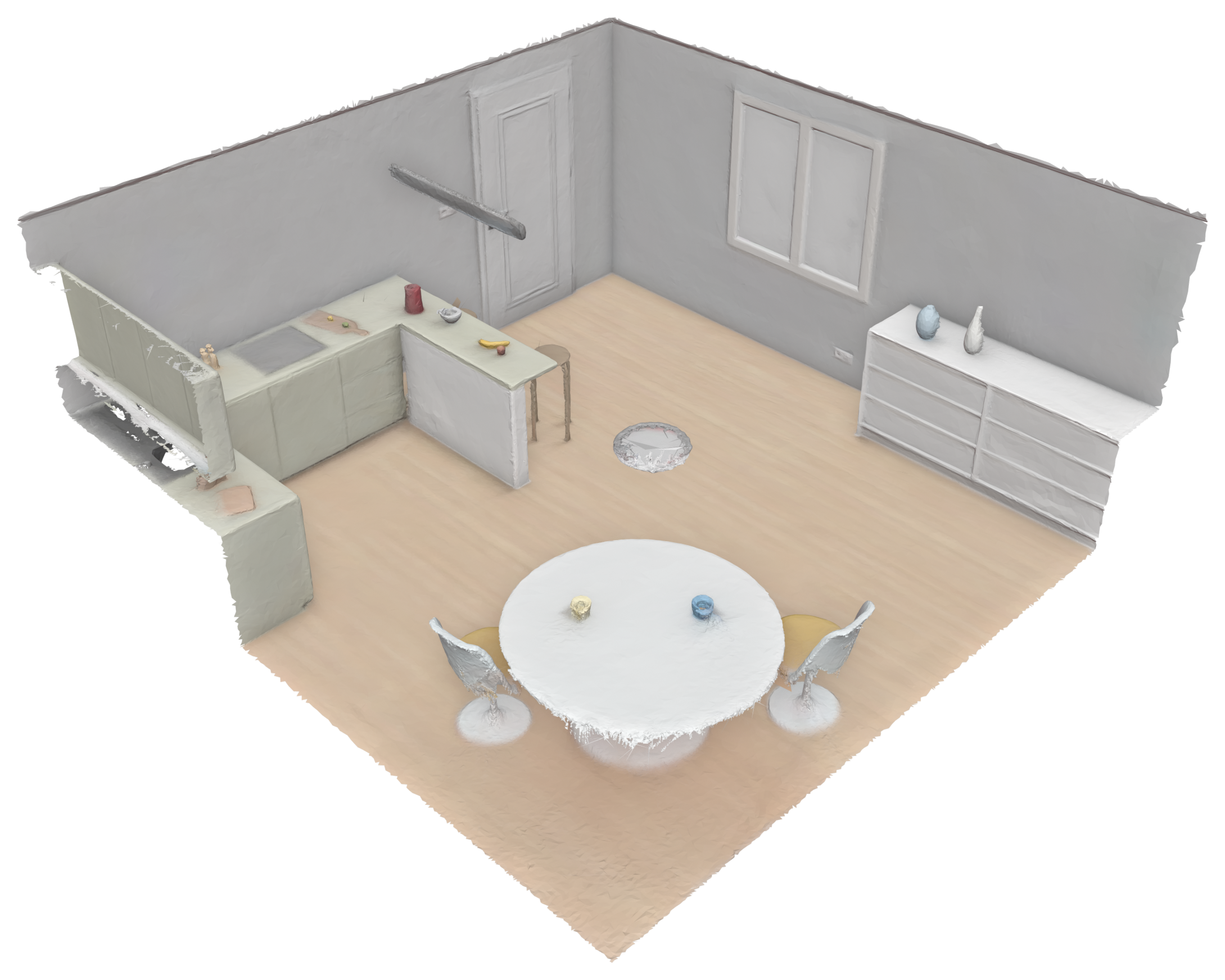}\hspace{\paperfiggap}%
    \includegraphics[width=\meshw]{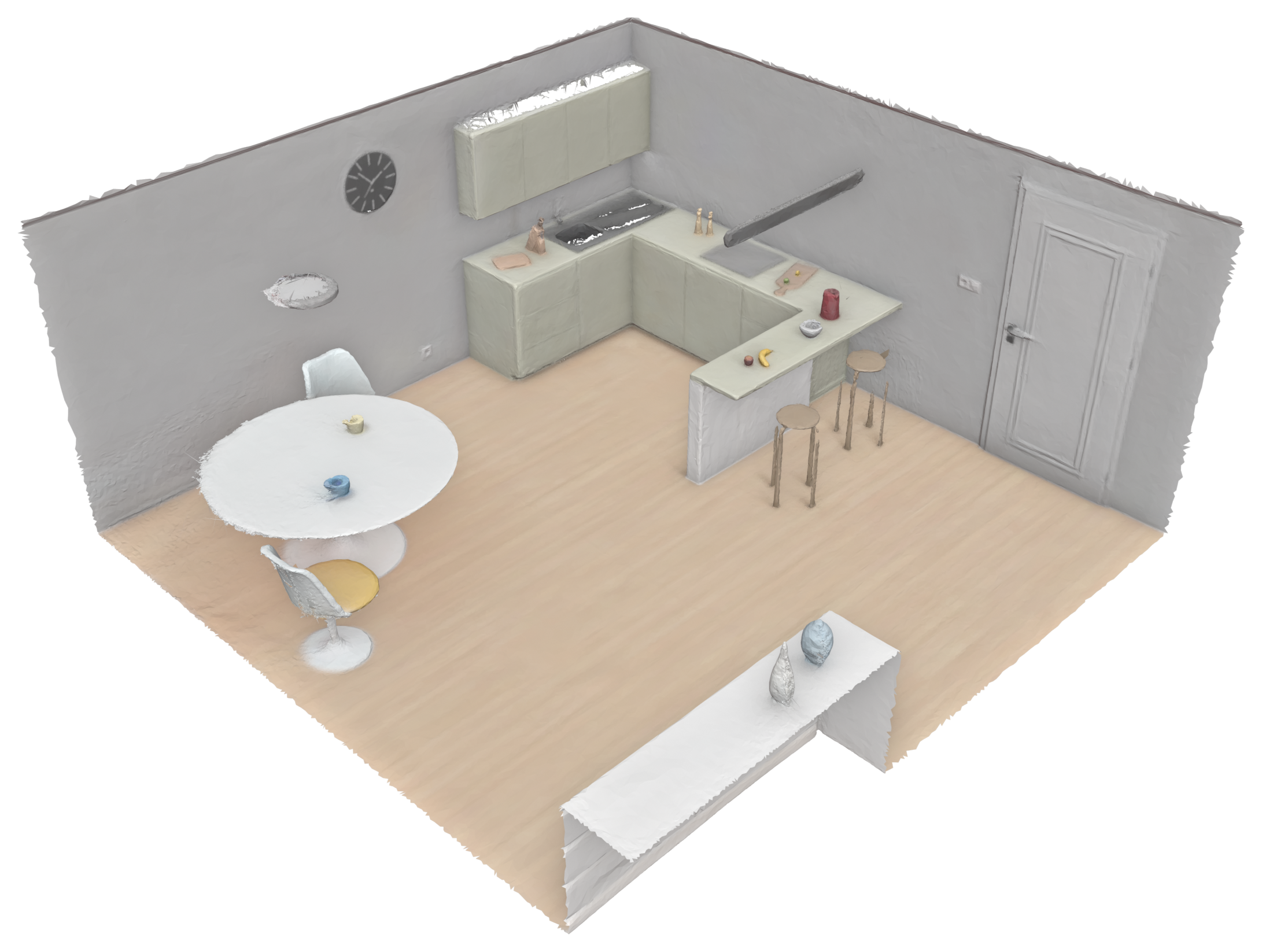}
    \caption{\emph{Kitchen}}
    \label{fig:proxy_meshes_kitchen}
  \end{subfigure}\hspace{\paperfiggroupgap}%
  \begin{subfigure}[t]{\meshgroupw}
    \centering
    \includegraphics[width=\meshw]{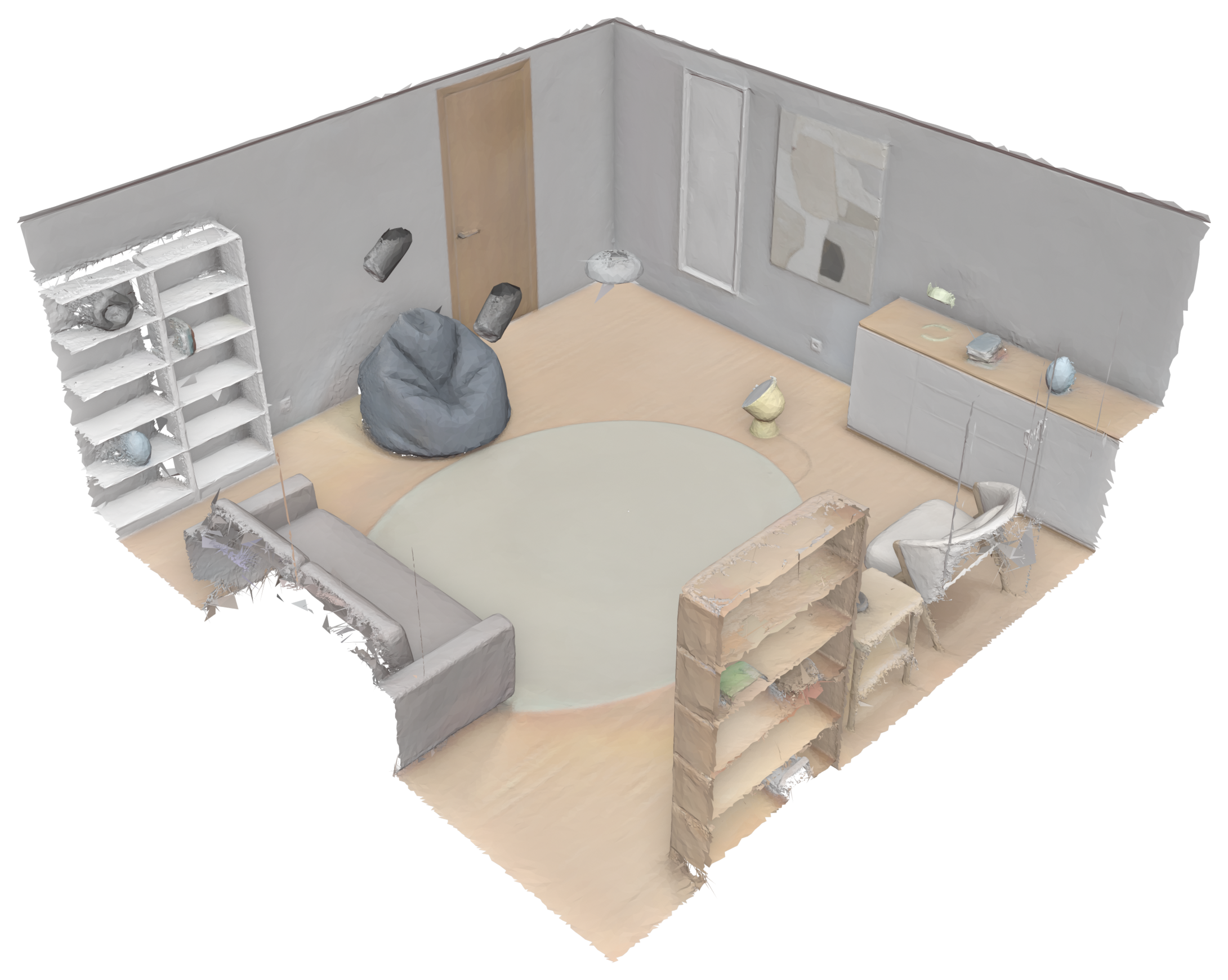}\hspace{\paperfiggap}%
    \includegraphics[width=\meshw]{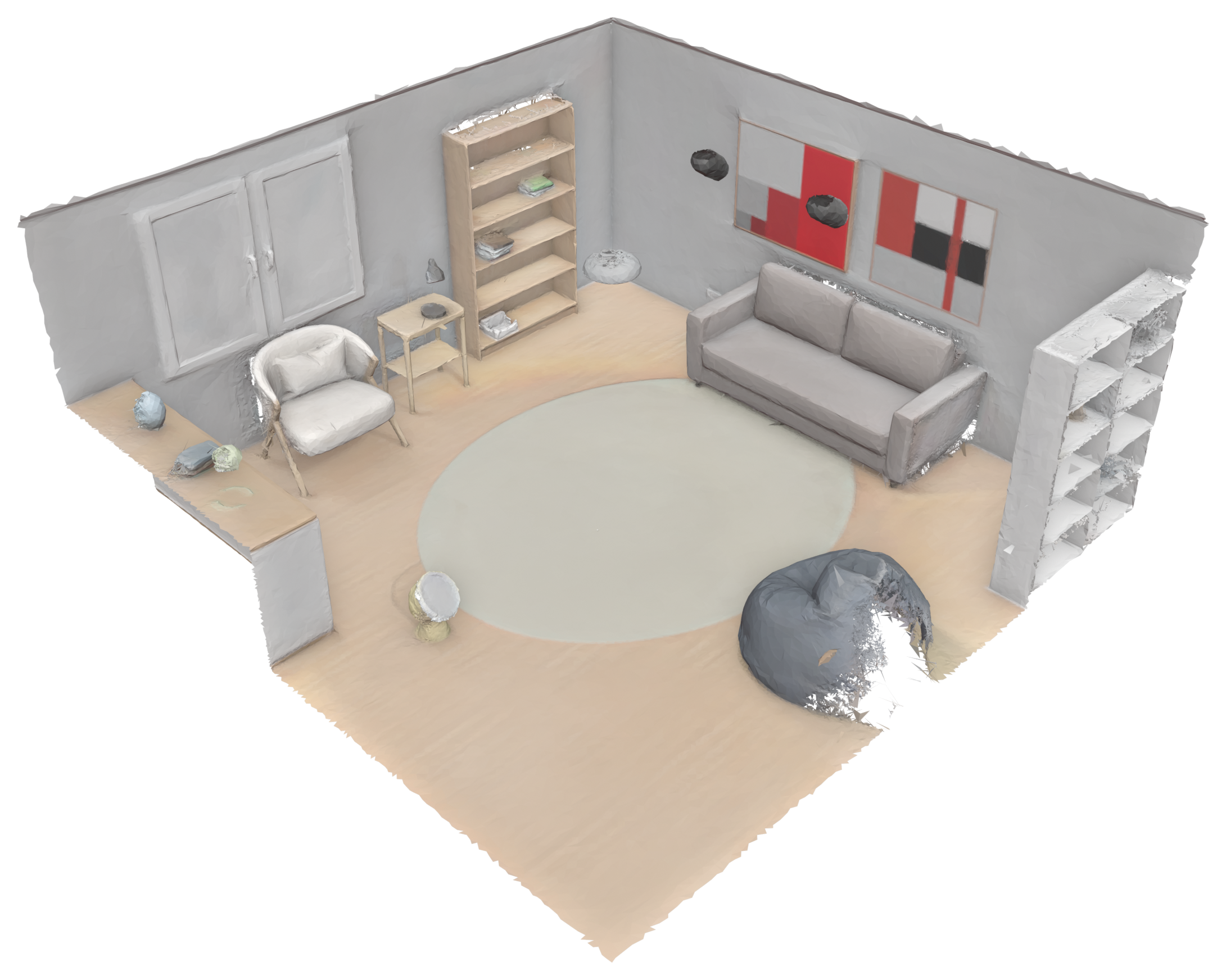}
    \caption{\emph{Living room}}
    \label{fig:proxy_meshes_livingroom}
  \end{subfigure}
  \caption{
  Proxy meshes of our two scenes used for the radiosity computation of the indirect illumination, each shown from two viewpoints.
  The meshes are extracted from the reconstructed 2D Gaussians by TSDF fusion and reduced to about 200k triangles by attribute-aware simplification.
  They are not meant to be accurate reconstructions and contain holes, floaters, and rough surfaces.
  However, they cover the large surfaces that dominate interreflections, such as walls, floor, and large furniture.
  The ceiling and the two walls nearest to the camera are removed for visualization only.
  }
  \label{fig:proxy_meshes}
\end{figure*}

The proxy mesh carries only diffuse interreflection.
Camera-visible emission and direct illumination remain on the Gaussians, so the proxy must cover the surfaces that dominate diffuse exchange rather than reproduce the reconstruction in full detail (\cref{fig:proxy_meshes}).

\paragraph{Extraction parameters.}
All length-type parameters scale with the scene size $D$, the diagonal of the per-axis 1\%-trimmed bounding box of the Gaussian positions, so one calibration transfers across scenes.
TSDF fusion uses the expected primary-ray depth rendered from the reconstruction, discarding pixels whose median depth lies more than 3\% behind it, with a voxel size that is a fixed fraction of $D$, a truncation distance of ten voxels, and for albedo-bake visibility a depth tolerance of six voxels or 1\% of depth, whichever is larger.
Simplification is quadric-error decimation with the Garland--Heckbert attribute extension:
each vertex is augmented with its albedo and normal, weighted in scene-normalized space so that material boundaries and creases are less likely to collapse while flat uniform regions collapse faster.
Conforming adaptive subdivision then splits every triangle whose longest edge exceeds a fixed fraction of $D$, bounding triangle extent for the solve, and sliver triangles are removed; if subdivision overshoots the triangle budget, the decimation target is lowered and the pass repeats. %
Reflectances are re-baked at the final centroids.
All constants are set in the released extraction code.

\subsection{Convergence of the Discrete Solve}
\label{sec:supp-radiosity-convergence}
We solve Eq.~\eqref{eq:radiosity} by Jacobi iteration from $B_0=E$:
\begin{equation}
 B_K=E+T_\rho B_{K-1}=\sum_{b=0}^{K}T_\rho^{b}E ,
 \label{eq:supp-neumann}
\end{equation}
so each term $T_\rho^{b}E$ carries the light that has undergone exactly $b$ diffuse reflections since emission.
Because $T_\rho$ acts per color channel, write $T_c=\operatorname{diag}(\boldsymbol\rho^M_{:,c})F$.
The series converges for every emission field $E$ to $B^*=(I-T_\rho)^{-1}E$ exactly when every spectral radius $\varrho(T_c)<1$.
Since $T_c$ is entry-wise nonnegative, its spectral radius is bounded by the maximum row sum:
\begin{equation}
 \varrho(T_c)\;\leq\;\|T_c\|_\infty
 =\max_j\;\rho^M_{jc}\sum_k F_{jk},
 \label{eq:supp-rowsum-bound}
\end{equation}
and we define the contraction factor $\beta=\max_c\|T_c\|_\infty$.

For exact form factors in a closed scene, rows of $F$ sum to at most one, so any physically valid albedo $\boldsymbol\rho^M<1$ gives $\beta<1$~\cite{cohen1993radiosity}.
Finite hemicube quadrature, however, does not inherit this guarantee.
The delta form factors of the discrete hemicube slightly over-integrate the hemisphere, so a fully enclosed patch acquires a row sum marginally above one, and $\beta<1$ must therefore be checked on the discretized operator rather than inferred from albedo alone.
We verify it directly on both benchmark operators (Tab.~\ref{tab:supp-beta}).
The maximum row sum of $F$ is 1.000135 in both scenes.
It is attained by fully enclosed patches and matches the total quadrature weight of the resolution-64 hemicube, confirming that the overshoot stems from the quadrature.
Production solves assign zero albedo to all patches that emit in the capture or in the target condition, consistent with the nonreflecting-emitter assumption, which can only lower row sums of $T_\rho$, so the contraction factors of Tab.~\ref{tab:supp-beta}, measured with the baked proxy albedo, bound every emission condition.
The solve therefore converges on both benchmark operators.

\begin{table}[t]
\centering
\small
\renewcommand{\arraystretch}{0.95}%
\edef\padTsep{\the\dimexpr\belowrulesep/2\relax}%
\edef\padBsep{\the\dimexpr\aboverulesep/2\relax}%
\setlength{\aboverulesep}{0pt}\setlength{\belowrulesep}{0pt}%
\def\padT{\rule{0pt}{\dimexpr\arraystretch\ht\strutbox+\padTsep\relax}}%
\def\padB{\rule[-\dimexpr\arraystretch\dp\strutbox+\padBsep\relax]{0pt}{0pt}}%
\begin{tabular}{@{}lrr@{}}
\toprule
\padT\padB & \emph{Kitchen} & \emph{Living room} \\
\midrule
\padT Triangles $N_T$      & 198{,}682 & 198{,}019 \\
Stored entries of $F$      & $1.08\times10^9$ & $1.01\times10^9$ \\
Mean entries per row       & 5{,}459 & 5{,}116 \\
Max row sum of $F$         & 1.000135 & 1.000135 \\
Rows with sum $>1$         & 6{,}360 & 2{,}519 \\
\padB Contraction $\beta$        & 0.974 & 0.973 \\
\bottomrule
\end{tabular}
\caption{
Discrete transport operators of the two benchmark scenes and their measured contraction factors.
Row sums exceed one only through hemicube quadrature, and $\beta<1$ guarantees convergence of Eq.~\eqref{eq:supp-neumann} for any emission.}
\label{tab:supp-beta}
\end{table}

\subsection{Truncation Residual and the Choice of K}
\label{sec:supp-truncation}
For the partial sum $B_K=\sum_{b=0}^{K}T_\rho^{b}E$, the residual of Eq.~\eqref{eq:radiosity} is the next reflected order:
\begin{equation}
 r_K=E+T_\rho B_K-B_K=T_\rho^{K+1}E .
 \label{eq:supp-residual-def}
\end{equation}
With the elementwise norm $\|Z\|_{\max}=\max_{j,c}|Z_{jc}|$ and the contraction factor $\beta$ of Sec.~\ref{sec:supp-radiosity-convergence}, the converged discrete solution satisfies, channelwise by the Neumann series:
\begin{equation}
 \|B^*-B_K\|_{\max}\;\leq\;\frac{\|r_K\|_{\max}}{1-\beta}.
 \label{eq:supp-residual-bound}
\end{equation}
The residual costs one additional sparse transport step, so the truncation error of any solve can be bounded a posteriori at roughly $1/K$ of the solve's own cost.
The production solve uses a fixed eight-step budget and does not evaluate this residual as a stopping criterion.
The bound concerns truncation of the fixed discrete solve only.

In practice the per-order power decays geometrically well below this worst-case rate, because $\beta$ bounds the single worst row while the propagating energy is governed by the operator's spectral behavior under the actual emission.
Let $\Delta P_b = \pi \sum_j \mathcal A_j \sum_c[T_\rho^{b}E]_{jc}$ denote the summed RGB diffuse power of reflection order $b$, where $\mathcal A_j$ is the area of triangle $j$ and $\pi B$ is the radiant exitance of outgoing radiance $B$.
We measured the series on the \emph{Kitchen} operator for all three evaluated conditions, extending the production solve to a 30-bounce reference.
Successive ratios $\Delta P_{b+1}/\Delta P_b$ lie between 0.49 and 0.54 for orders 2 through 20 in every condition, well below $\beta$ (Tab.~\ref{tab:supp-beta}).

Tab.~\ref{tab:supp-truncation} reports each order's share of the 30-bounce reference power.
Relative to this reference, orders 9--30 contribute at most 0.28\%.
The composite of Eq.~\eqref{eq:hybrid-image}, however, uses only $B_{\rm ind}$, the orders two through $K$, since the Gaussian leg supplies emission and the first reflection.
Relative to the reference's indirect component (orders 2--30), the omitted power is 0.96--1.06\% across conditions.
These measurements support $K=8$ for the evaluated \emph{Kitchen} conditions.
Increasing reflectance can slow this decay.
After an albedo edit, the residual bound remains applicable if the updated operator satisfies $\beta<1$, which must be reassessed for the changed reflectances.

\begin{table}[t]
\centering
\small
\renewcommand{\arraystretch}{0.95}%
\edef\padTsep{\the\dimexpr\belowrulesep/2\relax}%
\edef\padBsep{\the\dimexpr\aboverulesep/2\relax}%
\setlength{\aboverulesep}{0pt}\setlength{\belowrulesep}{0pt}%
\def\padT{\rule{0pt}{\dimexpr\arraystretch\ht\strutbox+\padTsep\relax}}%
\def\padB{\rule[-\dimexpr\arraystretch\dp\strutbox+\padBsep\relax]{0pt}{0pt}}%
\begin{tabular}{@{}lrrr@{}}
\toprule
\padT\padB Order $b$ & Relit~1 & Relit~2 & Relit~3 \\
\midrule
\padT 0 (emission)   & 49.42\% & 53.21\% & 48.66\% \\
\padB 1 (direct)           & 24.89\% & 23.29\% & 25.11\% \\
\midrule
\padT 2              & 12.66\% & 11.69\% & 12.73\% \\
3                    &  6.35\% &  5.78\% &  6.50\% \\
4                    &  3.25\% &  2.94\% &  3.36\% \\
5                    &  1.66\% &  1.50\% &  1.74\% \\
6                    &  0.85\% &  0.77\% &  0.91\% \\
7                    &  0.44\% &  0.40\% &  0.47\% \\
\padB 8                    &  0.23\% &  0.21\% &  0.25\% \\
\midrule
\padT Tail (orders 9--30)  &  0.25\% &  0.23\% &  0.28\% \\
\padB Tail relative to orders 2--30 & 0.97\% & 0.96\% & 1.06\% \\
\bottomrule
\end{tabular}
\caption{
Measured reflection-order decomposition on the \emph{Kitchen} operator for the three evaluated relighting conditions, against a 30-bounce reference.
Entries are the share of the reference's total power.
Orders 0 and 1 are supplied by the Gaussian leg (visible emission and NEE), and the radiosity leg contributes orders 2 through 8.
The omitted orders 9--30 are quoted against both the reference's total power and its indirect power (orders 2--30).
}
\label{tab:supp-truncation}
\end{table}

\subsection{Form-Factor Pruning}
\label{sec:supp-pruning}
Each row of $F$ stores the patches visible from one centroid, so the entry count grows with the scene's mutual visibility rather than with triangle count alone, and a more open scene yields denser rows. The sparsity of $F$ is therefore a scene property.
Our benchmark operators store $1.08\times10^9$ and $1.01\times10^9$ entries at comparable triangle counts (Tab.~\ref{tab:supp-beta}).
Per-row pruning reduces storage and sparse propagation cost when the operator exceeds the available memory.
Keeping the $k$ largest entries of each row caps the matrix at $N_Tk$ entries and hence bounds the sparse work per step, while keeping the fewest largest entries that reach a fraction $f$ of each row's sum adapts the cut to how concentrated that row's exchange is.
Both operations only remove nonnegative entries, so row sums and $\beta$ can only decrease, preserving the convergence guarantee of Sec.~\ref{sec:supp-radiosity-convergence}.

Tab.~\ref{tab:supp-pruning} quantifies this trade on the \emph{Kitchen} operator of Tab.~\ref{tab:supp-beta}.
Reducing the entry count also discards transport mass; the resulting image-quality trade-off is evaluated in Tab.~\ref{tab:supp-ff-prune}.
All synthetic-scene solves (except in the pruning ablation) use unpruned matrices. The \emph{FIPT-Classroom} operators ($3.13$--$3.14\times10^9$ entries, 23.3--23.4\,GiB) exceed the available memory budget on the 24\,GB GPU, so their timings use $f=0.98$, which keeps 88\% of the entries (20.6--20.7\,GiB).
Against the unpruned solve, the indirect field keeps 94.7--94.9\% of its energy in the evaluated \emph{FIPT-Classroom} conditions.
These numbers measure different quantities: Tab.~\ref{tab:supp-pruning} reports retained row mass of $F$, whereas the solved field depends on where radiance falls relative to the discarded coefficients.

\begin{table}[t]
\centering
\small
\renewcommand{\arraystretch}{0.95}%
\edef\padTsep{\the\dimexpr\belowrulesep/2\relax}%
\edef\padBsep{\the\dimexpr\aboverulesep/2\relax}%
\setlength{\aboverulesep}{0pt}\setlength{\belowrulesep}{0pt}%
\def\padT{\rule{0pt}{\dimexpr\arraystretch\ht\strutbox+\padTsep\relax}}%
\def\padB{\rule[-\dimexpr\arraystretch\dp\strutbox+\padBsep\relax]{0pt}{0pt}}%
\begin{tabular}{@{}lcc@{}}
\toprule
\padT\padB Pruning & Entries kept & Row mass kept \\
\midrule
\padT None               & $1.08\times10^9$ \hfill (100\%) & 100.0\% \\
$f=0.99$ & $0.98\times10^9$ \hfill (90.5\%) & 99.0\% \\
$f=0.90$ & $0.69\times10^9$ \hfill (63.5\%) & 90.1\% \\
$f=0.80$ & $0.49\times10^9$ \hfill (45.6\%) & 80.2\% \\
top-$k$, $k{=}3000$      & $0.53\times10^9$ \hfill (49.2\%) & 83.2\% \\
\padB top-$k$, $k{=}1000$      & $0.19\times10^9$ \hfill (17.6\%) & 60.1\% \\
\bottomrule
\end{tabular}
\caption{
Per-row pruning of the \emph{Kitchen} operator of Tab.~\ref{tab:supp-beta}.
$f$ denotes the per-row mass threshold.
Row mass kept is the retained fraction of the sum of all entries of $F$, not the retained power of a nonuniform illumination field.
Per-step solve time scales with stored entries.}
\label{tab:supp-pruning}
\end{table}

\section{Projection Operator Details}
\label{sec:supp-projection}
\paragraph{Candidates and prior.}
For every Gaussian, an exact closest-point query against the proxy BVH plus a local five-ring expansion around the closest triangle proposes candidate triangles, rejecting candidates farther from the surface than the hit-association distance gate below or $2\times$ the Gaussian's mean tangent scale (whichever is larger).
Each remaining candidate receives a geometric score $g_{it}$, the product of a tangent-plane Mahalanobis kernel $\exp[-\tfrac12(d_u^2/s_{i1}^2+d_v^2/s_{i2}^2)]$ with tangent scales floored at 0.01 scene units, a normal-displacement kernel $\exp[-\tfrac12(d_n/\sigma_n)^2]$ with $\sigma_n$ half the distance gate, and normal agreement $\max(0,\cos\theta)^2$ after aligning the sign of the Gaussian normal with its closest triangle, since reconstructed Gaussian normals are sign-ambiguous.
Here $(d_u,d_v,d_n)$ are the components of the displacement from the Gaussian center to the closest point of triangle $t$ along the Gaussian's tangent axes and aligned normal, and $\theta$ is the angle between the aligned normal and the triangle normal.
Training-view hit evidence adds triangles to these proposals and re-ranks them: each cached renderer hit of Sec.~\ref{sec:indirect} is associated with the triangle closest to its position, and the association is valid if that triangle lies within four TSDF voxels, its normal has a positive dot product with the aligned Gaussian normal, and it belongs to the same mesh component as the triangle closest to the Gaussian center.
Valid hits accumulate, per Gaussian $i$ and triangle $t$, the compositing mass $M_{it}$ and the number of supporting views.
With total mass $M_i=\sum_tM_{it}$ and saturating confidence $c_i=M_i/(M_i+m_0)$, $m_0=0.5$, candidates are ranked by $c_i\,M_{it}/M_i+(1-c_i)\,g_{it}$ with $g_{it}$ normalized per row, blending view-mass share with geometric evidence. %
Gaussians without valid hits have $c_i=0$ and are ranked by $g_{it}$ alone.
Up to 64 merged candidates per Gaussian are truncated to the 32 highest-scoring slots, and, renormalized, initialize the prior $P^0$.

\paragraph{Parameterization.}
$P$ is stored as a fixed-width array of 32 candidate triangle indices and logits per supported Gaussian, with rows realized by a masked softmax throughout optimization.
Supported rows are therefore nonnegative with unit sums by construction, which yields the constant-field preservation used in the main text.
Gaussians without a compatible candidate have zero rows.
These are predominantly floating, low-opacity Gaussians (36 of 200k in \emph{Kitchen} and 163 of 200k in \emph{Living room}).
At deployment they are omitted from the indirect render.

\paragraph{Probe fields.}
Training fields comprise analytic probes and solved radiosity fields.
The analytic families we use are constant fields, random fields, mesh-smoothed noise at three spatial scales, connected surface regions at four sizes, and hierarchical surface partitions, totaling 48 fields.
The solved family contains full radiosity solutions generated from 48 randomized emission regions using the production solve (with eight transport steps and zero emitter albedo on emitting patches); the few solutions that remain near-dark are discarded.
Training samples the solved family with weight 0.7 and the analytic family with weight 0.3.
The checkpoint is selected on held-out training views by the mean of two normalized errors: the error on held-out probe fields, which include indirect-only solutions $B_{\rm ind}$, and the error of the capture term.

\paragraph{Objective and schedule.}
The Charbonnier distance is $d(x)=\sqrt{x^2+\epsilon^2}-\epsilon$ with $\epsilon=10^{-3}$, applied to linear radiance and, with weight 0.3, to  $\log(1+x)$-compressed radiance, and averaged over pixels with weights $m_v$ ($d_1$ in Eq.~\eqref{eq:projection} uses unit weights).
The loss weights of Eq.~\eqref{eq:projection} are $\lambda_{\rm cap}=2$, $\lambda_{\rm cyc}=0.09$, and $\lambda_{\rm prior}=0.045$.
The reverse operator uses the same masses as the prior, $(Q\mathbf c)_t=\sum_iM_{it}\mathbf c_i/\sum_iM_{it}$, and the cycle loss is $\mathcal L_{\rm cyc}=\sum_tr_t\|(QPz)_t-z_t\|^2/\sum_tr_t\|z_t\|^2$, where $r_t$ is the number of views observing triangle $t$ relative to the maximum over triangles.
Triangles without valid hits have $(Q\mathbf c)_t=0$ and $r_t=0$, so they do not enter $\mathcal L_{\rm cyc}$.
The prior averages $D_{\rm KL}(P_i\|P_i^0)$ with row weights $\beta_i$ of 0.3 for Gaussians with $M_i\ge0.05$ from at least two views, 10 for the remaining rows with a single candidate, and 3 otherwise, each multiplied by $1.5-0.5\min(1,M_i/0.2)$.
A flux-conservation ratio through $Q$ is logged as a diagnostic but never optimized.
We fit with Adam at learning rate 0.05 for 2,500 steps with 200 warmup steps and cosine decay, sampling four cached training views and four probe fields per step, with validation every 250 steps.
The fit runs once per scene, and applying $P$ at runtime is one gather and one weighted sum over at most 32 entries per Gaussian.

\section{Composition Details}
\label{sec:supp-composition}
\subsection{Blending at Fractional Emitter Coverage}
\label{sec:supp-coverage-blend}
The factor $(1-w_v^e)$ in Eq.~\eqref{eq:hybrid-image} assigns reflected light to the non-emissive pixel fraction: it prevents adding reflection where emission is already composited and suppresses indirect radiance that the projection interpolates onto nonreflecting emitter surfels, whose own triangles carry no indirect radiance but whose $P$ rows mix neighboring triangles.
Emitter hits are not excluded elsewhere: they remain occluders in the transmittance of $W_v^{\rm ind}$, and the receiver attributes of \cref{sec:direct} composite their zero albedo.
At fractional coverage $0<w_v^e<1$ the blend is therefore an image-space approximation: reflected light composited behind a partially covering emitter is attenuated once by transmittance and again by $(1-w_v^e)$, and, as a scalar per-pixel factor, $(1-w_v^e)$ does not resolve depth ordering, so reflection composited in front of an emitter is scaled as well.
Both effects would require per-hit separation of emission and reflection, and both vanish where $w_v^e=0$ or emission saturates the pixel.
The blend assumes opaque emitting surfaces and does not model translucent fixtures.

\section{Emission Edits}
\label{sec:supp-emission-edits}
\subsection{Editing Inputs and Data Provenance}
\label{sec:supp-editing-inputs}
Our pipeline distinguishes four kinds of data. 
Reconstructed information: Gaussians, albedos, the proxy mesh, $F$, and $P$ are derived from the capture-condition reconstruction, which, following EAG-PT~\cite{yang2026eagpt}, uses posed linear-radiance images, normal maps, and masks of the fixtures lit at capture. The depth maps gating triangle selection are rendered from the reconstruction. 
User editing controls: each edit selects its fixtures through capture-view masks and specifies their radiances and emitting-face areas, with authored 3D anchors grouping the selection into independently controllable fixtures.
Fixtures that never emit during capture are visible in the capture images and are marked there by hand or by raycasting a marked 3D region against the proxy, as done for the real scenes. 
Benchmark annotations: for the synthetic benchmarks these controls are generated from the modeled scenes, with binary emitter masks rendered in the training views that carry no target-condition shading. 
Ground truth, used for evaluation only: path-traced renders of each target condition on held-out test views, never read during editing.

\subsection{Source-Area Calibration}
\label{sec:supp-area-calibration}
Benchmark source selection votes proxy triangles from the target condition's emitter masks, gated by rendered depth.
The selected area $\mathcal A_{\rm sel}$ of a fixture can exceed its authored emitting-face area $\mathcal A_{\rm auth}$, since voting whole triangles from coverage masks over-covers small faces.
Injecting the full fixture radiance on every selected triangle would then mis-scale the injected power by exactly this area ratio.
With $\boldsymbol\ell_f$ denoting the RGB radiance of fixture $f$, shared by its emitting Gaussians ($\boldsymbol\ell_i=\boldsymbol\ell_f$, \cref{sec:composition}), we therefore assign each selected triangle of fixture $f$ the emission:
\begin{equation}
E_j=\boldsymbol\ell_f\cdot\min\!\left(1,\;\frac{ \mathcal A_{\rm auth}}{\mathcal A_{\rm sel}}\right),
 \label{eq:supp-area-cap}
\end{equation}
which restores the authored per-channel power $\pi \mathcal A_{\rm auth}\boldsymbol\ell_f$ exactly for over-covered opaque fixtures: a slightly larger but proportionally dimmer patch carries the intended power.
Under-coverage stems from missing proxy geometry, and we do not compensate for it by brightening the remaining triangles.
The calibration is computed when a condition's emission is assigned and is retained across geometry edits.

\paragraph{Area provenance.}
$\mathcal A_{\rm auth}$ is a user editing control: together with the target radiance it specifies the fixture's radiant power $\pi \mathcal A_{\rm auth}\boldsymbol\ell_f$.
For the synthetic benchmarks it is generated from the modeled scenes together with the other annotations, and inserted fixtures carry the area of their authored geometry.
For the real captures no model exists, and the area is estimated from the reconstruction alone: the fixture's masks or a marked 3D region are raycast onto the proxy, and $\mathcal A_{\rm auth}$ is the summed area of the depth-gated face triangles.
For fixtures selected directly from such masks, $\mathcal A_{\rm auth}$ coincides with $\mathcal A_{\rm sel}$ and the cap is inactive.

\paragraph{Area treatment in direct light.}
Proxy triangles tile the surface without overlap, so injected power scales with the selected area, and Eq.~\eqref{eq:supp-area-cap} restores the authored power by dimming over-coverage.
For overlapping front-facing Gaussian emitters with equal radiance, alpha compositing along each ray reduces their combined emission to that radiance times their accumulated coverage, including attenuation by intervening occluders.
Since shadow rays stop $10^{-4}$ scene units short of the sampled point, emitter layers closer to it than that add without mutual attenuation.
This avoids treating overlapping layers as independent opaque sources, but does not guarantee that their reconstructed coverage matches the authored emitting face.
Assigning unequal per-splat radiances no longer permits this common-radiance factorization; a uniform fixture-wide rescaling preserves it.
On the Gaussian leg, the area input only shapes the emissive weights $e_i$ of turn-on fixtures, and the emitting extent is bounded by the selection, which admits a splat only when most of its compositing weight over the training views falls inside the fixture mask; a proxy triangle is selected when at least half of the pixels it covers are masked.
EAG-PT receives the same masks and target radiances and likewise assigns uniform radiance to its selected splats, so the methods differ in emitter selection and proxy calibration rather than in the per-splat assignment.

\section{Fixture Edits}
\subsection{Similarity Edits on Gaussians}
An edit starts from a per-fixture selection of Gaussians.
Fixtures are segmented once per scene from the reconstruction's emissive clusters near known fixture anchors, grown by a rim step that attaches the adjacent non-emitting housing.
Relocation edits additionally include the fully opaque housing and cord, which are invisible to emissive clustering, through two explicit spatial regions, so the whole assembly travels together.

The evaluated fixture edits are similarities, which preserve an orthonormal tangent frame through rotation and apply one shared scale. 
A general invertible affine map would also yield an ellipse representable by two scales and a rotation after refactorizing its tangent covariance. 
Similarities permit the direct parameter updates used here and preserve the fixture's internal overlap structure. 
The oriented normal of Sec.~\ref{sec:onesided} is transformed with the fixture and never re-derived, so one-sidedness survives every edit.

\subsection{Re-injection Scene and Frozen Operators}
\label{sec:supp-fixture-scene}

\paragraph{Temporary tracing scene.}
Fixture edits leave the proxy mesh unchanged.
The temporary scene of Sec.~\ref{sec:injection} exists only for ray queries: a static acceleration structure holds the frozen proxy with the fixture's triangles removed, both its emitting surface and surrounding body, and a second, per-edit structure holds their transformed copy.
Excluding the fixture from the static structure keeps its vacated location from occluding the fresh injection, and the transformed copy lets a moved or scaled fixture shadow its own emission.
The opening this leaves in the fused surface is not repaired. Rays passing through it deposit nothing, and receiver triangles, areas, and indices are unchanged, which preserves the discretization behind $F$, $T_\rho$, and $P$.
Rays resolve at their nearest hit under the front-face-only exchange of the form factors: a first hit on the fixture body absorbs the ray, since emitting surfaces are nonreflecting under the source split and reflection off the rest of the selected body is omitted from the fresh injection; a receiver hit on the back face absorbs; a front-face receiver hit deposits.
Nearest-hit resolution also covers transformed fixture triangles that intersect static geometry, with no boolean operations on the mesh.
In Eq.~\eqref{eq:fixture-injection}, the factor $\pi$ of the flux $\pi\mathcal A_k'E_k$ leaving a Lambertian source triangle cancels the $1/\pi$ of the receiver's Lambertian BRDF, consistent with the $\pi$-normalized form factors of Sec.~\ref{sec:indirect}.

\paragraph{Frozen operators and the moved fixture.}
Beyond the fresh injection, the fixture's own entries in the frozen operators are retained rather than masked.
Its triangles stop emitting through $E_{\neg f}$ and receive no fresh injection, but they remain receivers at their pre-edit position inside the retained gathers $T_\rho^b$, and $P$'s triangle-to-Gaussian correspondence is frozen, so the fixture's Gaussians are shaded with indirect radiance sampled at the pre-edit proxy location.
Through $T_\rho$ the triangles at the vacated position keep receiving and re-radiating light in the retained orders, while the stale $P$ rows affect only the shading of the fixture's own Gaussians.
On the emitting surface the composite suppresses projected indirect radiance wherever emission saturates the pixel (\cref{sec:supp-coverage-blend}), so the stale rows mainly concern the non-emitting housing.
Tab.~\ref{tab:fixture_edits} evaluates the combined effect against path-traced references of the edited scenes, including pendant relocations up to 2\,m.

\subsection{Compact In-Place Scaling}
\label{sec:alphasq}
For compact in-place fixture scalings, a simpler alternative to our emitter re-injection method (described in Sec.~\ref{sec:injection}) approximates the indirect change by rescaling the fixture's original indirect field.
It assumes that scaling a compact fixture in place changes mainly how much indirect light the scene receives, and hardly where that light lands.
We observe this directly in path-traced ground truth, where the normalized indirect field stays nearly unchanged across fixture scales.
Gaussian geometry and direct illumination are updated exactly as described in the main paper; only the indirect update differs.

\paragraph{Indirect-field update.}
Let $\mathcal G_f^e$ contain the fixture's emitting Gaussians, and let $a>0$ be the in-place scale factor.
At fixed source radiance, the opacity integral satisfies $A_i'=a^2A_i$.
Hence the summed source sampling weight $\sum_{i\in\mathcal G_f^e}\bar\ell_i A_i$ scales by $\eta=a^2$. 
If the edit should instead preserve the fixture's power, dividing the emitter radiances by $a^2$ gives $\eta=1$.
The frozen-pattern update carries this factor over to indirect light by scaling the fixture's share of the proxy emission $E_f$.
Let $E=E_{\neg f}+E_f$, where $E_f$ includes the fixture's initial source-area calibration and $E_{\neg f}$ contains emission from all other sources.
We retain $E_f$'s support and calibration factor, and set $E'=E_{\neg f}+\eta E_f$.
At fixed $T_\rho$, linearity gives:
\begin{equation}
B_{\rm ind}'=B_{\rm ind}+(\eta-1)B_{{\rm ind},f},\enspace
B_{{\rm ind},f}=\sum_{b=2}^{K}T_\rho^bE_f.
\label{eq:fixture-power}
\end{equation}
We implement this update by sparse re-solving with $E'$.
The source strength therefore scales exactly, and the approximation lies entirely in freezing the indirect pattern, which relies on the edit preserving the scene's occlusion structure.
It degrades for receivers near the fixture and breaks down under translation, where a pure move leaves $\eta=1$ and the frozen field simply stays behind.
Emitter re-injection (\cref{sec:injection}) addresses this failure mode.

\begin{figure*}[t]
  \centering
  \paperfigcaptionstyle
  \edef\scw{\the\dimexpr(\columnwidth-\paperfiglabelwidth-4\paperfiggap)/3\relax}%
  \edef\sch{\the\dimexpr\scw*2/3\relax}%
  \def\sclab#1{\paperfigrowlabel{\sch}{#1}}%
  \def\schdr#1{\paperfigheader{\scw}{#1}}%
  \paperfigsetup
  \begin{tabular}{c@{\hspace{\paperfiggap}}c@{\hspace{\paperfiggroupgap}}c@{\hspace{\paperfiggap}}c@{\hspace{\paperfiggroupgap}}c@{\hspace{\paperfiggap}}c}
    & \schdr{$1\times$} & \schdr{$2\times$} & \schdr{$|2{\times} - 1{\times}|$} & \schdr{$3\times$} & \schdr{$|3{\times} - 1{\times}|$} \\[\paperfighdrgap]
    \sclab{Direct} &
    \includegraphics[width=\scw]{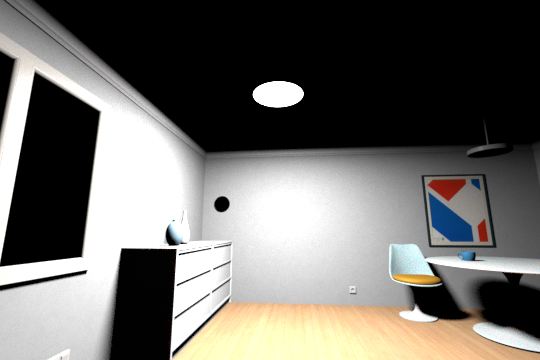} &
    \includegraphics[width=\scw]{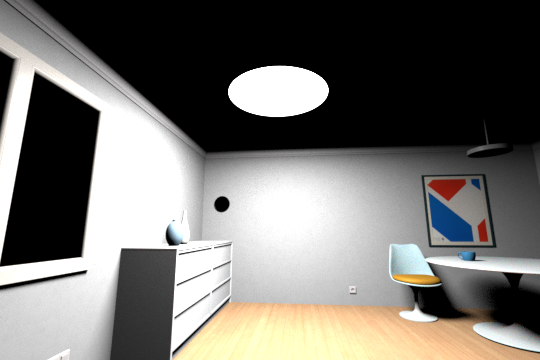} &
    \includegraphics[width=\scw]{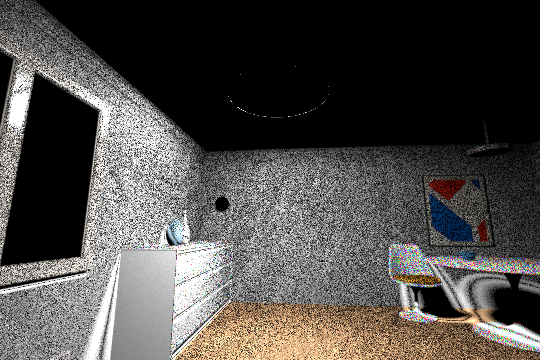} &
    \includegraphics[width=\scw]{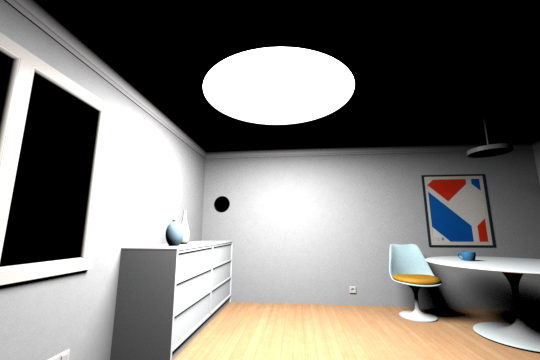} &
    \includegraphics[width=\scw]{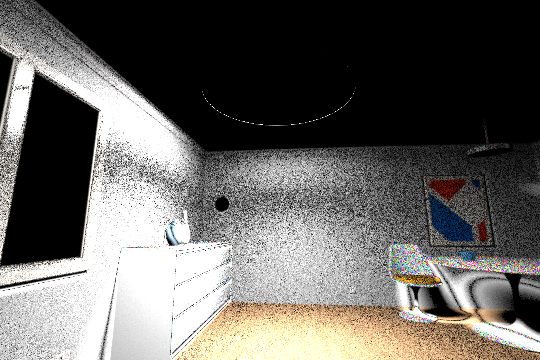} \\[\paperfiggap]
    \sclab{Indirect} &
    \includegraphics[width=\scw]{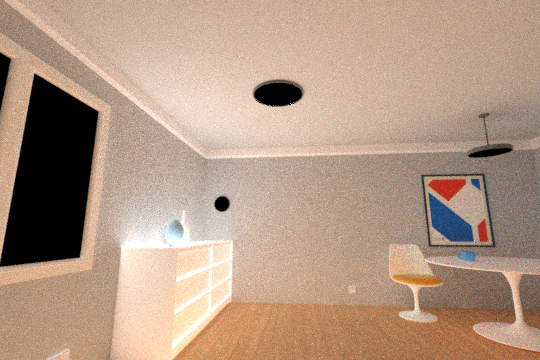} &
    \includegraphics[width=\scw]{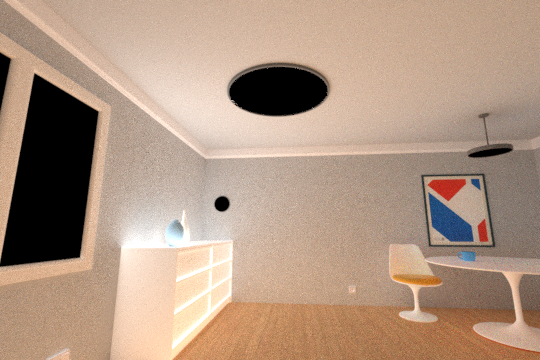} &
    \includegraphics[width=\scw]{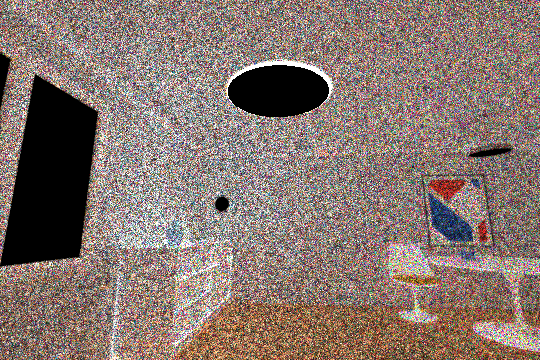} &
    \includegraphics[width=\scw]{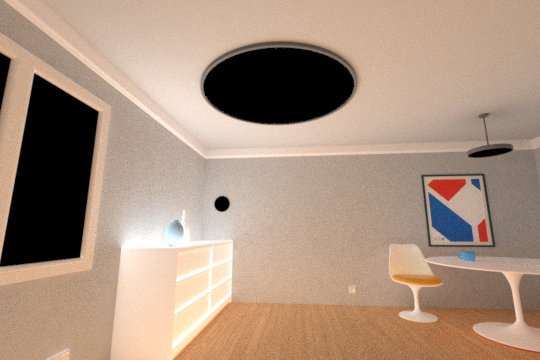} &
    \includegraphics[width=\scw]{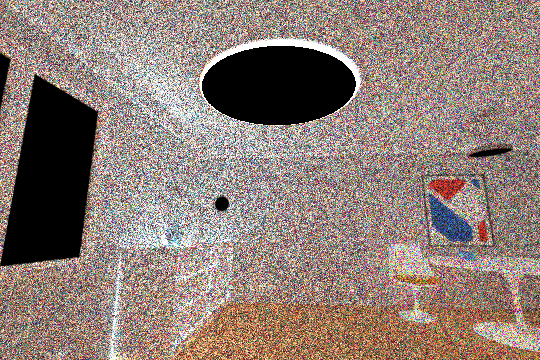} \\
  \end{tabular}
  \caption{
  Direct and indirect illumination from a single ceiling disk light in \emph{Kitchen} as it is scaled by $a$ about its center at fixed radiance, path traced at 8192\,spp without denoising. The direct row includes the visible emitter.
  Each field is divided by its mean over non-emitting pixels, so a pure change of emitted power would leave it unchanged. Next to each scaled field, its absolute difference to the $1\times$ field is shown, amplified $4\times$. 
  The direct field changes near the fixture and at shadow boundaries, whereas the displayed normalized indirect field remains approximately stable. 
  The ring in the indirect differences is the enlarged fixture body.
  }
  \label{fig:scaling_analysis}
\end{figure*}

\paragraph{Ground-truth shape analysis.}
The $a^2$ source-strength factor is exact, so the approximation depends on whether the spatial shape of the indirect field is preserved.
We test this independently of our reconstruction by path tracing direct and indirect illumination from a single ceiling disk at scales $a\in\{1,2,3\}$ (8192\,spp without denoising, Fig.~\ref{fig:scaling_analysis}).
Dividing each field by its mean over non-emitting pixels removes the overall intensity, so the comparison sees only the spatial distributions.
Under this normalization the direct field visibly changes near the fixture and at shadow boundaries, while the indirect field stays approximately stable across the displayed scales.
The ring in the indirect differences is the enlarged fixture body itself.

\paragraph{Comparison with emitter re-injection.}
We next evaluate the practical effect of the frozen-pattern approximation by scaling the \emph{Kitchen} ceiling disk about its center by $a\in[0.7,4]$ under Relit~2 at fixed source radiance.
Tab.~\ref{tab:fixture_edits} compares the resulting composites against path-traced references of each edited scene.
Both variants update direct illumination on the transformed Gaussians; they differ in how they update indirect light.
Across the in-place scalings, their composite PSNR differs by at most 0.06\,dB, supporting the simpler update as an approximation to re-injection in these evaluated cases.
Both variants become less accurate at larger scales, so their close agreement alone does not establish that the indirect pattern remains unchanged.
When scaling is combined with translations of 10--20\,cm, re-injection improves PSNR by about 0.2--0.5\,dB.
Translation changes the spatial distribution of the fixture's illumination, which rescaling its original indirect field cannot capture, and the gap grows with pendant translation distance.

\begin{table}[t]
  \centering
  \small
  \renewcommand{\arraystretch}{0.95}%
  \edef\padTsep{\the\dimexpr\belowrulesep/2\relax}%
  \edef\padBsep{\the\dimexpr\aboverulesep/2\relax}%
  \setlength{\aboverulesep}{0pt}\setlength{\belowrulesep}{0pt}%
  \def\padT{\rule{0pt}{\dimexpr\arraystretch\ht\strutbox+\padTsep\relax}}%
  \def\padB{\rule[-\dimexpr\arraystretch\dp\strutbox+\padBsep\relax]{0pt}{0pt}}%
  \begin{tabular}{@{}lccccc@{}}
    \toprule
    \padT & \multicolumn{2}{c}{Frozen pattern} & \multicolumn{2}{c}{Re-injection} & \\
    \cmidrule(lr){2-3}\cmidrule(lr){4-5}
    \padB Edit & PSNR$\uparrow$ & SSIM$\uparrow$ & PSNR$\uparrow$ & SSIM$\uparrow$ & $\Delta$PSNR \\
    \midrule
    \padT Unedited        & 36.99 & 0.983 & -- & -- & -- \\
    \multicolumn{6}{@{}l}{\textit{Ceiling disk: in-place scaling}} \\
    $0.7\times$           & 38.04 & 0.985 & 38.03 & 0.985 & $-0.02$ \\
    $1.5\times$           & 34.99 & 0.979 & 34.98 & 0.979 & $-0.01$ \\
    $2\times$             & 33.52 & 0.974 & 33.50 & 0.974 & $-0.02$ \\
    $3\times$             & 31.01 & 0.965 & 30.95 & 0.965 & $-0.06$ \\
    $4\times$             & 29.02 & 0.958 & 28.98 & 0.958 & $-0.03$ \\
    \multicolumn{6}{@{}l}{\textit{Ceiling disk: scaling and translation}} \\
    $1.5\times$, 10\,cm   & 34.83 & 0.979 & 34.98 & 0.979 & $+0.16$ \\
    $2\times$, 10\,cm     & 33.21 & 0.974 & 33.39 & 0.974 & $+0.18$ \\
    \padB $2\times$, 20\,cm     & 32.92 & 0.974 & 33.37 & 0.974 & $+0.45$ \\
    \midrule
    \multicolumn{6}{@{}l}{\padT
      \textit{Dining pendant: translation}} \\
    Unedited & 36.96 & 0.984 & --    & --     & --      \\
    0.5\,m   & 36.21 & 0.983 & 36.70 & 0.983 & $+0.49$ \\
    1\,m     & 35.08 & 0.982 & 36.73 & 0.983 & $+1.64$ \\
    \padB 2\,m     & 32.73 & 0.976 & 36.35 & 0.982 & $+3.62$ \\
    \bottomrule
  \end{tabular}
  \caption{
  Fixture geometry edits in \emph{Kitchen}, evaluated against a path-traced reference of each edited scene: ceiling-disk scaling and translation under Relit~2 and dining-pendant translation under Relit~3 (Fig.~\ref{fig:fixture_edits}).
  Both variants update direct illumination on the transformed Gaussians.
  The frozen-pattern update scales the fixture's original indirect field by $a^2$ ($a=1$ for pure translation), whereas re-injection recomputes its first-reflection field.
  $\Delta$ is re-injection minus frozen-pattern PSNR.
  For in-place scaling the two agree within 0.06\,dB, and once the fixture moves, re-injection gains with distance.
  Both decline as the disk grows, while $0.7\times$ scores above the unedited scene; %
  we attribute both to the inflated emitter silhouette of the reconstruction, which scales with the disk.
  The unedited rows repeat the production composites of Tab.~\ref{tab:relighting} for their conditions.
  }
  \label{tab:fixture_edits}
\end{table}

\section{Synthetic Benchmark Scenes}
\label{sec:supp-scenes}
To our knowledge, existing datasets for quantitative relighting evaluation are object-centric or, at room scale, lack complex light fixtures.
We therefore built \emph{Kitchen} and \emph{Living room} in Blender to evaluate emitter-driven relighting of entire interiors against path-traced references.
For every relit condition, they provide a single-bounce reference in addition to the full global-illumination reference, so that the direct and indirect parts of our hybrid method can be evaluated separately (Tab.~\ref{tab:transport_components}).
Both are closed, furnished rooms of about $4.75{\times}4.75$\,m with a 2.25\,m ceiling (2.2M and 0.9M triangles).
All light originates from emissive surfaces visible in the camera images, with no environment illumination and no point or area light primitives.
Each scene contains eight fixtures (Tab.~\ref{tab:supp-scenes}).
\emph{Kitchen} combines a window, three flush ceiling disks, a wall disk, and an under-cabinet strip with two free-hanging pendants.
\emph{Living room} combines two windows and a pendant with two aimed ceiling spots and a floor, a table, and a desk lamp.

\paragraph{Views and reference passes.}
Each scene provides 450 posed views at $540{\times}360$ pixels with a $90^\circ$ horizontal field of view.
Camera positions are sampled uniformly over the room, and a subset of the poses is steered toward the fixtures so that every emitting surface is observed.
Every eighth view is held out, giving 393 training and 57 test views.
All images are path traced in Blender Cycles with adaptive sampling of up to 1024\,spp (noise threshold 0.01) and OpenImageDenoise, and are stored as linear radiance without tone mapping.
The capture condition provides radiance, diffuse albedo, depth, geometric normals, and binary emitter masks for all views.
Each relit condition provides test-view radiance with full global illumination, a matching single-bounce reference that contains only emission and direct light, and emitter masks for the training views.

\paragraph{Lighting conditions.}
Tab.~\ref{tab:supp-scenes} lists the emitted radiance of every fixture.
All capture emitters are neutral white, and three \emph{Living room} lamps do not emit at capture.
The relit conditions switch on these lamps, switch off the windows that dominate the capture, dim or boost individual emitters, and tint emitters.

\paragraph{Edit references.}
For \emph{Kitchen}, we also path trace edited scenes on the test views:
one wall recolored red under the capture condition and under Relit~3 (\cref{sec:exp-albedo}), eight scaling and translation edits of ceiling disk~1 under Relit~2, and three pendant translations under Relit~3 (Tab.~\ref{tab:fixture_edits}).

\newcommand{\scchip}[1]{{\setlength{\fboxsep}{0pt}\setlength{\fboxrule}{0.3pt}%
  \fcolorbox[rgb]{0.45,0.45,0.45}{#1}{\rule{0pt}{0.52em}\rule{0.52em}{0pt}}}}
\newcommand{\scT}[2]{#1\,\scchip{#2}}                    %
\newcommand{\scW}[1]{#1\,\hphantom{\scchip{1,1,1}}}      %

\begin{table}[t]
  \centering
  \small
  \renewcommand{\arraystretch}{0.95}%
  \edef\padTsep{\the\dimexpr\belowrulesep/2\relax}%
  \edef\padBsep{\the\dimexpr\aboverulesep/2\relax}%
  \setlength{\aboverulesep}{0pt}\setlength{\belowrulesep}{0pt}%
  \def\padT{\rule{0pt}{\dimexpr\arraystretch\ht\strutbox+\padTsep\relax}}%
  \def\padB{\rule[-\dimexpr\arraystretch\dp\strutbox+\padBsep\relax]{0pt}{0pt}}%
  \begin{tabular}{@{}lccccc@{}}
    \toprule
    \padT\padB Fixture & Area & Capture & \multicolumn{1}{c}{Relit~1} & \multicolumn{1}{c}{Relit~2} & \multicolumn{1}{c@{}}{Relit~3} \\
    \midrule
    \multicolumn{6}{@{}l}{\padT\emph{Kitchen}} \\
    Window                     & 1.19  & 10       & \scW{0}  & \scW{0}    & \scT{6}{1,0.65,0.35} \\   %
    Ceiling disk 1              & 0.163 & 4        & \scW{0}  & \scW{15}   & \scW{2} \\
    Ceiling disks 2, 3   & 0.163 & 4        & \scW{0}  & \scT{4}{1,0.77,0.48} & \scW{2} \\     %
    Dining pendant             & 0.113 & 15       & \scT{16}{1,0.87,0.68} & \scW{0.05} & \scW{14} \\ %
    Linear pendant             & 0.056 & 0.2      & \scW{15} & \scW{0.1}  & \scW{0.1} \\
    Wall disk                  & 0.042 & 0.2      & \scT{10}{0.63,1,0.70} & \scT{10}{0.77,0.54,1} & \scW{0.1} \\ %
    Cabinet strip              & 0.021 & 6        & \scT{4}{1,0.92,0.80} & \scT{2}{1,0.92,0.80} & \scW{0.05} \\ %
    \padB Emitted flux (\%)   &       & 100 & \scW{17} & \scW{22} & \scW{39} \\
    \midrule
    \multicolumn{6}{@{}l}{\padT\emph{Living room}} \\
    Window, west               & 1.19  & 8        & \scT{8}{1,0.78,0.51} & \scW{0} & \scT{0.88}{0.46,0.60,1} \\ %
    Window, south              & 0.76  & 8        & \scT{8}{1,0.78,0.51} & \scW{0} & \scT{0.88}{0.46,0.60,1} \\
    Pendant                    & 0.087 & 76       & \scW{0}  & \scW{49}   & \scT{25}{1,0.67,0.40} \\  %
    Ceiling spots, $2\times$   & 0.017 & 15       & \scW{60} & \scW{0}    & \scW{30} \\
    Floor lamp                 & 0.066 & 0        & \scW{0}  & \scT{150}{1,0.25,0.13} & \scW{0} \\   %
    Table lamp                 & 0.016 & 0        & \scW{0}  & \scT{150}{1,0.80,0.57} & \scW{0} \\   %
    Desk lamp                  & 0.013 & 0        & \scW{0}  & \scW{0}    & \scT{60}{1,0.80,0.57} \\  %
    \padB Emitted flux (\%)   &       & 100 & \scW{53} & \scW{37} & \scW{14} \\
    \bottomrule
  \end{tabular}
  \caption{
  Fixtures of our synthetic scenes with their room-facing emitting area $\mathcal A_f$ in m$^2$ (per fixture) and their emitted radiance in each lighting condition, in scene-linear units.
  Colored squares mark tinted emission, and entries without a square are neutral white.
  Emitted flux is $\pi\sum_f \mathcal A_f\,Y(\boldsymbol\ell_f)$ over all fixtures, with $Y$ the Rec.~709 luminance, relative to the capture condition.
  }
  \label{tab:supp-scenes}
\end{table}

\begin{figure*}[p]
  \centering
  \paperfigcaptionstyle
  \edef\relw{\the\dimexpr(\linewidth-3\paperfiggap-3\paperfiggroupgap-0.5pt)/7\relax}%
  \def\relhdr#1{\paperfigheader{\relw}{#1}}%
  \paperfigsetup
  \begin{tabular}{c@{\hspace{\paperfiggroupgap}}c@{\hspace{\paperfiggap}}c@{\hspace{\paperfiggroupgap}}c@{\hspace{\paperfiggap}}c@{\hspace{\paperfiggroupgap}}c@{\hspace{\paperfiggap}}c}
    \relhdr{Capture} & \relhdr{Relit~1} & \relhdr{GT} & \relhdr{Relit~2} & \relhdr{GT} & \relhdr{Relit~3} & \relhdr{GT} \\[\paperfighdrgap]
        \includegraphics[width=\relw]{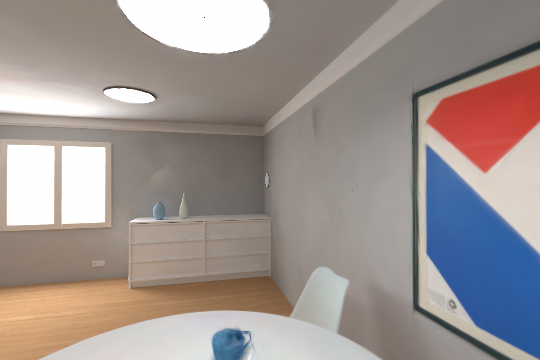} &
    \includegraphics[width=\relw]{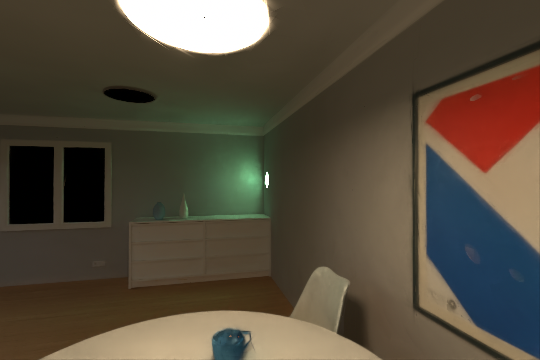} &
    \includegraphics[width=\relw]{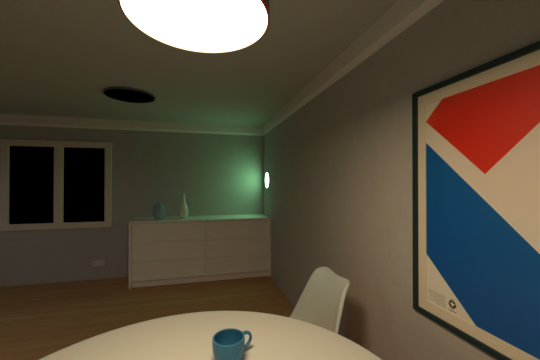} &
    \includegraphics[width=\relw]{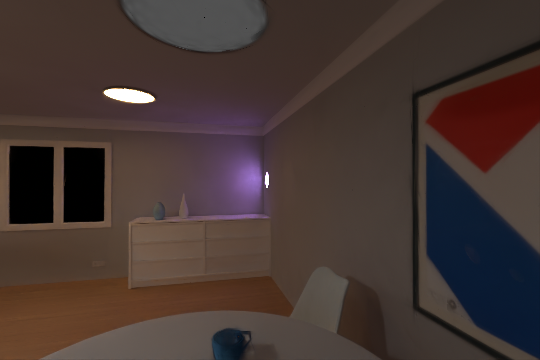} &
    \includegraphics[width=\relw]{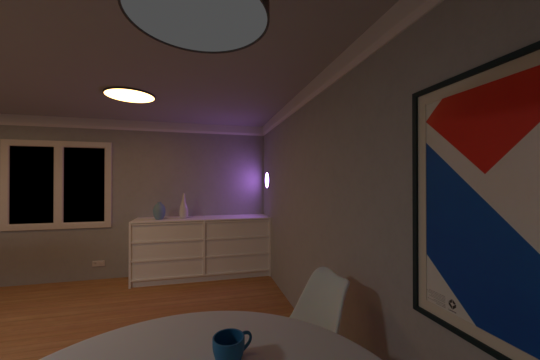} &
    \includegraphics[width=\relw]{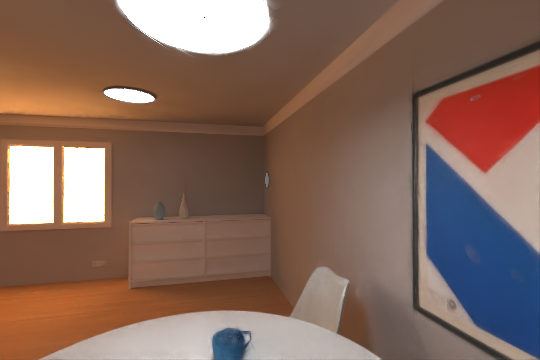} &
    \includegraphics[width=\relw]{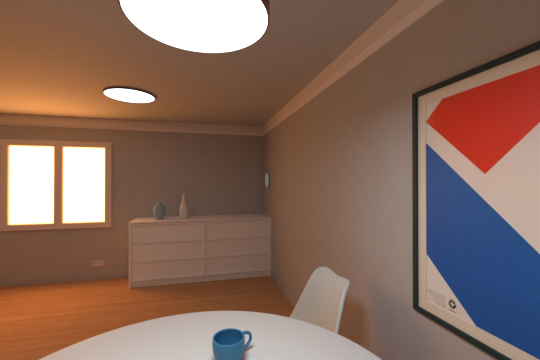} \\[\paperfiggap]
    \includegraphics[width=\relw]{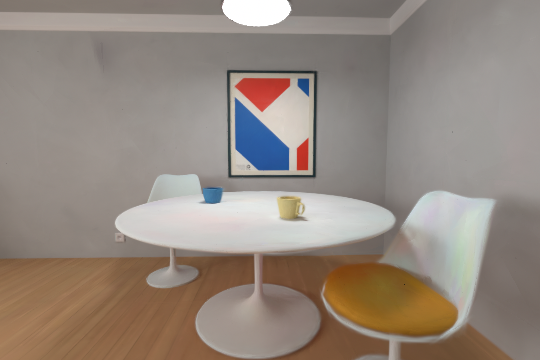} &
    \includegraphics[width=\relw]{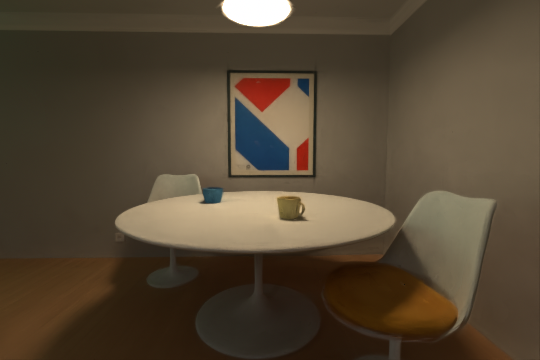} &
    \includegraphics[width=\relw]{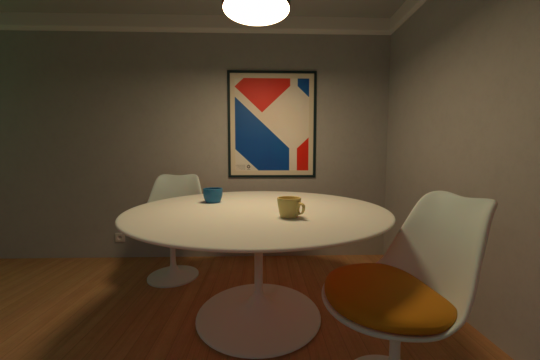} &
    \includegraphics[width=\relw]{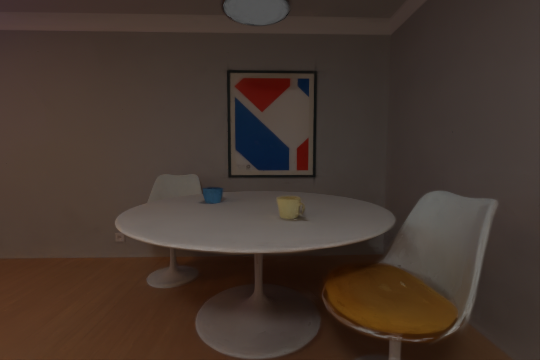} &
    \includegraphics[width=\relw]{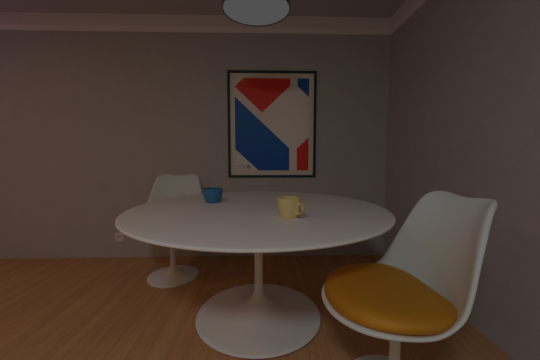} &
    \includegraphics[width=\relw]{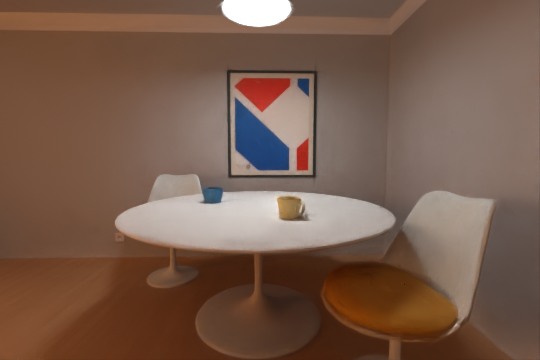} &
    \includegraphics[width=\relw]{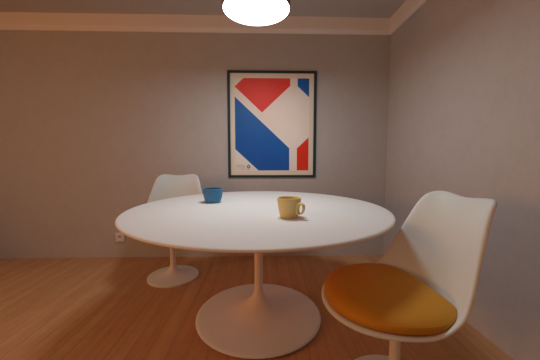} \\[\paperfiggap]
    \includegraphics[width=\relw]{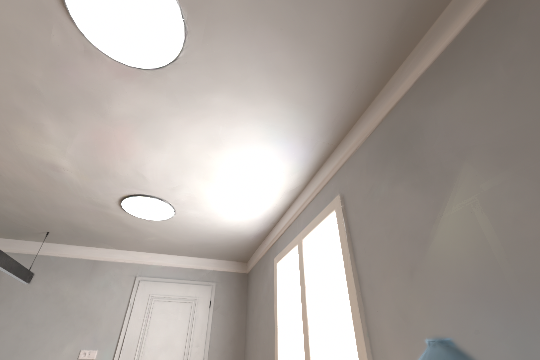} &
    \includegraphics[width=\relw]{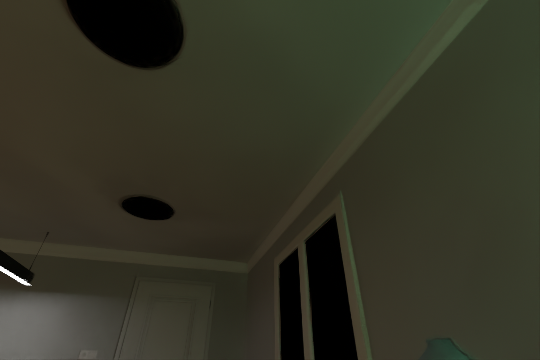} &
    \includegraphics[width=\relw]{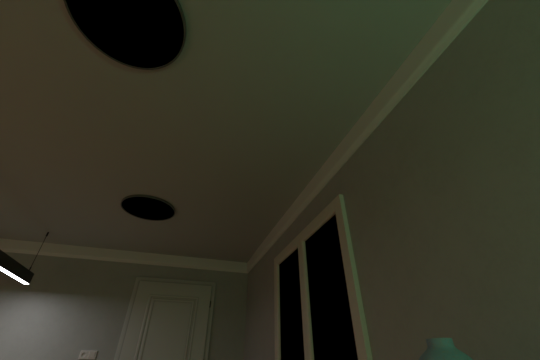} &
    \includegraphics[width=\relw]{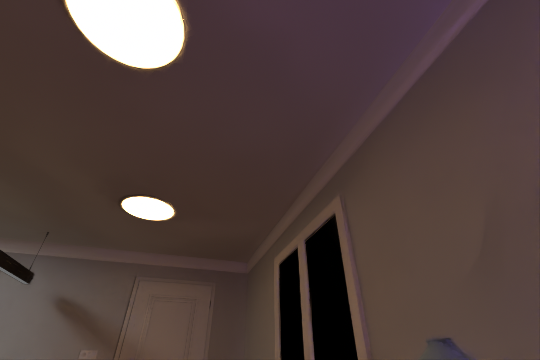} &
    \includegraphics[width=\relw]{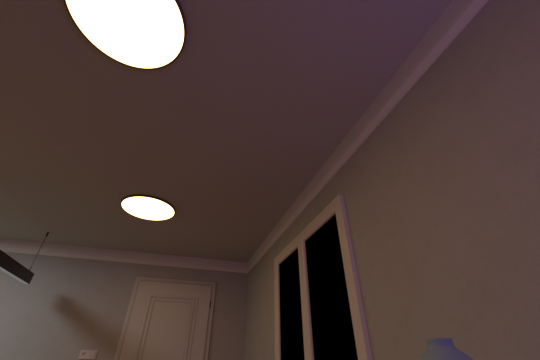} &
    \includegraphics[width=\relw]{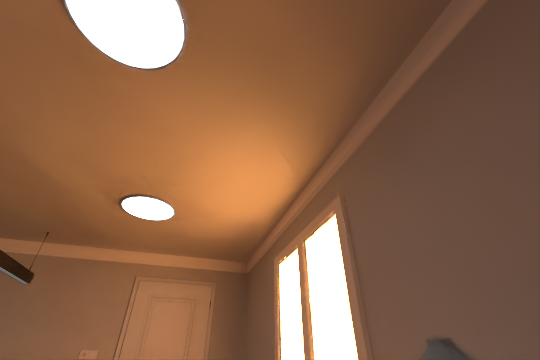} &
    \includegraphics[width=\relw]{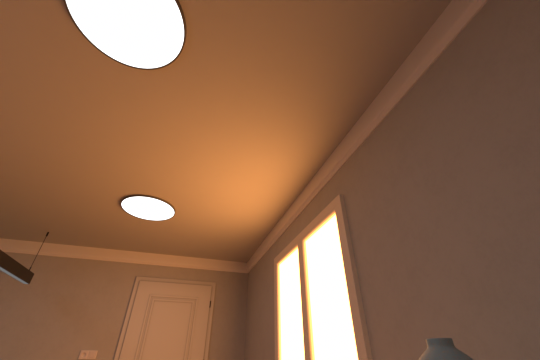} \\[\paperfiggap]
    \includegraphics[width=\relw]{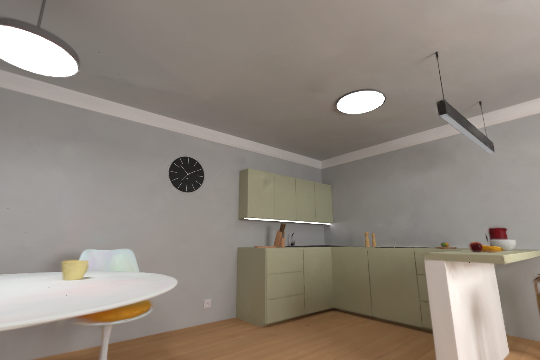} &
    \includegraphics[width=\relw]{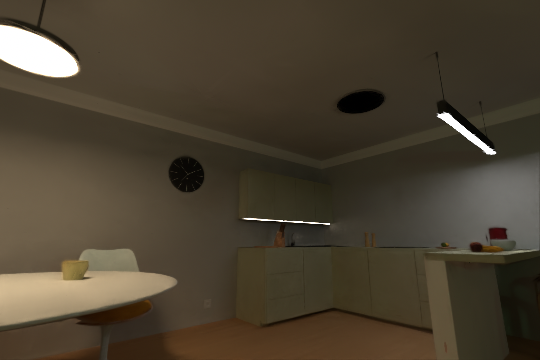} &
    \includegraphics[width=\relw]{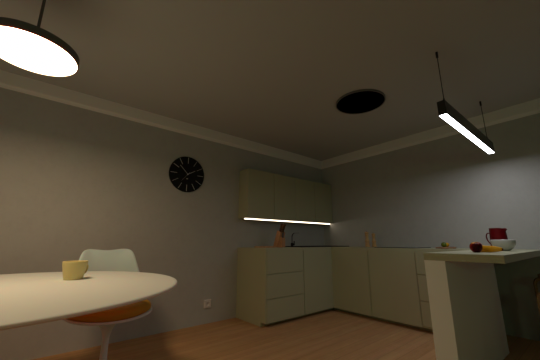} &
    \includegraphics[width=\relw]{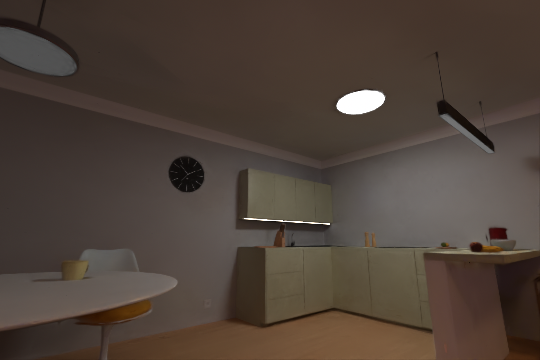} &
    \includegraphics[width=\relw]{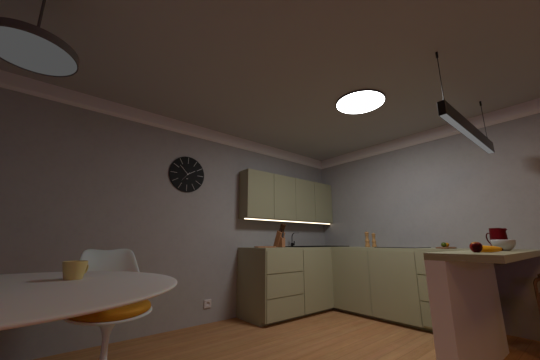} &
    \includegraphics[width=\relw]{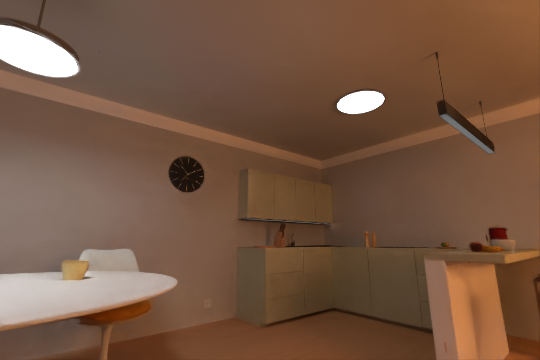} &
    \includegraphics[width=\relw]{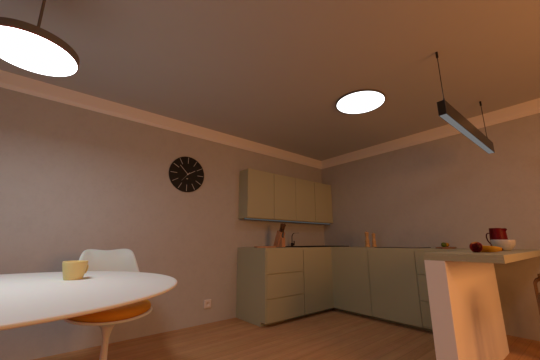} \\[\paperfiggap]
    \includegraphics[width=\relw]{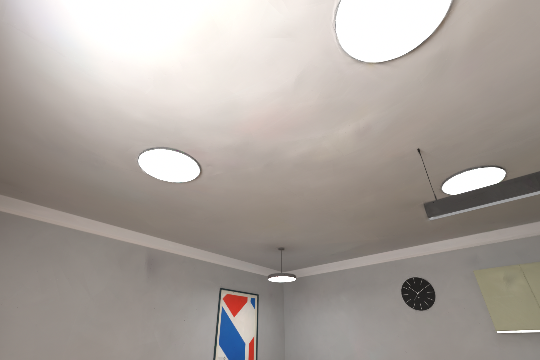} &
    \includegraphics[width=\relw]{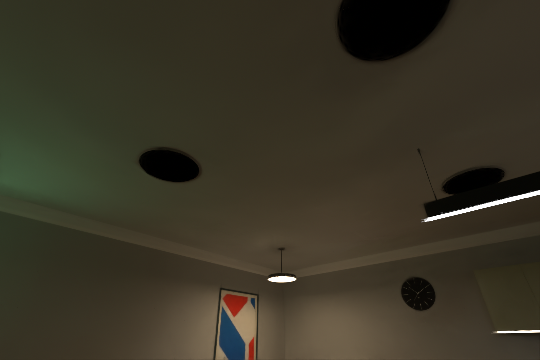} &
    \includegraphics[width=\relw]{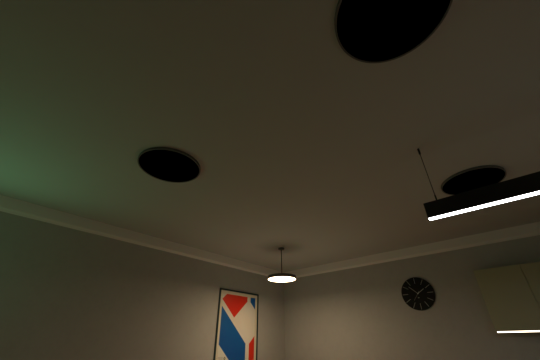} &
    \includegraphics[width=\relw]{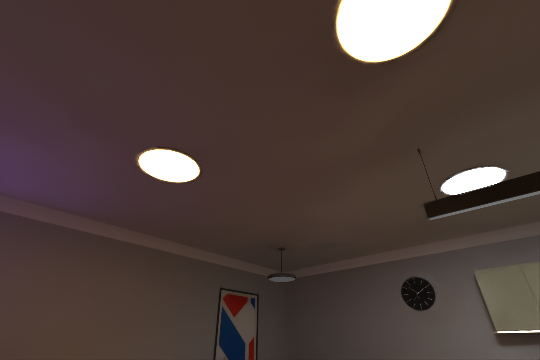} &
    \includegraphics[width=\relw]{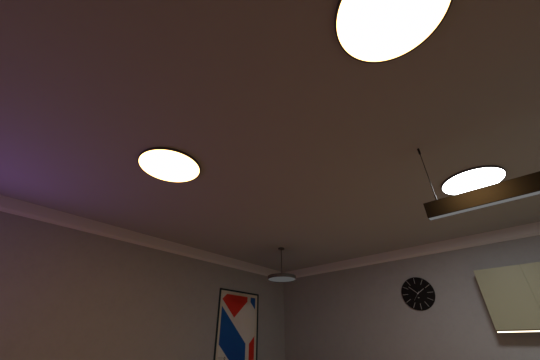} &
    \includegraphics[width=\relw]{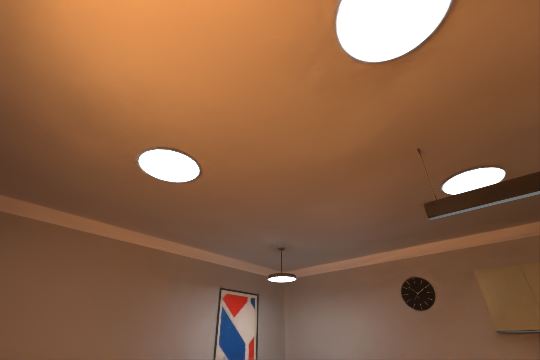} &
    \includegraphics[width=\relw]{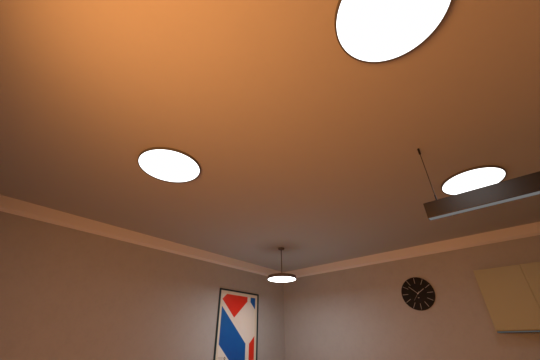} \\[\paperfiggap]
    \includegraphics[width=\relw]{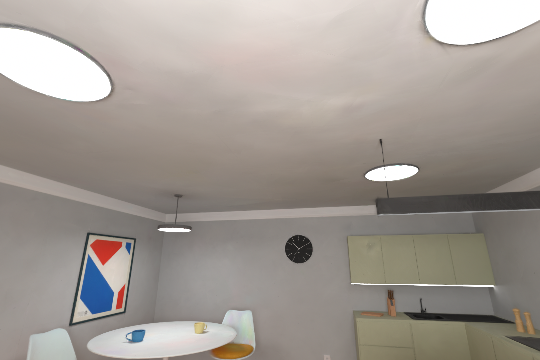} &
    \includegraphics[width=\relw]{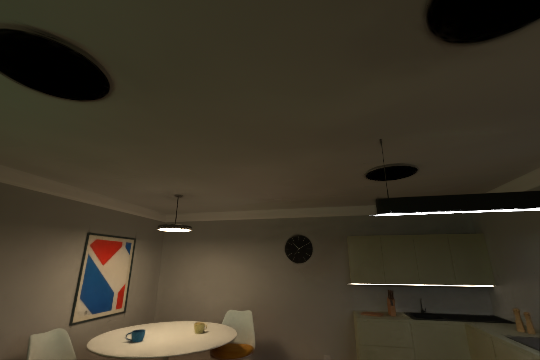} &
    \includegraphics[width=\relw]{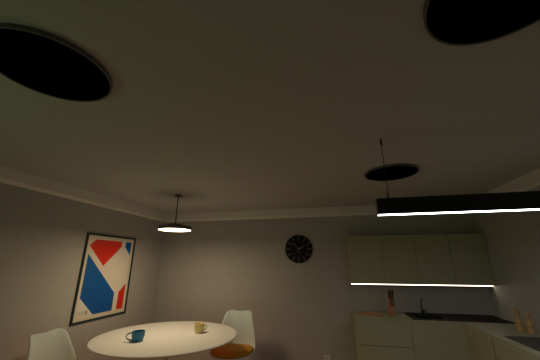} &
    \includegraphics[width=\relw]{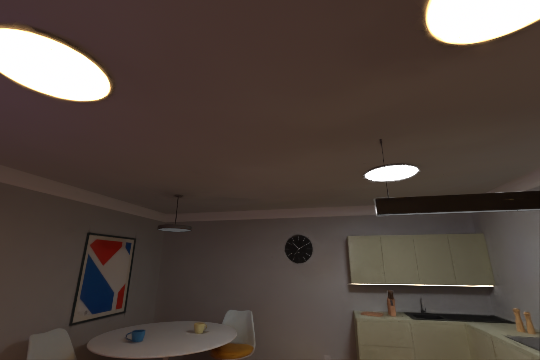} &
    \includegraphics[width=\relw]{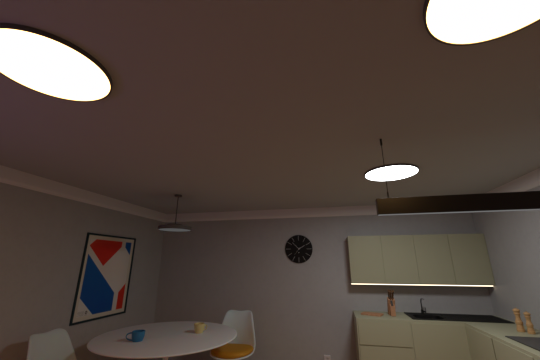} &
    \includegraphics[width=\relw]{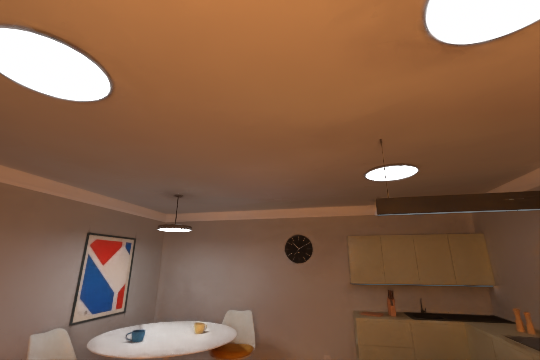} &
    \includegraphics[width=\relw]{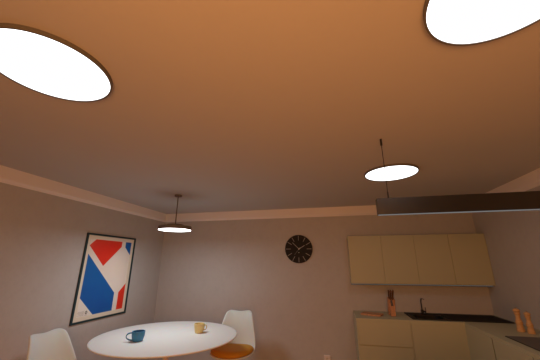} \\[\paperfiggap]
    \includegraphics[width=\relw]{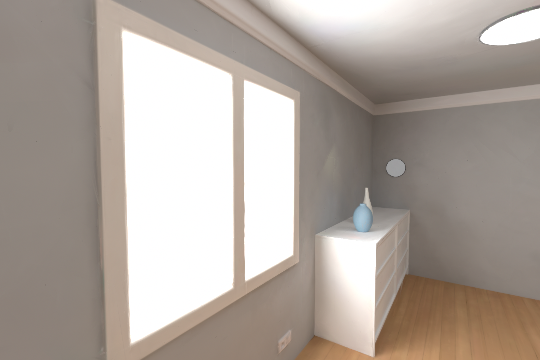} &
    \includegraphics[width=\relw]{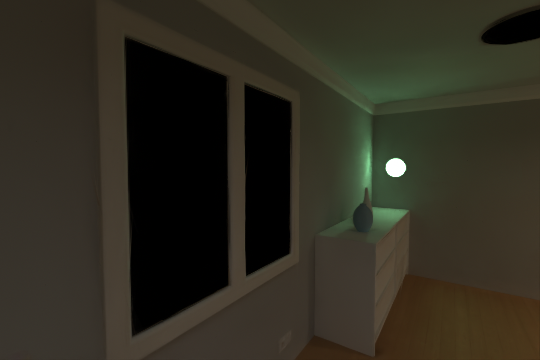} &
    \includegraphics[width=\relw]{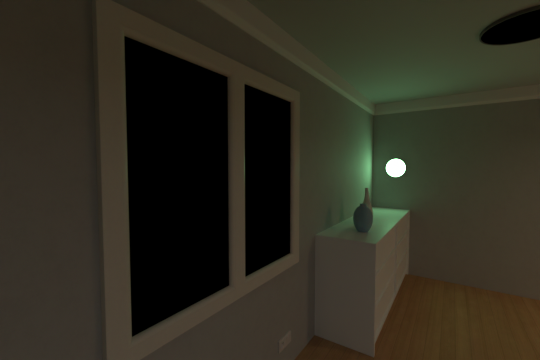} &
    \includegraphics[width=\relw]{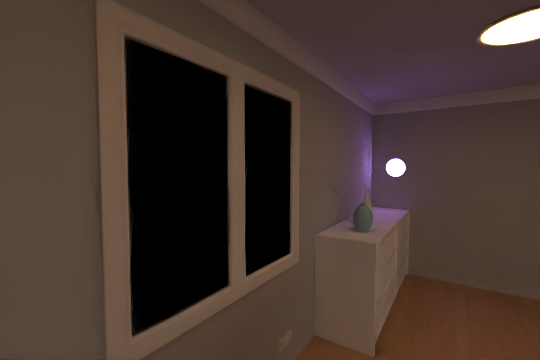} &
    \includegraphics[width=\relw]{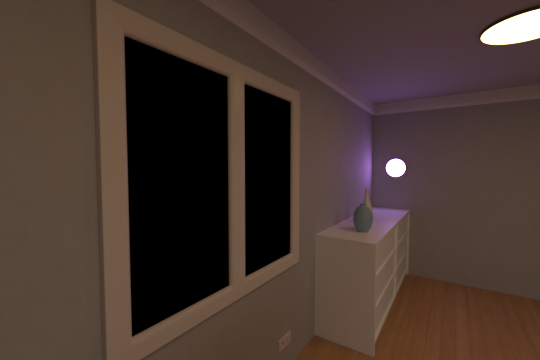} &
    \includegraphics[width=\relw]{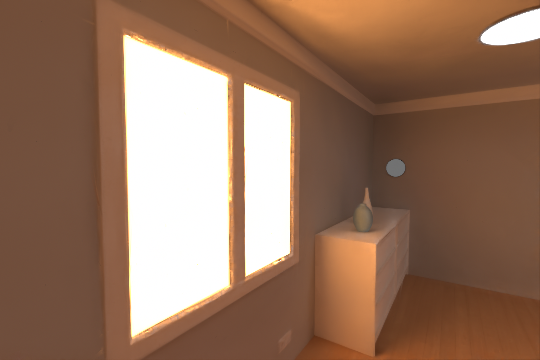} &
    \includegraphics[width=\relw]{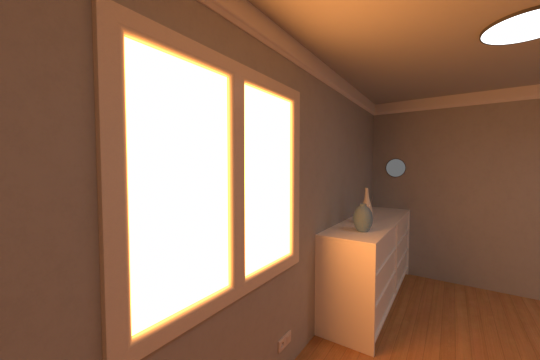} \\[\paperfiggap]
    \includegraphics[width=\relw]{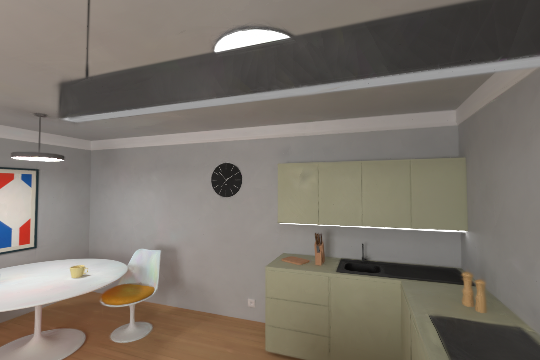} &
    \includegraphics[width=\relw]{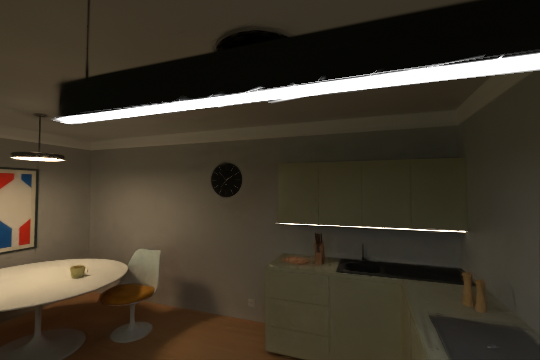} &
    \includegraphics[width=\relw]{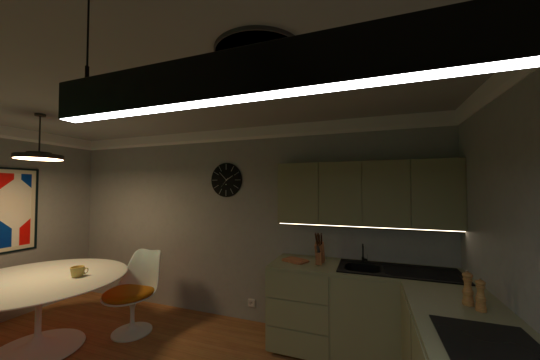} &
    \includegraphics[width=\relw]{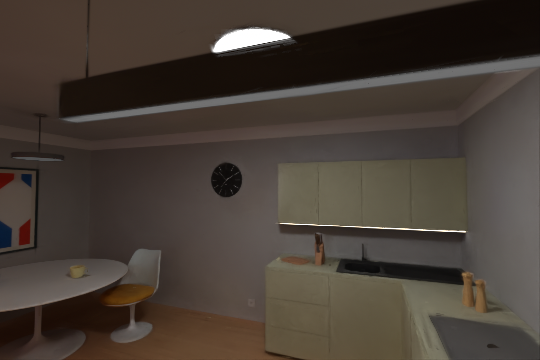} &
    \includegraphics[width=\relw]{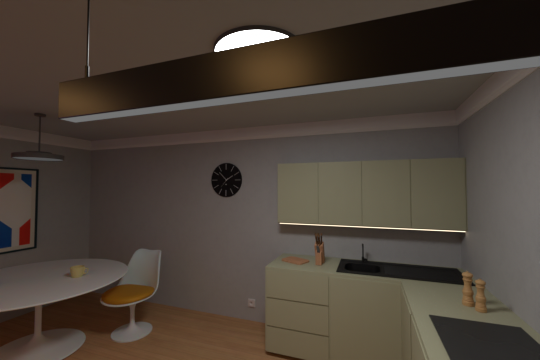} &
    \includegraphics[width=\relw]{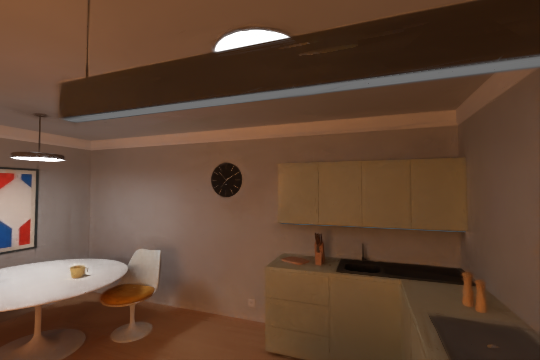} &
    \includegraphics[width=\relw]{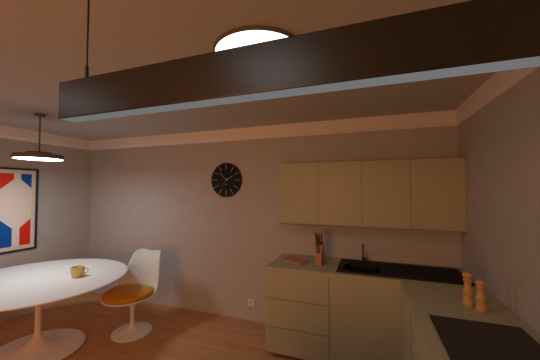} \\[\paperfiggap]
    \includegraphics[width=\relw]{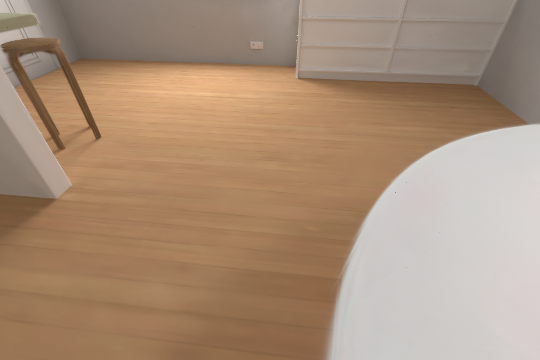} &
    \includegraphics[width=\relw]{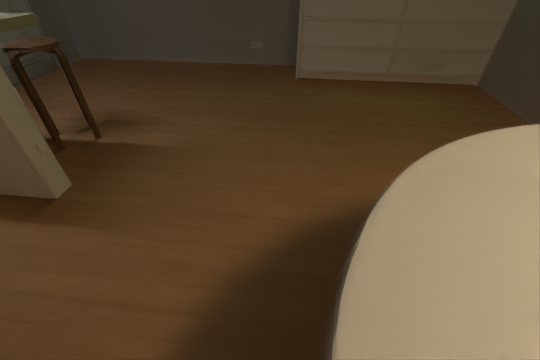} &
    \includegraphics[width=\relw]{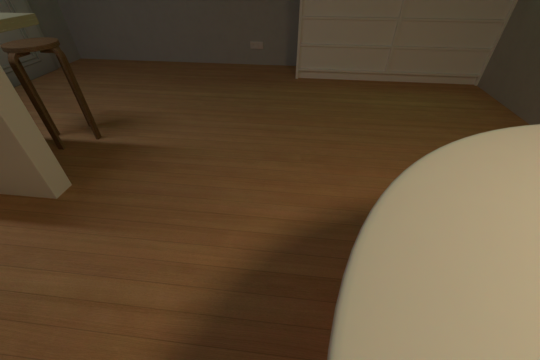} &
    \includegraphics[width=\relw]{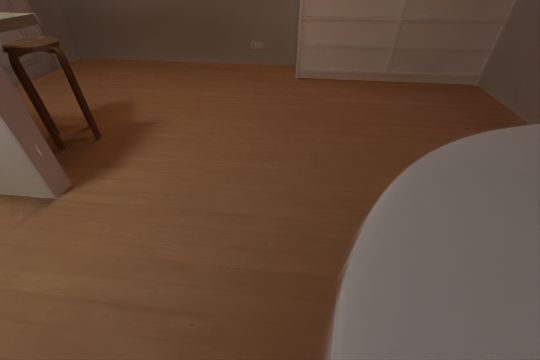} &
    \includegraphics[width=\relw]{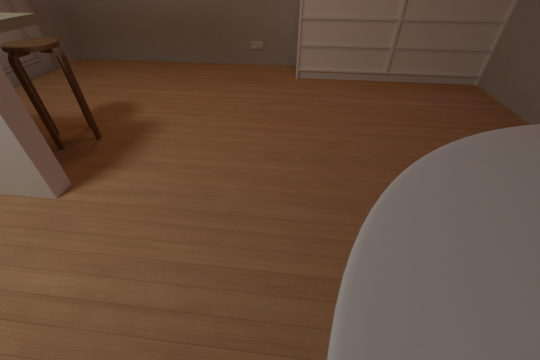} &
    \includegraphics[width=\relw]{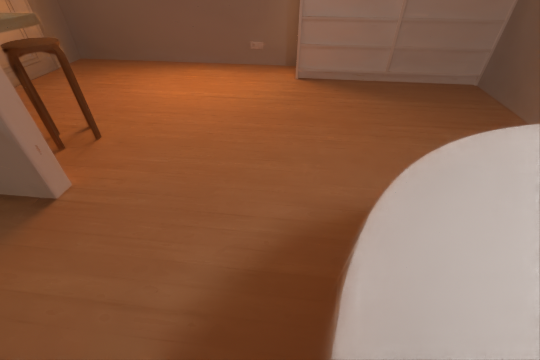} &
    \includegraphics[width=\relw]{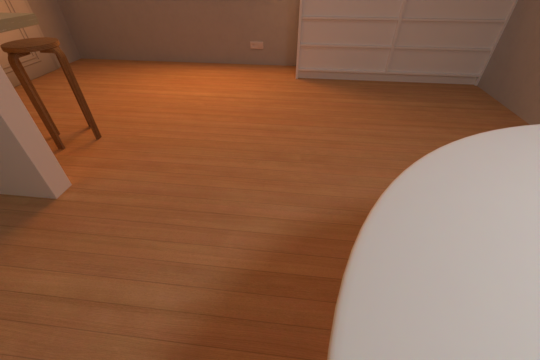} \\[\paperfiggap]
    \includegraphics[width=\relw]{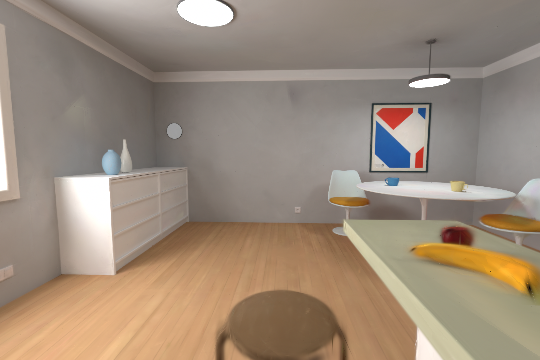} &
    \includegraphics[width=\relw]{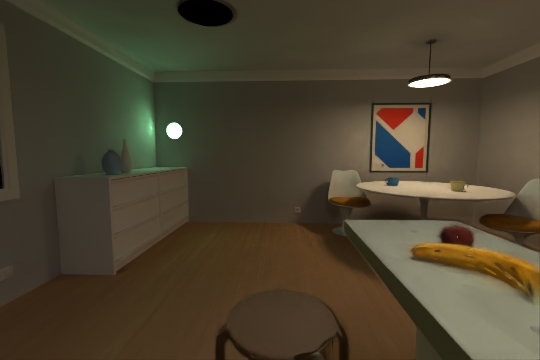} &
    \includegraphics[width=\relw]{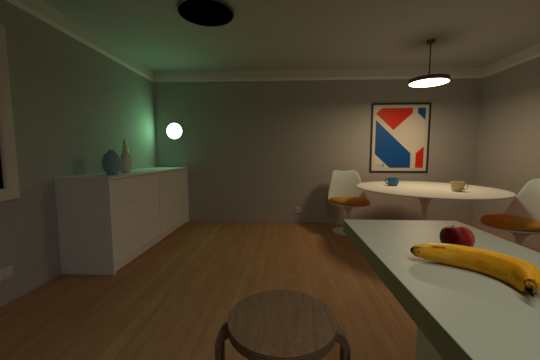} &
    \includegraphics[width=\relw]{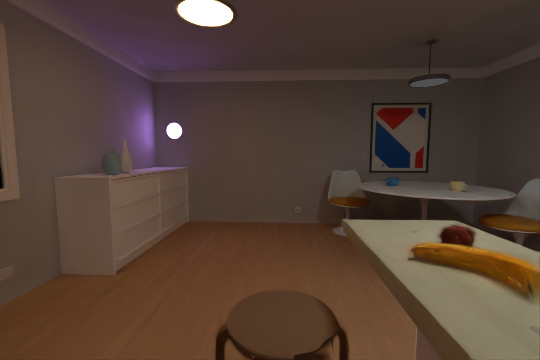} &
    \includegraphics[width=\relw]{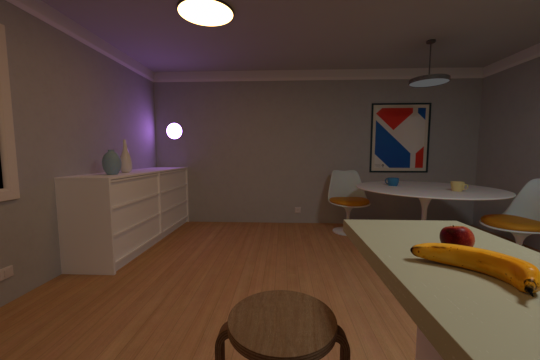} &
    \includegraphics[width=\relw]{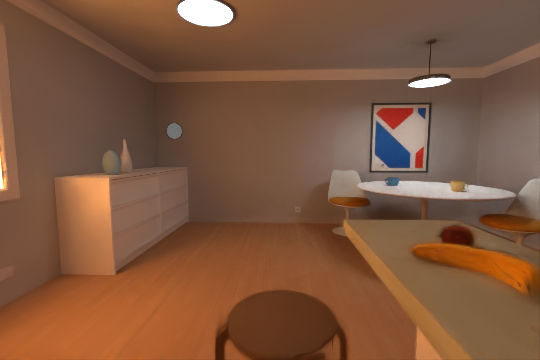} &
    \includegraphics[width=\relw]{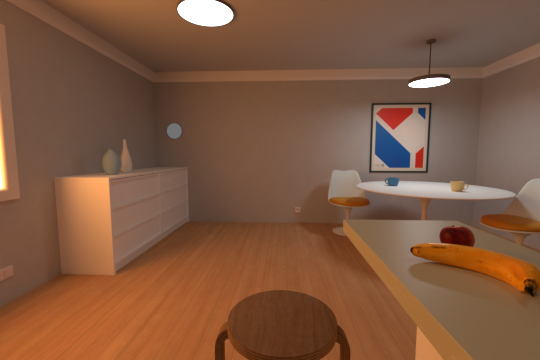} \\[\paperfiggap]
    \includegraphics[width=\relw]{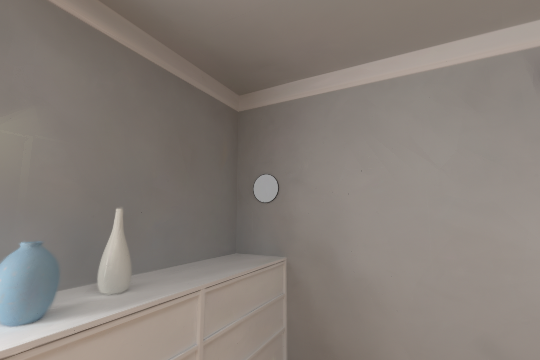} &
    \includegraphics[width=\relw]{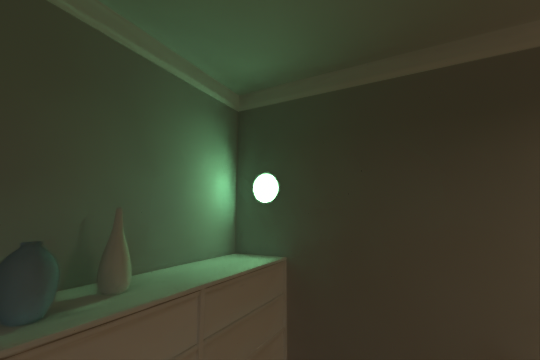} &
    \includegraphics[width=\relw]{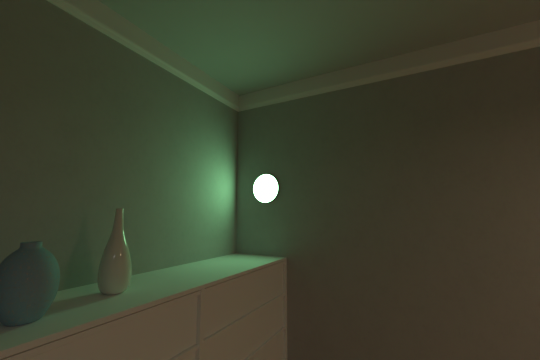} &
    \includegraphics[width=\relw]{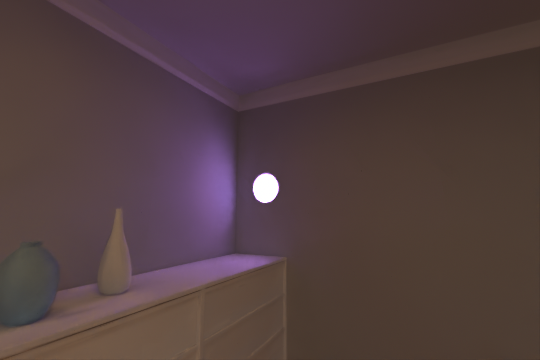} &
    \includegraphics[width=\relw]{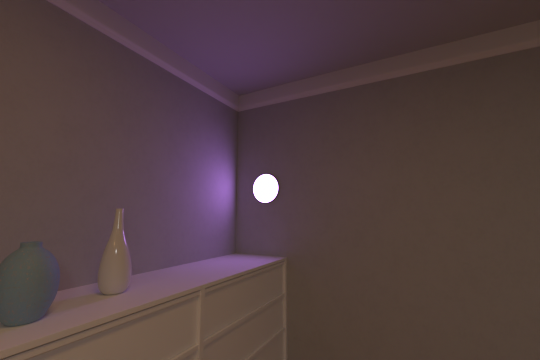} &
    \includegraphics[width=\relw]{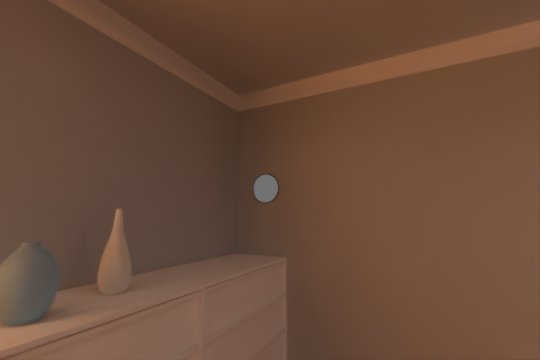} &
    \includegraphics[width=\relw]{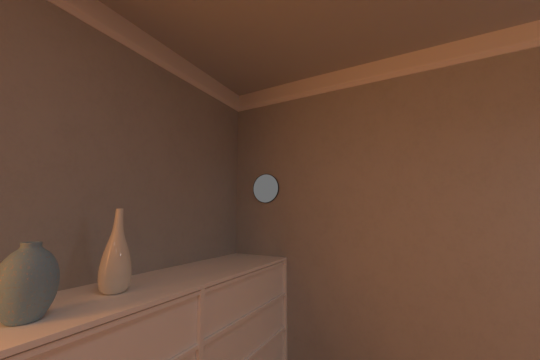} \\
  \end{tabular}
  \caption{\emph{Kitchen} relighting for different test views, one per row: our reconstruction under capture lighting (Capture) and our relighting to the three \emph{Kitchen} conditions of Tab.~\ref{tab:relighting}, which change emission strength and color and switch lights off, each next to its path-traced ground truth (GT).}
  \label{fig:relighting_kitchen_all}
\end{figure*}

\begin{figure*}[p]
  \centering
  \paperfigcaptionstyle
  \edef\relw{\the\dimexpr(\linewidth-3\paperfiggap-3\paperfiggroupgap-0.5pt)/7\relax}%
  \def\relhdr#1{\paperfigheader{\relw}{#1}}%
  \paperfigsetup
  \begin{tabular}{c@{\hspace{\paperfiggroupgap}}c@{\hspace{\paperfiggap}}c@{\hspace{\paperfiggroupgap}}c@{\hspace{\paperfiggap}}c@{\hspace{\paperfiggroupgap}}c@{\hspace{\paperfiggap}}c}
    \relhdr{Capture} & \relhdr{Relit~1} & \relhdr{GT} & \relhdr{Relit~2} & \relhdr{GT} & \relhdr{Relit~3} & \relhdr{GT} \\[\paperfighdrgap]
    \includegraphics[width=\relw]{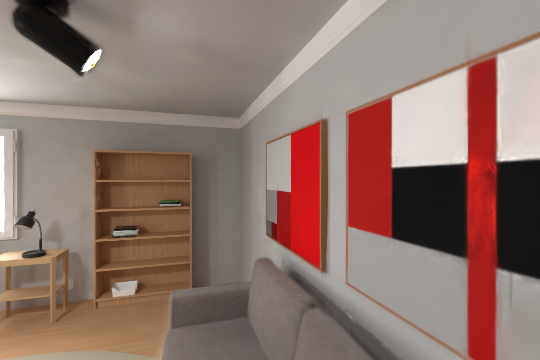} &
    \includegraphics[width=\relw]{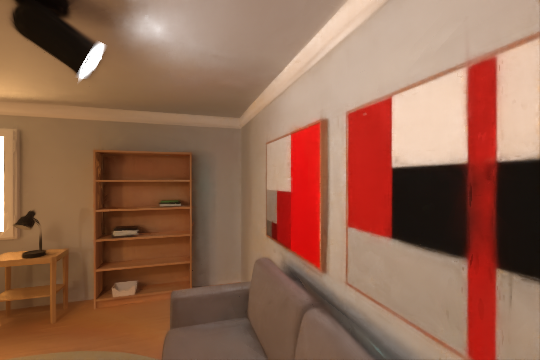} &
    \includegraphics[width=\relw]{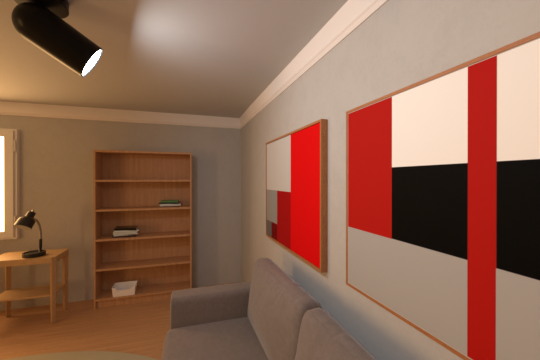} &
    \includegraphics[width=\relw]{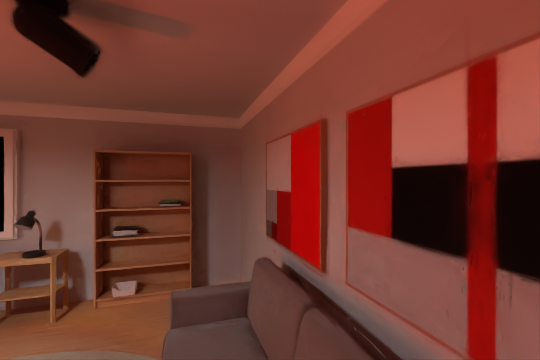} &
    \includegraphics[width=\relw]{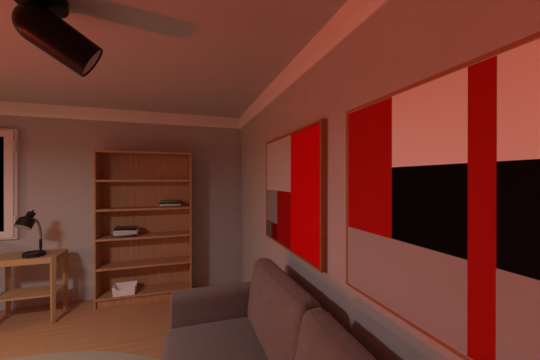} &
    \includegraphics[width=\relw]{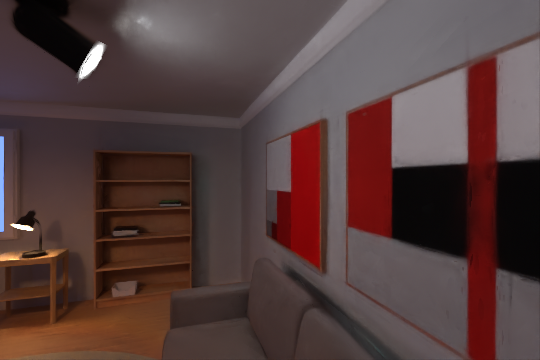} &
    \includegraphics[width=\relw]{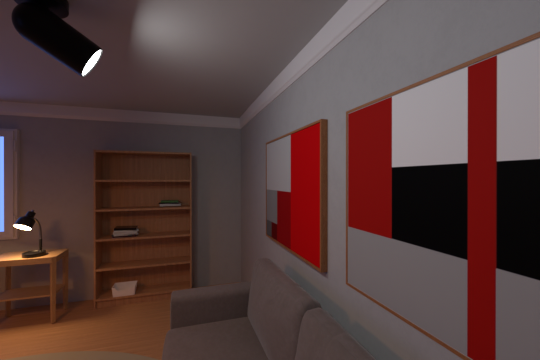} \\[\paperfiggap]
    \includegraphics[width=\relw]{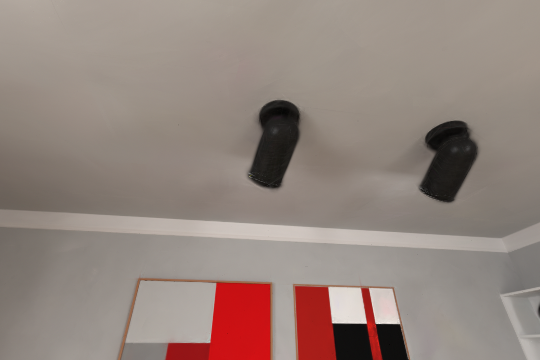} &
    \includegraphics[width=\relw]{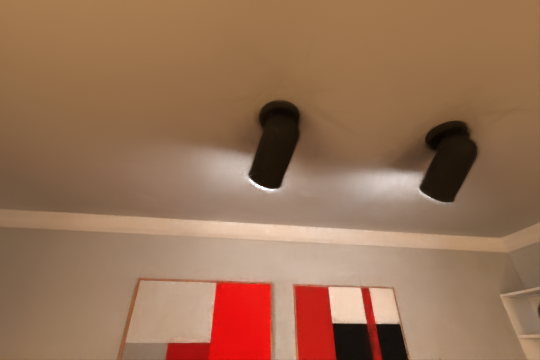} &
    \includegraphics[width=\relw]{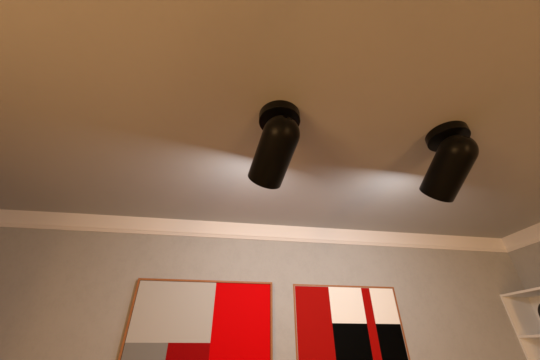} &
    \includegraphics[width=\relw]{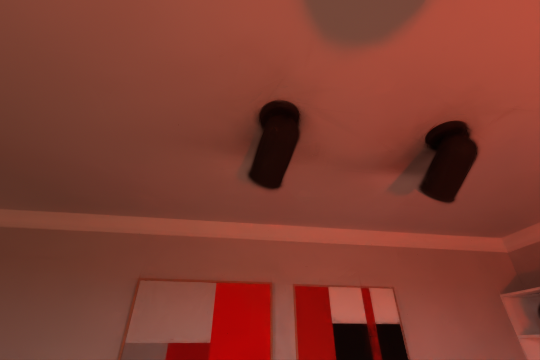} &
    \includegraphics[width=\relw]{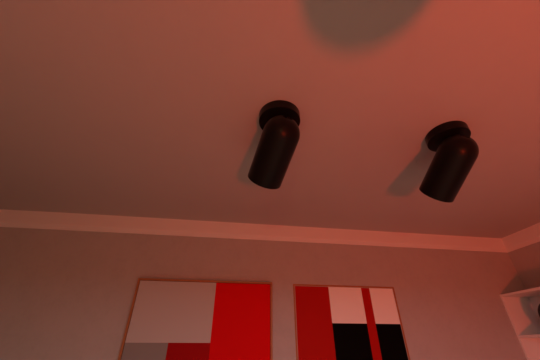} &
    \includegraphics[width=\relw]{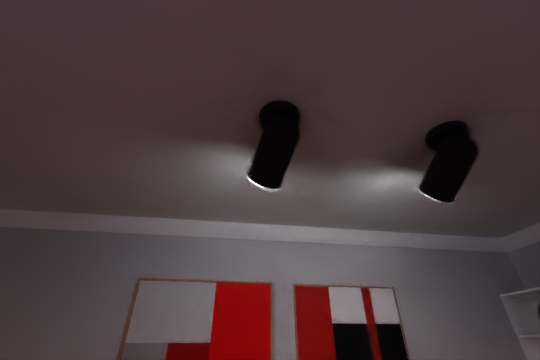} &
    \includegraphics[width=\relw]{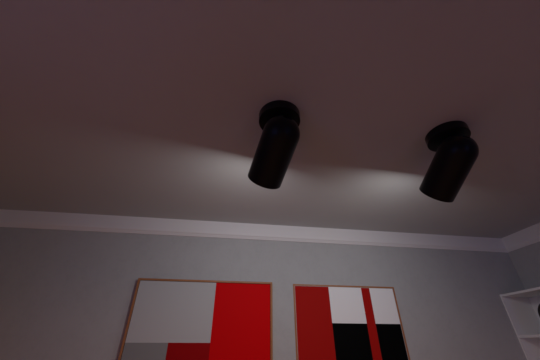} \\[\paperfiggap]
    \includegraphics[width=\relw]{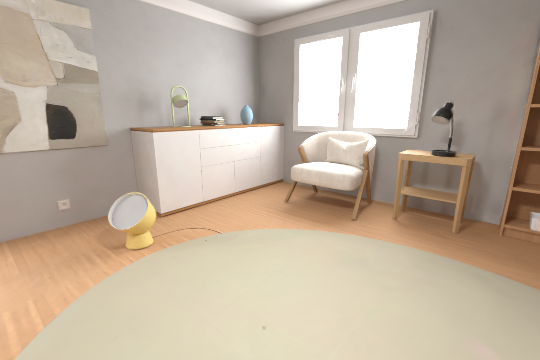} &
    \includegraphics[width=\relw]{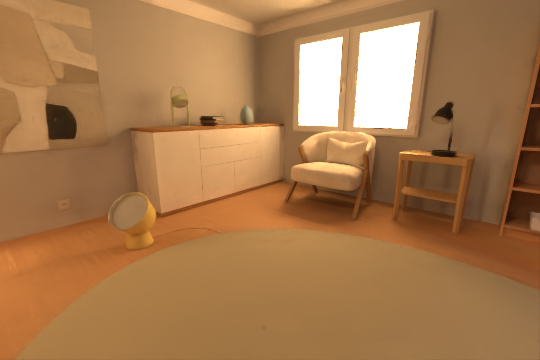} &
    \includegraphics[width=\relw]{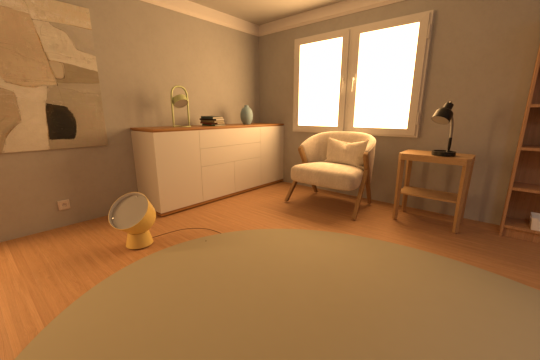} &
    \includegraphics[width=\relw]{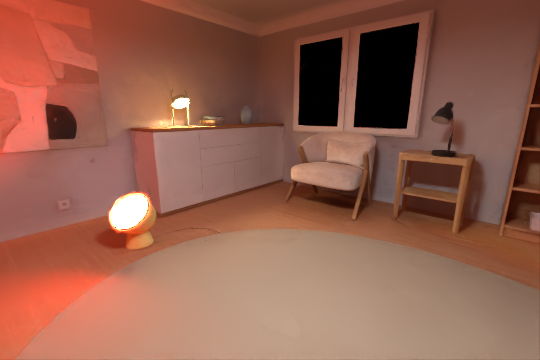} &
    \includegraphics[width=\relw]{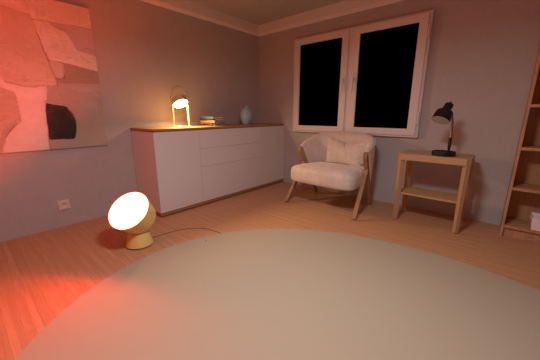} &
    \includegraphics[width=\relw]{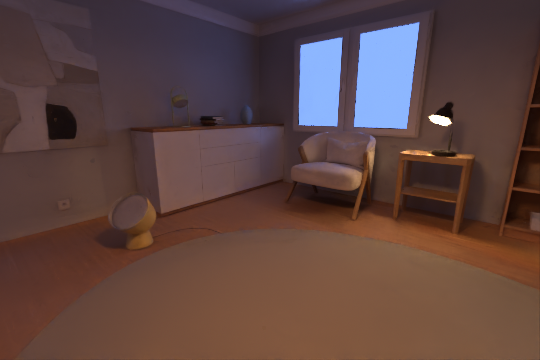} &
    \includegraphics[width=\relw]{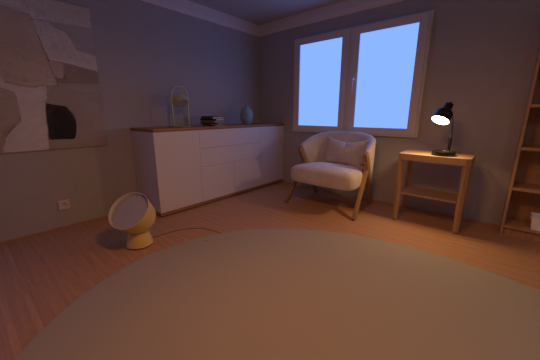} \\[\paperfiggap]
    \includegraphics[width=\relw]{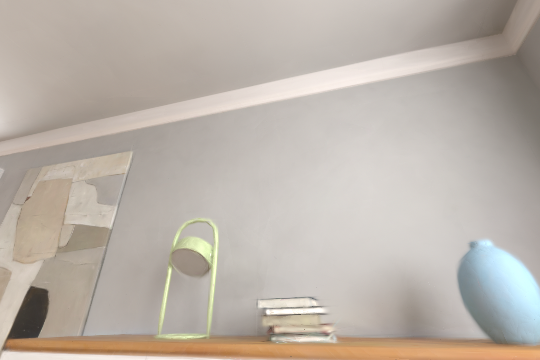} &
    \includegraphics[width=\relw]{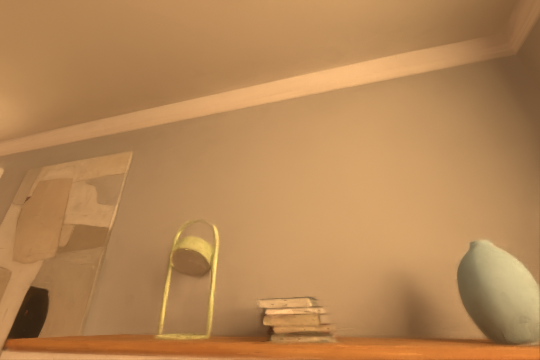} &
    \includegraphics[width=\relw]{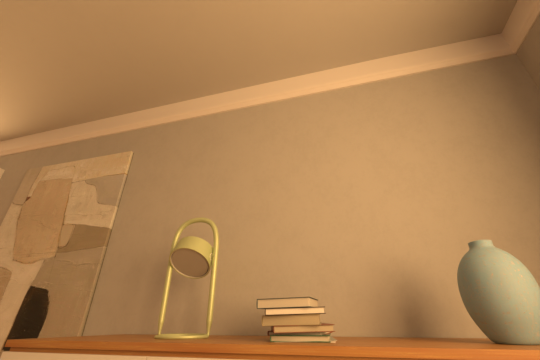} &
    \includegraphics[width=\relw]{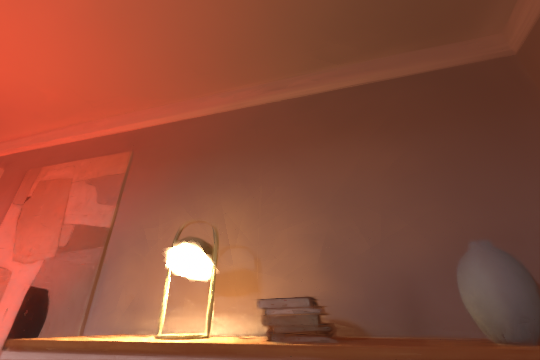} &
    \includegraphics[width=\relw]{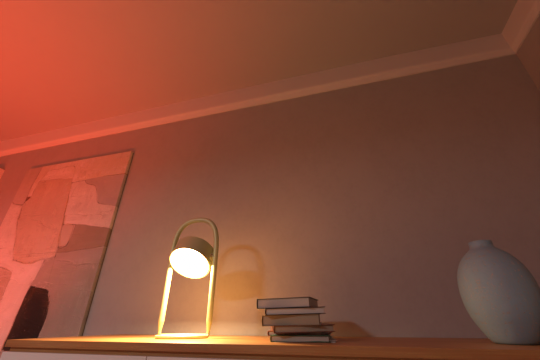} &
    \includegraphics[width=\relw]{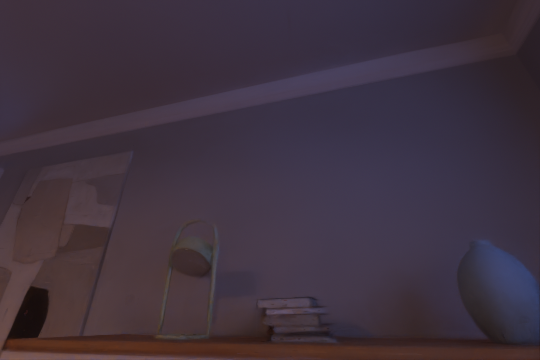} &
    \includegraphics[width=\relw]{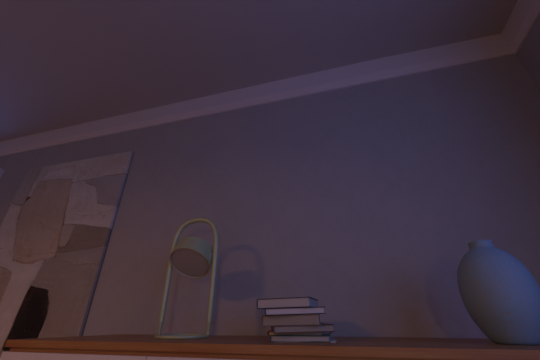} \\[\paperfiggap]
    \includegraphics[width=\relw]{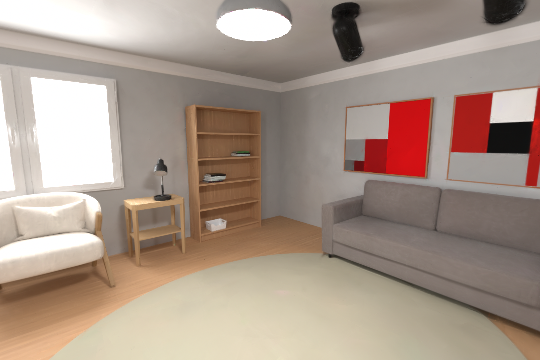} &
    \includegraphics[width=\relw]{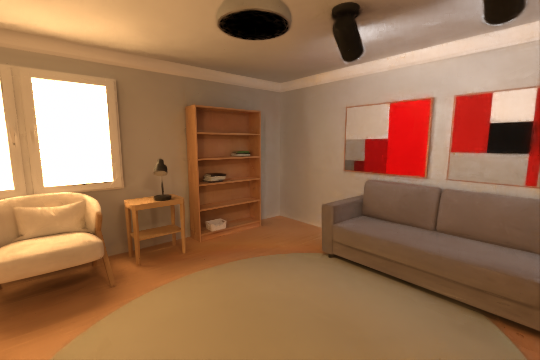} &
    \includegraphics[width=\relw]{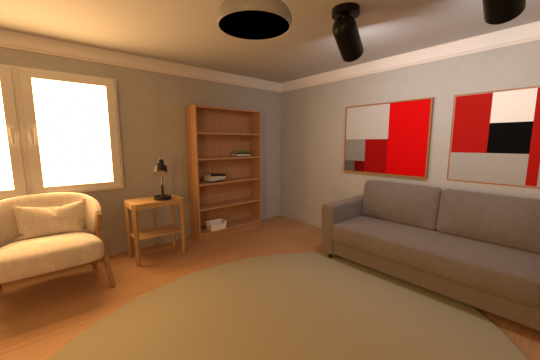} &
    \includegraphics[width=\relw]{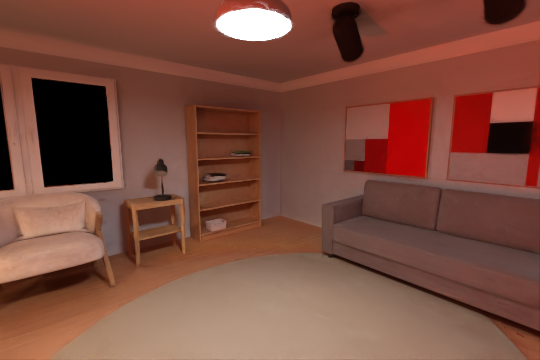} &
    \includegraphics[width=\relw]{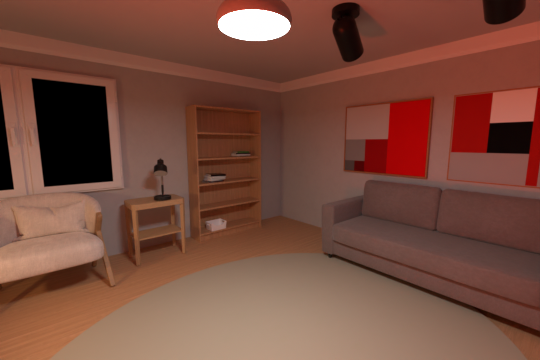} &
    \includegraphics[width=\relw]{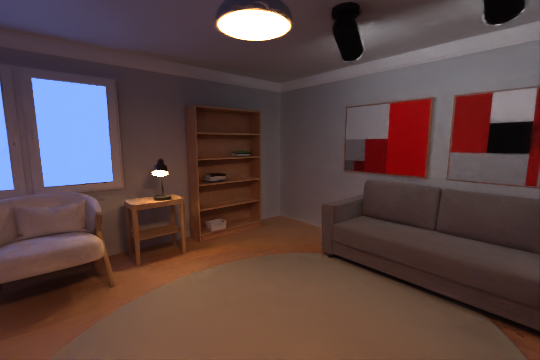} &
    \includegraphics[width=\relw]{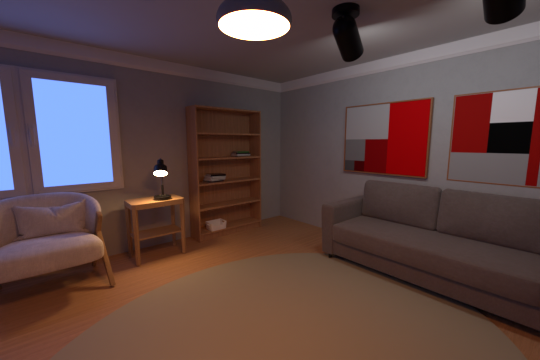} \\[\paperfiggap]
    \includegraphics[width=\relw]{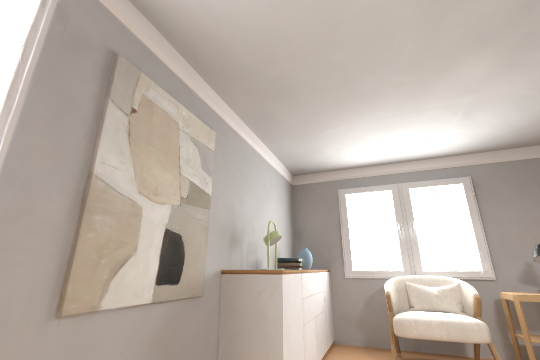} &
    \includegraphics[width=\relw]{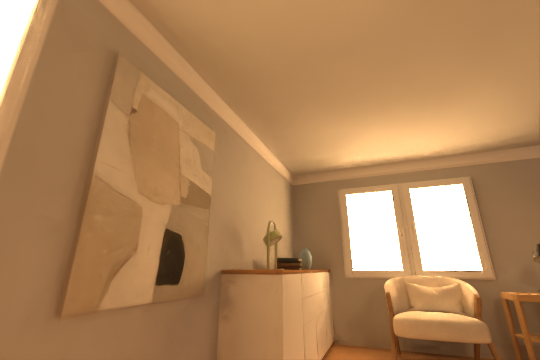} &
    \includegraphics[width=\relw]{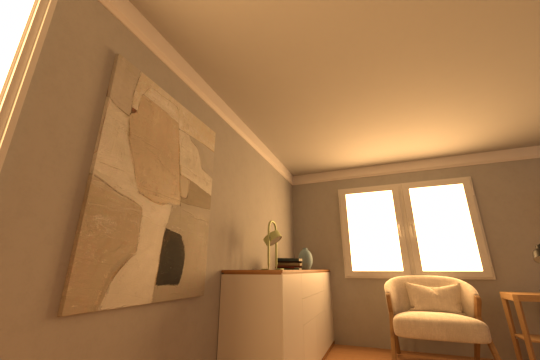} &
    \includegraphics[width=\relw]{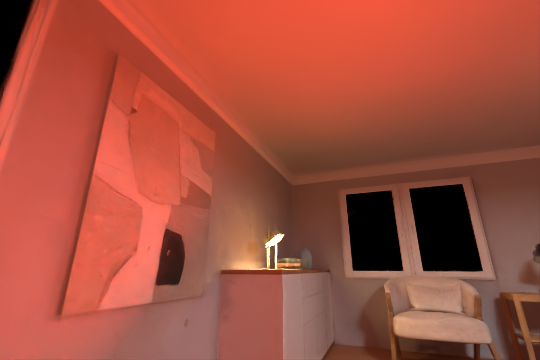} &
    \includegraphics[width=\relw]{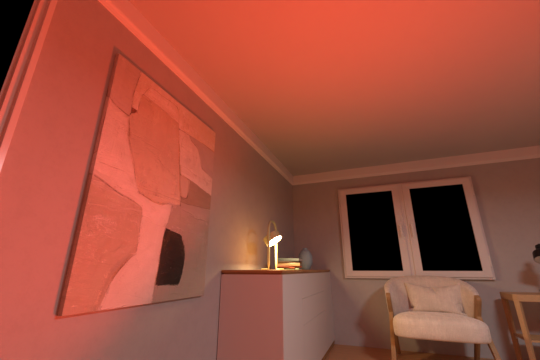} &
    \includegraphics[width=\relw]{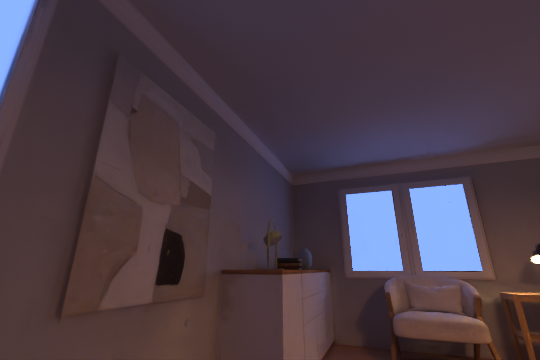} &
    \includegraphics[width=\relw]{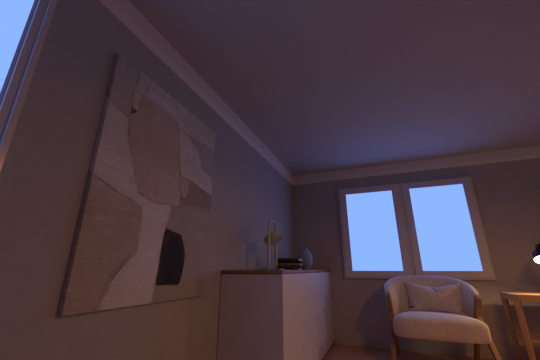} \\[\paperfiggap]
    \includegraphics[width=\relw]{figures/livingroom_relighting_page/cam29_capture.png} &
    \includegraphics[width=\relw]{figures/livingroom_relighting_page/cam29_relit1.png} &
    \includegraphics[width=\relw]{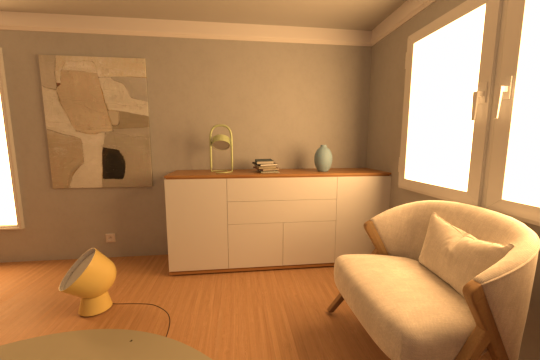} &
    \includegraphics[width=\relw]{figures/livingroom_relighting_page/cam29_relit2.png} &
    \includegraphics[width=\relw]{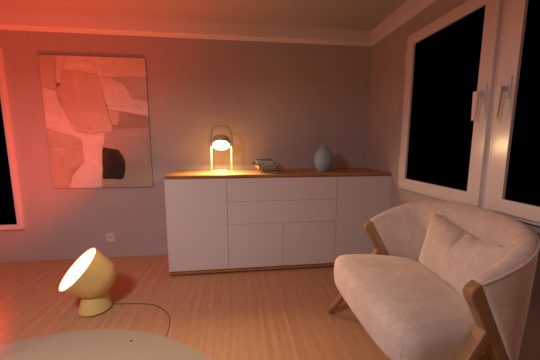} &
    \includegraphics[width=\relw]{figures/livingroom_relighting_page/cam29_relit3.png} &
    \includegraphics[width=\relw]{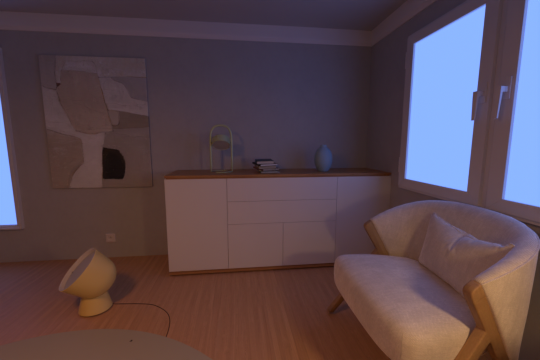} \\[\paperfiggap]
    \includegraphics[width=\relw]{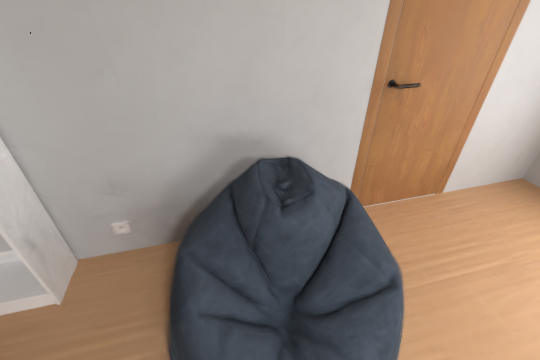} &
    \includegraphics[width=\relw]{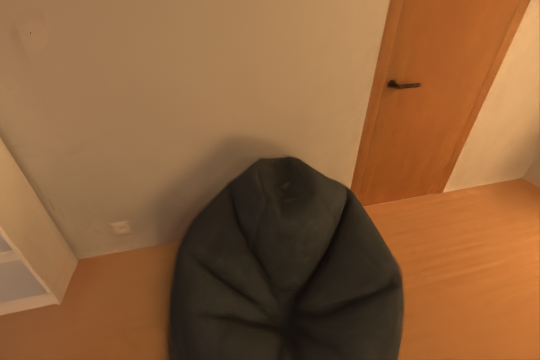} &
    \includegraphics[width=\relw]{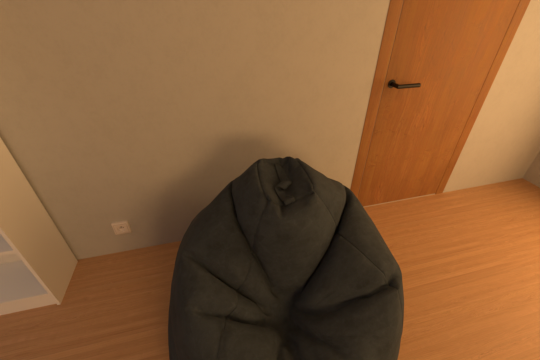} &
    \includegraphics[width=\relw]{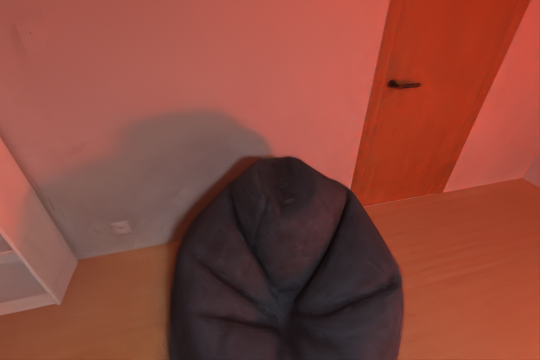} &
    \includegraphics[width=\relw]{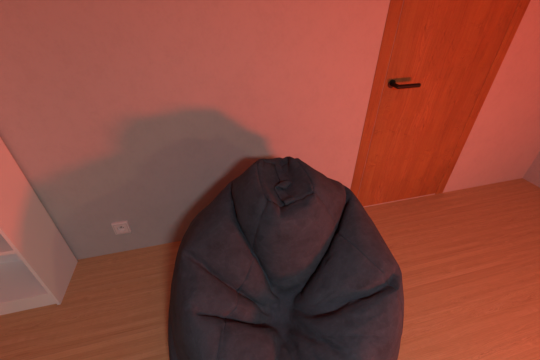} &
    \includegraphics[width=\relw]{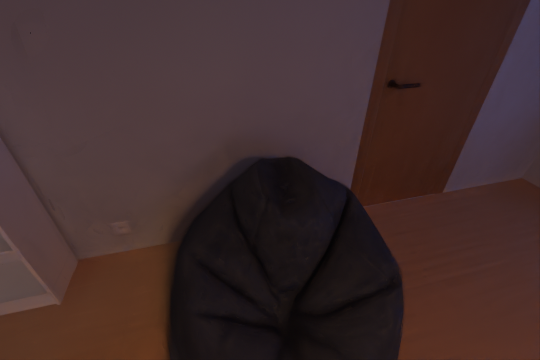} &
    \includegraphics[width=\relw]{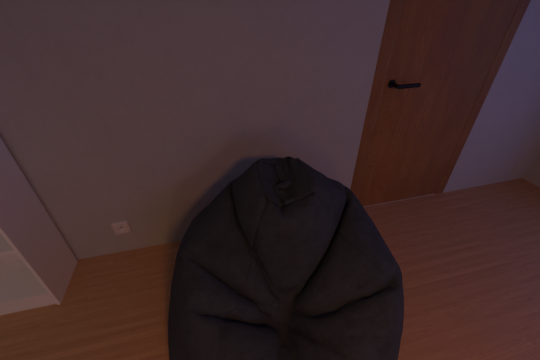} \\[\paperfiggap]
    \includegraphics[width=\relw]{figures/livingroom_relighting_page/cam35_capture.png} &
    \includegraphics[width=\relw]{figures/livingroom_relighting_page/cam35_relit1.png} &
    \includegraphics[width=\relw]{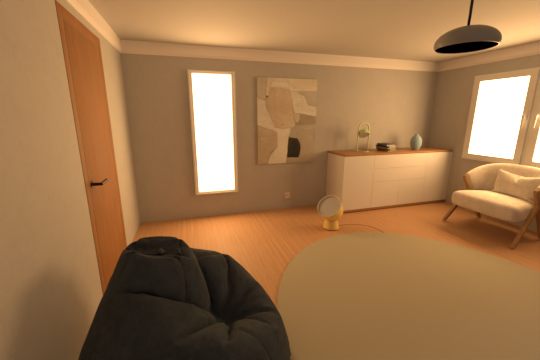} &
    \includegraphics[width=\relw]{figures/livingroom_relighting_page/cam35_relit2.png} &
    \includegraphics[width=\relw]{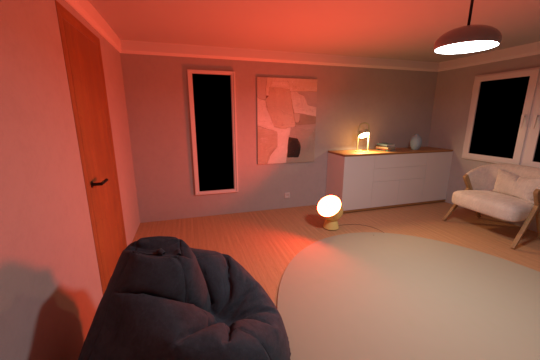} &
    \includegraphics[width=\relw]{figures/livingroom_relighting_page/cam35_relit3.png} &
    \includegraphics[width=\relw]{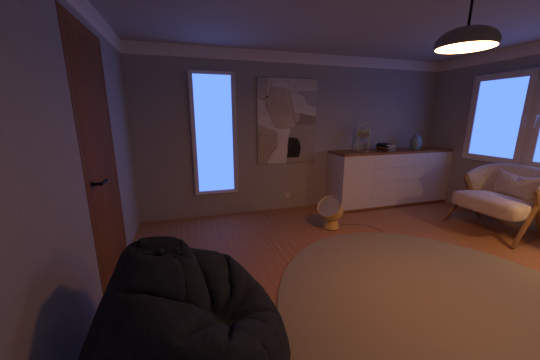} \\[\paperfiggap]
    \includegraphics[width=\relw]{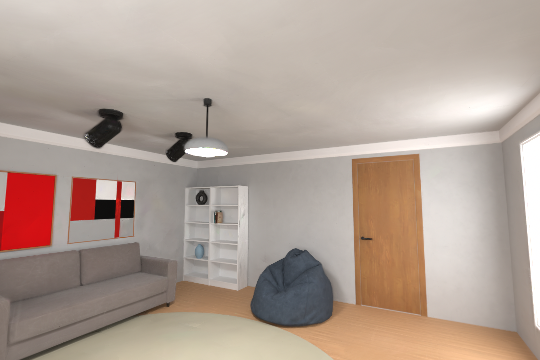} &
    \includegraphics[width=\relw]{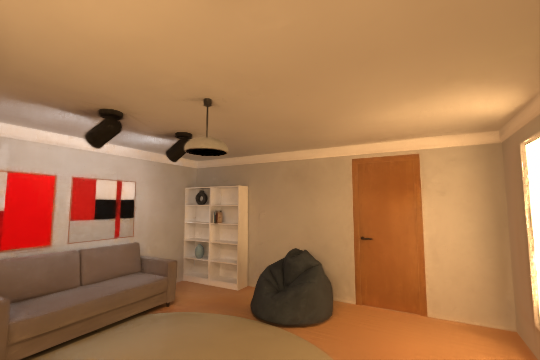} &
    \includegraphics[width=\relw]{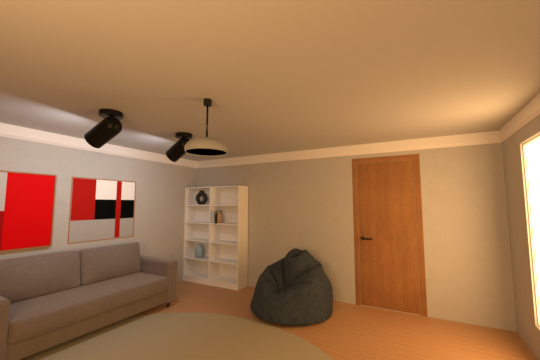} &
    \includegraphics[width=\relw]{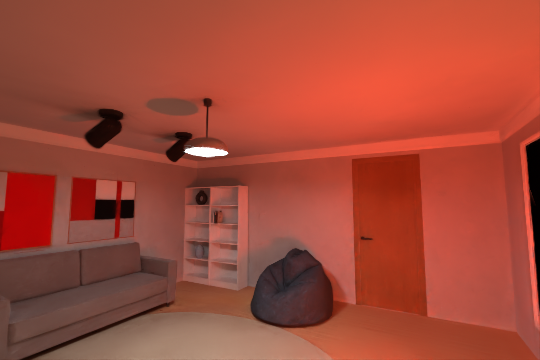} &
    \includegraphics[width=\relw]{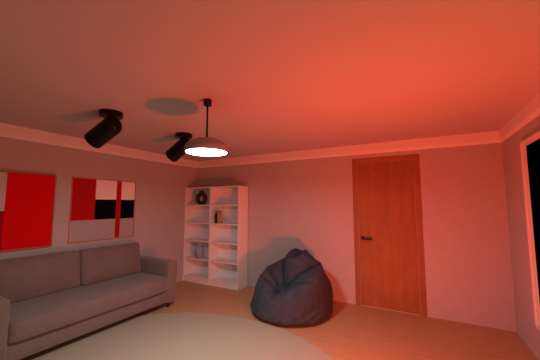} &
    \includegraphics[width=\relw]{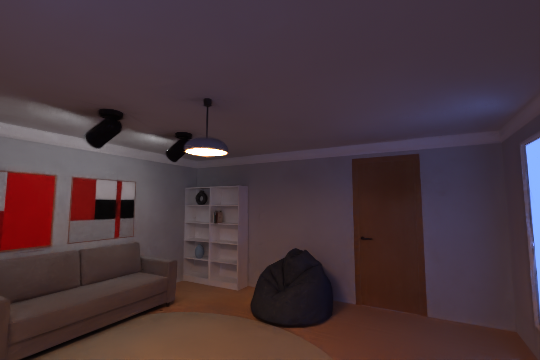} &
    \includegraphics[width=\relw]{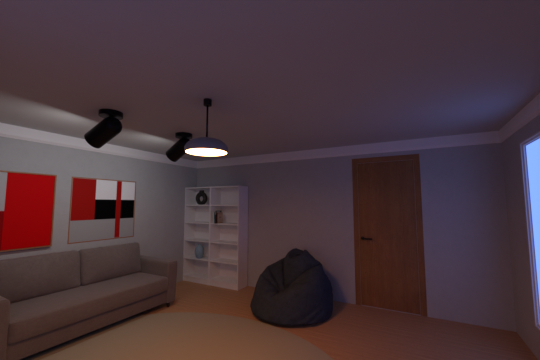} \\[\paperfiggap]
    \includegraphics[width=\relw]{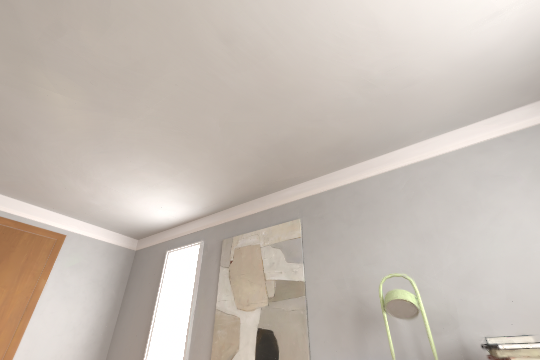} &
    \includegraphics[width=\relw]{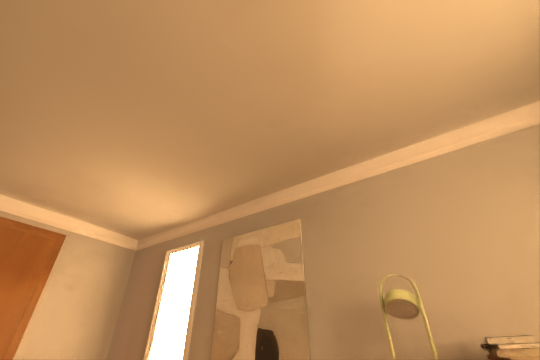} &
    \includegraphics[width=\relw]{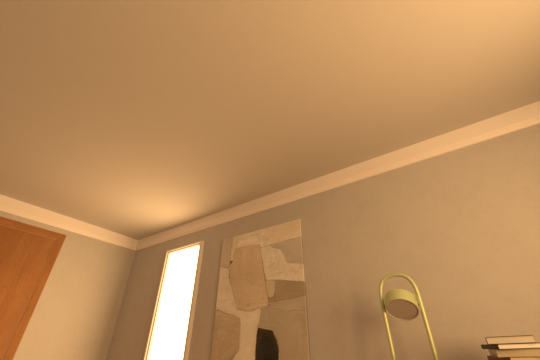} &
    \includegraphics[width=\relw]{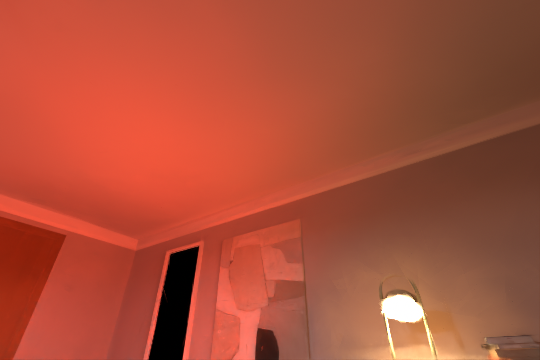} &
    \includegraphics[width=\relw]{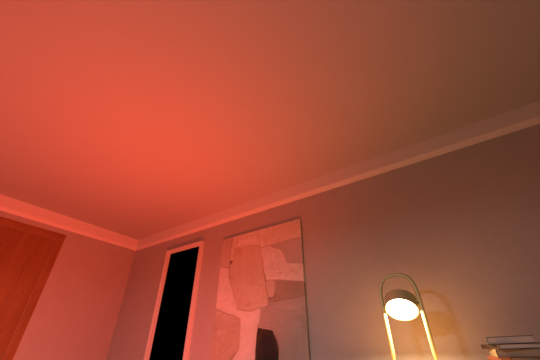} &
    \includegraphics[width=\relw]{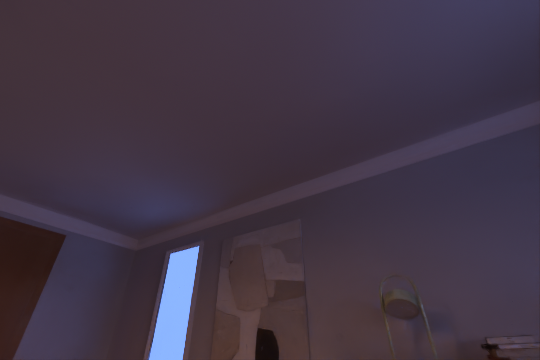} &
    \includegraphics[width=\relw]{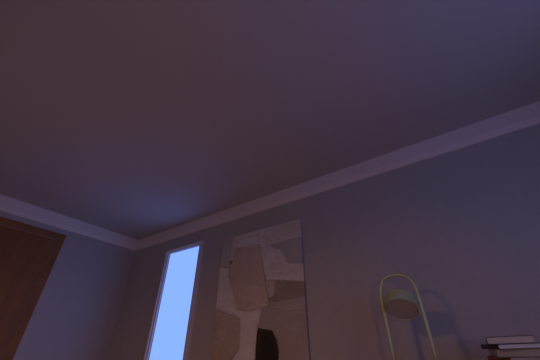} \\
  \end{tabular}
  \caption{\emph{Living room} relighting for different test views, one per row: our reconstruction under capture lighting (Capture) and our relighting to the three conditions of Fig.~\ref{fig:relighting}, which change emission strength and color, switch lights off, and let surfaces that are non-emitting during capture become emissive, each next to its path-traced ground truth (GT).}
  \label{fig:relighting_livingroom_all}
\end{figure*}

\begin{table*}[t]
  \centering
  \small
  \renewcommand{\arraystretch}{0.95}%
  \edef\padTsep{\the\dimexpr\belowrulesep/2\relax}%
  \edef\padBsep{\the\dimexpr\aboverulesep/2\relax}%
  \setlength{\aboverulesep}{0pt}\setlength{\belowrulesep}{0pt}%
  \def\padT{\rule{0pt}{\dimexpr\arraystretch\ht\strutbox+\padTsep\relax}}%
  \def\padB{\rule[-\dimexpr\arraystretch\dp\strutbox+\padBsep\relax]{0pt}{0pt}}%
\begin{tabular}{@{}l*{12}{c}@{}}
    \toprule
    \padT & \multicolumn{4}{c}{Relit~1} & \multicolumn{4}{c}{Relit~2} & \multicolumn{4}{c}{Relit~3} \\
    \cmidrule(r){2-5}\cmidrule(lr){6-9}\cmidrule(l){10-13}
    \padB & PSNR$\uparrow$ & SSIM$\uparrow$ & LPIPS$\downarrow$ & FLIP$\downarrow$ & PSNR$\uparrow$ & SSIM$\uparrow$ & LPIPS$\downarrow$ & FLIP$\downarrow$ & PSNR$\uparrow$ & SSIM$\uparrow$ & LPIPS$\downarrow$ & FLIP$\downarrow$ \\
    \midrule
    \multicolumn{13}{@{}l}{\padT\emph{Kitchen}} \\
    \multicolumn{13}{@{}l}{\textit{vs.\ single-bounce ground truth}} \\
    \quad Emission + NEE (ours) & \textbf{35.01} & \textbf{0.930} & \textbf{0.058} & \textbf{0.295} & \textbf{36.65} & \textbf{0.972} & \textbf{0.062} & \textbf{0.181} &
    \textbf{36.30} & \textbf{0.972} & \textbf{0.067} & \textbf{0.170} \\
    \quad Radiosity $E{+}T_\rho E$ & 30.48 & 0.857 & 0.131 & 0.384 & 33.46 & 0.938 & 0.101 & 0.250 & 33.69 & 0.965 & 0.077 & 0.217 \\
    \multicolumn{13}{@{}l}{\textit{vs.\ full ground truth}} \\
    \quad Emission + NEE only & 22.02 & 0.659 & 0.107 & 0.681 & 21.72 & 0.733 & 0.089 & 0.601 & 20.26 & 0.819 & 0.100 & 0.643 \\
    \quad Radiosity only & 32.84 & 0.973 & 0.092 & 0.179 & 35.52 & 0.979 & 0.079 & 0.132 & 35.96 & 0.983 & 0.071 & \textbf{0.117} \\
    \padB \quad Composite (ours) & \textbf{35.95} & \textbf{0.983} & \textbf{0.059} & \textbf{0.139} & \textbf{36.99} & \textbf{0.983} & \textbf{0.058} & \textbf{0.125} & \textbf{36.96} & \textbf{0.984} & \textbf{0.063} & \textbf{0.117} \\
    \midrule
    \multicolumn{13}{@{}l}{\padT\emph{Living room}} \\
    \multicolumn{13}{@{}l}{\textit{vs.\ single-bounce ground truth}} \\
    \quad Emission + NEE (ours) & \textbf{31.28} & \textbf{0.955} & \textbf{0.094} & \textbf{0.193} & \textbf{32.97} & \textbf{0.956} & \textbf{0.082} & \textbf{0.214} & \textbf{34.75} & \textbf{0.969} & \textbf{0.069} & \textbf{0.201} \\
    \quad Radiosity $E{+}T_\rho E$ & 30.50 & 0.939 & 0.113 & 0.205 & 27.69 & 0.887 & 0.155 & 0.395 & 32.12 & 0.939 & 0.093 & 0.283 \\
    \multicolumn{13}{@{}l}{\textit{vs.\ full ground truth}} \\
    \quad Emission + NEE only & 18.97 & 0.845 & 0.129 & 0.589 & 19.83 & 0.723 & 0.133 & 0.690 & 22.69 & 0.778 & 0.098 & 0.634 \\
    \quad Radiosity only & \textbf{31.92} & 0.969 & 0.095 & \textbf{0.159} & 30.51 & 0.965 & 0.107 & 0.215 & 34.31 & 0.977 & 0.076 & 0.164 \\
    \padB \quad Composite (ours) & 31.90 & \textbf{0.972} & \textbf{0.084} & 0.165 & \textbf{33.22} & \textbf{0.972} & \textbf{0.078} & \textbf{0.163} & \textbf{35.57} & \textbf{0.981} & \textbf{0.062} & \textbf{0.149} \\
    \bottomrule
  \end{tabular}
  \caption{
  Accuracy of the transport components for both scenes and all relit conditions (\cref{fig:relighting_kitchen_all,fig:relighting_livingroom_all}).
  Against the single-bounce ground truth (emission and one reflection), we compare our Gaussian leg (visible emission and NEE direct light) with the first two radiosity orders $E+T_\rho E$ projected from the proxy mesh.
  Against the full ground truth, we compare the Gaussian leg alone (no interreflection), the radiosity solution alone (its projected first two orders and projected higher orders composited in image space), and our composite of the Gaussian leg with the projected higher radiosity orders.
  The composite rows repeat Tab.~\ref{tab:relighting}.
  }
  \label{tab:transport_components}
\end{table*}

\section{Additional Results}
\subsection{Relighting}
\label{sec:relighting-additional-results}
\cref{fig:relighting_kitchen_all,fig:relighting_livingroom_all} show additional test views for all relighting conditions of both our scenes next to their path-traced ground truth, and Tab.~\ref{tab:transport_components} evaluates the transport components of the hybrid image separately and additionally reports LPIPS~\cite{zhang2018unreasonable} (VGG variant).

\subsection{Albedo Editing}
\label{sec:exp-albedo}
\begin{figure}[t]
  \centering
  \paperfigcaptionstyle
  \edef\editw{\the\dimexpr0.325\linewidth\relax}%
  \def\edithdr#1{\paperfigheader{\editw}{#1}}%
  \paperfigsetup
  \begin{tabular}{c@{\hspace{\paperfiggap}}c@{\hspace{\paperfiggap}}c}
    \edithdr{Capture} & \edithdr{Edited (ours)} & \edithdr{GT} \\[\paperfighdrgap]
    \includegraphics[width=\editw]{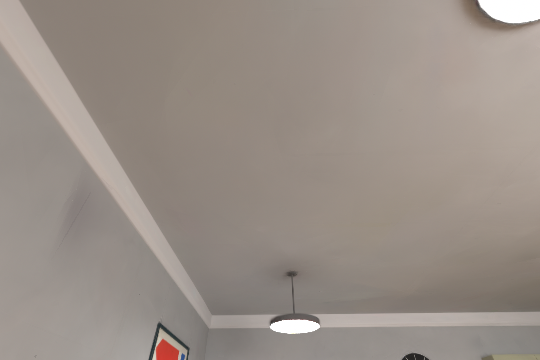} &
    \includegraphics[width=\editw]{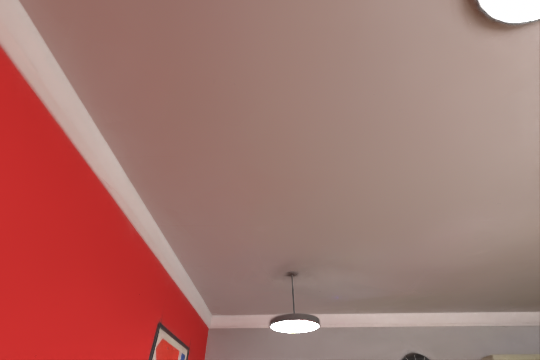} &
    \includegraphics[width=\editw]{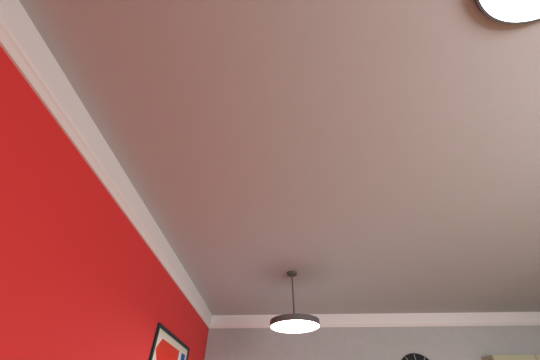} \\[\paperfiggap]
    \includegraphics[width=\editw]{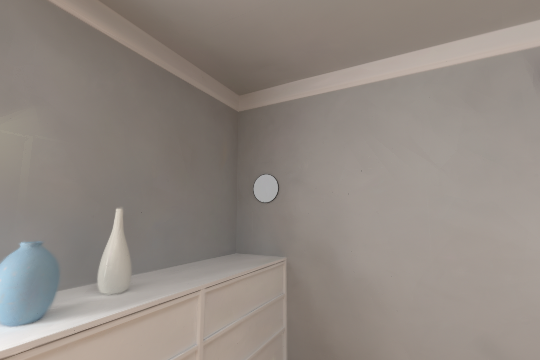} &
    \includegraphics[width=\editw]{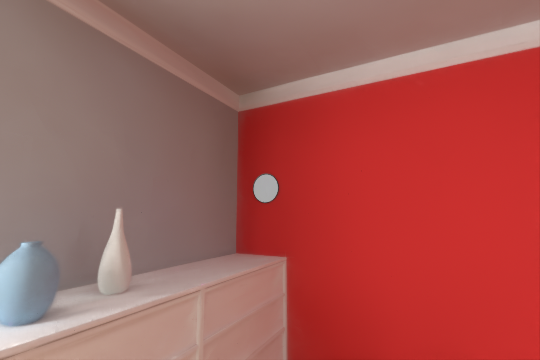} &
    \includegraphics[width=\editw]{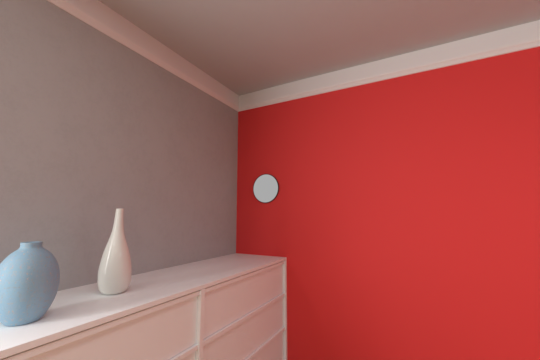} \\[\paperfiggap]
    \includegraphics[width=\editw]{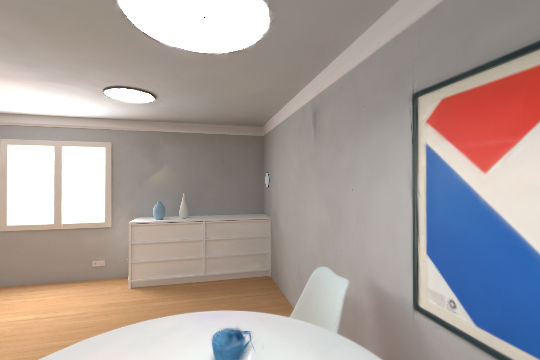} &
    \includegraphics[width=\editw]{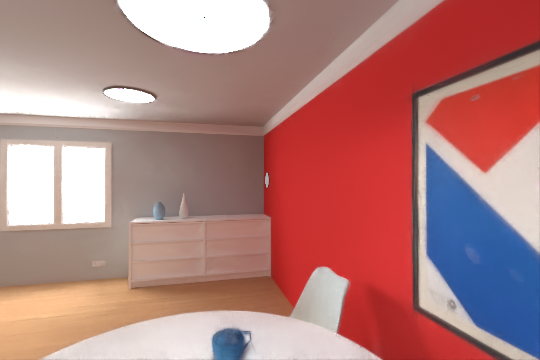} &
    \includegraphics[width=\editw]{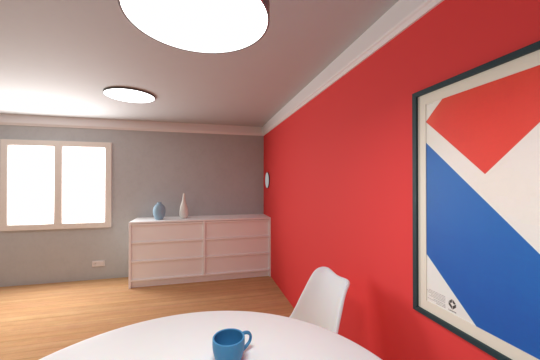} \\
  \end{tabular}
  \caption{
  Wall-albedo editing in \emph{Kitchen} from three viewpoints, one per row:
  the reconstructed scene under capture lighting (left), our result with the wall changed to red (middle), and the corresponding path-traced ground truth (right).
  Capture lighting is held fixed across these panels, so the red tint on the ceiling and the neighboring surfaces shows interreflection from the edited wall.  
  }
  \label{fig:albedo_edit}
\end{figure}

Albedo edits change the light reflected at each bounce while leaving geometric exchange unchanged.
We evaluate this capability by changing one \emph{Kitchen} wall's albedo to red under Relit~3, retaining $F$ and $P$ and comparing against independently path-traced references with the same lighting and material edit.
Over all 57 test views, the edited composite achieves 36.10\,dB PSNR, 0.978 SSIM, and 0.061 LPIPS.
For comparison, Relit~3 with unchanged materials scores 36.96\,dB, 0.984, and 0.063 against its corresponding reference (Tab.~\ref{tab:transport_components}).
Fig.~\ref{fig:albedo_edit} shows the wall edit under capture lighting, including red interreflection on the ceiling and adjacent wall.

\subsection{Fixture Editing with EAG-PT}
\label{sec:fixture-eagpt}
We also evaluate EAG-PT on the dining-pendant translations of Tab.~\ref{tab:fixture_edits}.
EAG-PT stores no precomputed light transport that an edit could update, so emitter re-injection does not apply and every edited scene is path traced in full (7 bounces, 1024\,spp, denoised), at 133--135\,s per view on an RTX~4090 at $540{\times}360$.
We apply the same similarity edit to the same fixture Gaussians of EAG-PT's relit representation, so both methods see identical pendant geometry.

\begin{figure}[t]
  \centering
  \paperfigcaptionstyle
  \edef\osw{\the\dimexpr(\linewidth-\paperfiglabelwidth-4\paperfiggap)/3\relax}%
  \edef\osh{\the\dimexpr\osw*2/3\relax}%
  \def\oslab#1{\paperfigrowlabel{\osh}{#1}}%
  \def\oshdr#1{\paperfigheader{\osw}{#1}}%
  \paperfigsetup
  \begin{tabular}{c@{\hspace{\paperfiggap}}c@{\hspace{\paperfiggap}}c@{\hspace{\paperfiggap}}c}
    & \oshdr{Unedited} & \oshdr{Moved 1\,m} & \oshdr{Moved 2\,m} \\[\paperfighdrgap]
    \oslab{GT} &
    \includegraphics[width=\osw]{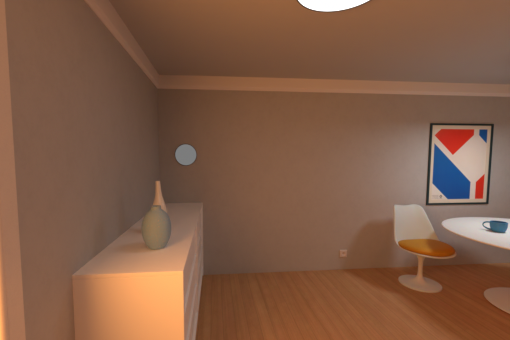} &
    \includegraphics[width=\osw]{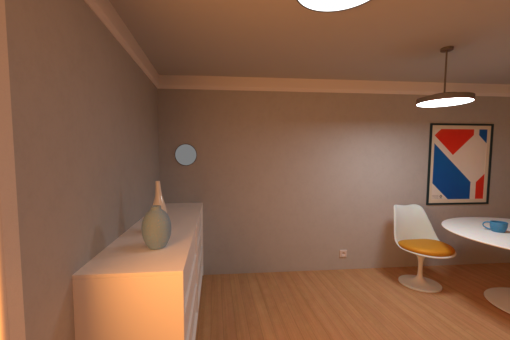} &
    \includegraphics[width=\osw]{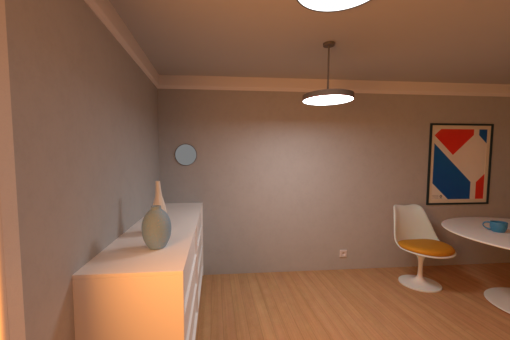} \\[\paperfiggap]
    \oslab{EAG-PT} &
    \includegraphics[width=\osw]{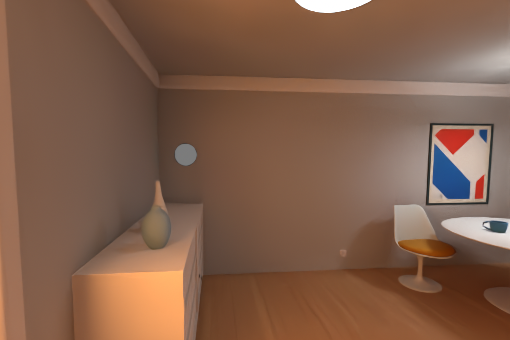} &
    \includegraphics[width=\osw]{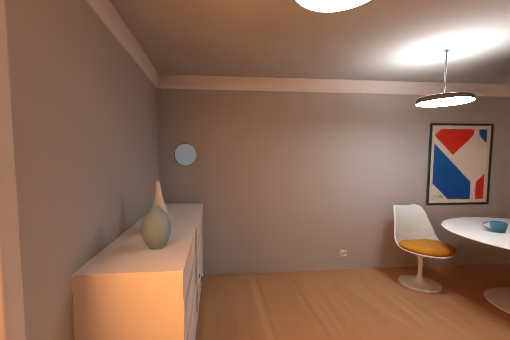} &
    \includegraphics[width=\osw]{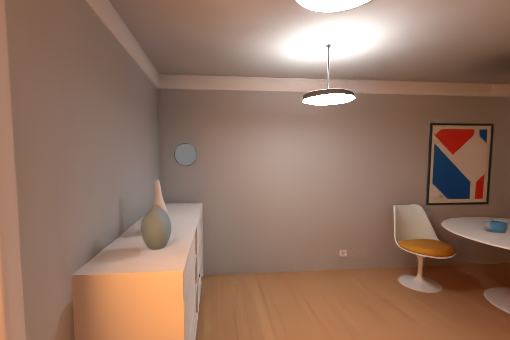} \\
  \end{tabular}
  \caption{
  EAG-PT under dining-pendant translation, for the view shown for our method in Fig.~\ref{fig:fixture_edits}.
  As the pendant moves, a glow appears on the ceiling around it that is absent from the reference.
  PSNR for this view falls from 26.62\,dB to 20.03\,dB and 19.00\,dB.
  }
  \label{fig:eagpt_pendant}
\end{figure}

Path tracing re-evaluates visibility and transport for the moved fixture, yet Tab.~\ref{tab:eagpt_pendant} shows that EAG-PT loses 3.15\,dB between the original position and the 2\,m translation, whereas our re-injection result loses 0.61\,dB, and our lead grows from 4.80 to 7.35\,dB.
We attribute this decline to two errors in the representation that partly cancel at the original position.
First, EAG-PT's two-sided emission model sends pendant light upward onto the ceiling, which does not happen in the reference.
Second, the recovered ceiling albedo is darkened directly above the original pendant position, falling from about 0.56 beyond 0.5\,m to 0.24 within 0.1\,m, a pattern consistent with compensation for excess incident light during albedo recovery.
After translation, the upward light lands on brighter ceiling and produces a visible glow (Fig.~\ref{fig:eagpt_pendant}), while the dark patch remains at the original position.
The effect depends on the view: two views that gain the glow in frame lose more than 18\,dB at 2\,m, while one that sees the glow at the original position gains 11\,dB once the pendant leaves the frame.

\begin{table}[t]
\centering
\small
\renewcommand{\arraystretch}{0.95}%
\edef\padTsep{\the\dimexpr\belowrulesep/2\relax}%
\edef\padBsep{\the\dimexpr\aboverulesep/2\relax}%
\setlength{\aboverulesep}{0pt}\setlength{\belowrulesep}{0pt}%
\def\padT{\rule{0pt}{\dimexpr\arraystretch\ht\strutbox+\padTsep\relax}}%
\def\padB{\rule[-\dimexpr\arraystretch\dp\strutbox+\padBsep\relax]{0pt}{0pt}}%
\begin{tabular}{@{}lccccc@{}}
  \toprule
  \padT & \multicolumn{2}{c}{EAG-PT} & \multicolumn{2}{c}{Ours (re-injection)} & \\
  \cmidrule(lr){2-3}\cmidrule(lr){4-5}
  \padB Translation & PSNR$\uparrow$ & SSIM$\uparrow$ & PSNR$\uparrow$ & SSIM$\uparrow$ & $\Delta$PSNR \\
  \midrule
  \padT None & 32.14 & 0.975 & 36.96 & 0.984 & $+4.80$ \\
  0.5\,m   & 31.19 & 0.972 & 36.70 & 0.983 & $+5.51$ \\
  1\,m     & 29.98 & 0.968 & 36.73 & 0.983 & $+6.75$ \\
  \padB 2\,m     & 29.00 & 0.964 & 36.35 & 0.982 & $+7.35$ \\
  \bottomrule
\end{tabular}
\caption{
Dining-pendant translation under Relit~3 in \emph{Kitchen}, against ground truth of each edited scene.
Ours repeats the re-injection rows of Tab.~\ref{tab:fixture_edits}; the untranslated row uses the unedited production composite.
$\Delta$ is ours minus EAG-PT.
}
\label{tab:eagpt_pendant}
\end{table}

\subsection{Resolution Scaling}
\label{sec:resolution-scaling}
\cref{tab:resolution-runtime} lists the per-condition latencies behind the ranges quoted in the main paper: a $9\times$ increase in pixel count raises our edit-to-first-image latency by only 3.1--4.0$\times$, while EAG-PT's per-view path tracing grows by about $8.7\times$.

\begin{table}[t]
  \centering
  \small
  \renewcommand{\arraystretch}{0.95}%
  \edef\padTsep{\the\dimexpr\belowrulesep/2\relax}%
  \edef\padBsep{\the\dimexpr\aboverulesep/2\relax}%
  \setlength{\aboverulesep}{0pt}\setlength{\belowrulesep}{0pt}%
  \def\padT{\rule{0pt}{\dimexpr\arraystretch\ht\strutbox+\padTsep\relax}}%
  \def\padB{\rule[-\dimexpr\arraystretch\dp\strutbox+\padBsep\relax]{0pt}{0pt}}%
  \providecommand{\tbd}{\textcolor{red}{tbd}}%
  \begin{tabular}{@{}l*{6}{c}@{}}
    \toprule
    \padT & \multicolumn{2}{c}{Relit~1} & \multicolumn{2}{c}{Relit~2} & \multicolumn{2}{c}{Relit~3} \\
    \cmidrule(r){2-3}\cmidrule(lr){4-5}\cmidrule(l){6-7}
    \padB Resolution & EAG-PT & Ours & EAG-PT & Ours & EAG-PT & Ours \\
    \midrule
    \padT $540{\times}360$ & 167.9 & \textbf{0.250} & 184.4 & \textbf{0.211} & 166.5 & \textbf{0.248} \\
    $1080{\times}720$  & 660.0  & \textbf{0.529} & 723.2  & \textbf{0.380} & 654.5  & \textbf{0.481} \\
    \padB $1620{\times}1080$ & 1473.9  & \textbf{0.993} & 1606.9  & \textbf{0.653} & 1444.9  & \textbf{0.862} \\
    \bottomrule
  \end{tabular}
  \caption{
  Runtime scaling with image resolution in \emph{Living room} (seconds).
  EAG-PT reports path tracing per view; ours reports edit-to-first-image latency.
  For our method, only the per-view rendering grows with pixel count; source assignment, emitter injection, radiosity solve, and applying $P$ do not depend on resolution.
  }
  \label{tab:resolution-runtime}
\end{table}

\section{Ablations}
\label{sec:ablations}
\paragraph{Direct light estimation.}
\cref{tab:nee-ablation} compares our denoised NEE estimator at 8\,spp with denoised single-bounce path tracing at 1024\,spp under otherwise identical stages. 
This replacement changes composite PSNR by at most 0.47\,dB, while raising edit-to-first-image latency from about 0.3\,s to about 25\,s. 

\begin{table*}[t]
  \centering
  \small
  \renewcommand{\arraystretch}{0.95}%
  \edef\padTsep{\the\dimexpr\belowrulesep/2\relax}%
  \edef\padBsep{\the\dimexpr\aboverulesep/2\relax}%
  \setlength{\aboverulesep}{0pt}\setlength{\belowrulesep}{0pt}%
  \def\padT{\rule{0pt}{\dimexpr\arraystretch\ht\strutbox+\padTsep\relax}}%
  \def\padB{\rule[-\dimexpr\arraystretch\dp\strutbox+\padBsep\relax]{0pt}{0pt}}%
  \begin{tabular}{@{}ll*{8}{c}@{}}
    \toprule
    \padT & & \multicolumn{3}{c}{Emission + direct}
      & \multicolumn{3}{c}{Composite}
      & \multicolumn{2}{c}{Time (s)} \\
    \cmidrule(lr){3-5}
    \cmidrule(lr){6-8}
    \cmidrule(l){9-10}
    \padB & Method
      & PSNR$\uparrow$ & SSIM$\uparrow$ & FLIP$\downarrow$
      & PSNR$\uparrow$ & SSIM$\uparrow$ & FLIP$\downarrow$
      & Trace$\downarrow$ & Edit$\downarrow$ \\
    \midrule
    \padT Relit~1 & NEE 8\,spp
      & 35.01 & 0.930 & 0.295
      & \textbf{35.95} & \textbf{0.983} & \textbf{0.139}
      & \textbf{0.059} & \textbf{0.244} \\
      \padB & PT 1024\,spp
      & \textbf{35.60} & \textbf{0.938} & \textbf{0.271}
      & 35.92 & \textbf{0.983} & 0.143
      & 24.45 & 24.77 \\
    \midrule
    \padT Relit~2 & NEE 8\,spp
      & 36.65 & 0.972 & 0.181
      & \textbf{36.99} & \textbf{0.983} & \textbf{0.125}
      & \textbf{0.092} & \textbf{0.307} \\
      \padB & PT 1024\,spp
      & \textbf{36.74} & \textbf{0.974} & \textbf{0.168}
      & 36.52 & \textbf{0.983} & 0.133
      & 24.34 & 24.79 \\
    \midrule
    \padT Relit~3 & NEE 8\,spp
      & 36.30 & 0.972 & 0.170
      & 36.96 & 0.984 & \textbf{0.117}
      & \textbf{0.088} & \textbf{0.287} \\
      \padB & PT 1024\,spp
      & \textbf{37.63} & \textbf{0.977} & \textbf{0.157}
      & \textbf{37.26} & \textbf{0.986} & 0.124
      & 24.22 & 24.70 \\
    \bottomrule
  \end{tabular}
  \caption{
    Direct-light estimator ablation on \emph{Kitchen}.
    Emission-plus-direct renders are evaluated against single-bounce references, and composites against full ground truth.
    Both estimators use the same denoising; all non-direct stages remain unchanged.
    \emph{Trace} reports per-frame ray-tracing time; denoising adds 1--2\,ms.
    \emph{Edit} reports edit-to-first-image latency, including denoising.
  }
  \label{tab:nee-ablation}
\end{table*}

\cref{tab:nee-spp} sweeps the sampling budget from 2 to 64\,spp under identical denoising. 
Increasing NEE from 4 to 8\,spp improves mean emission-plus-direct PSNR by 0.26\,dB (render-and-denoise time: from 50.9 to 80.9\,ms), while 64\,spp adds only $\sim$0.4\,dB over 8\,spp at $6.1\times$ the cost, so we use 8\,spp.

\begin{table*}[t]
  \centering
  \small
  \renewcommand{\arraystretch}{0.95}%
  \edef\padTsep{\the\dimexpr\belowrulesep/2\relax}%
  \edef\padBsep{\the\dimexpr\aboverulesep/2\relax}%
  \setlength{\aboverulesep}{0pt}\setlength{\belowrulesep}{0pt}%
  \def\padT{\rule{0pt}{\dimexpr\arraystretch\ht\strutbox+\padTsep\relax}}%
  \def\padB{\rule[-\dimexpr\arraystretch\dp\strutbox+\padBsep\relax]{0pt}{0pt}}%
  \begin{tabular}{@{}l*{12}{c}@{}}
    \toprule
    \padT & \multicolumn{4}{c}{Relit~1} & \multicolumn{4}{c}{Relit~2} & \multicolumn{4}{c@{}}{Relit~3} \\
    \cmidrule(r){2-5}\cmidrule(lr){6-9}\cmidrule(l){10-13}
    \padB Method
      & PSNR$\uparrow$ & SSIM$\uparrow$ & FLIP$\downarrow$ & Time (s)$\downarrow$
      & PSNR$\uparrow$ & SSIM$\uparrow$ & FLIP$\downarrow$ & Time (s)$\downarrow$
      & PSNR$\uparrow$ & SSIM$\uparrow$ & FLIP$\downarrow$ & Time (s)$\downarrow$ \\
    \midrule
    \padT NEE 2\,spp
      & 34.68 & 0.926 & 0.304 & 0.030
      & 36.20 & 0.968 & 0.194 & 0.039
      & 35.33 & 0.968 & 0.188 & 0.037 \\
    NEE 4\,spp
      & 34.86 & 0.928 & 0.300 & 0.041
      & 36.47 & 0.970 & 0.187 & 0.057
      & 35.85 & 0.970 & 0.178 & 0.055 \\
    \textbf{NEE 8\,spp}
      & 35.01 & 0.930 & 0.295 & 0.061
      & 36.65 & 0.972 & 0.181 & 0.093
      & 36.30 & 0.972 & 0.170 & 0.089 \\
    NEE 16\,spp
      & 35.07 & 0.931 & 0.293 & 0.098
      & 36.76 & 0.973 & 0.177 & 0.164
      & 36.70 & 0.974 & 0.163 & 0.160 \\
    NEE 32\,spp
      & 35.08 & 0.931 & 0.292 & 0.174
      & 36.80 & 0.973 & 0.175 & 0.304
      & 36.98 & 0.975 & 0.159 & 0.299 \\
    \padB NEE 64\,spp
      & 35.09 & 0.932 & 0.291 & 0.325
      & 36.82 & 0.973 & 0.174 & 0.581
      & 37.21 & 0.976 & 0.156 & 0.575 \\
    \midrule
    \padT\padB PT 1024\,spp
      & 35.60 & 0.938 & 0.271 & 24.45
      & 36.74 & 0.974 & 0.168 & 24.34
      & 37.63 & 0.977 & 0.157 & 24.22 \\
    \bottomrule
  \end{tabular}
  \caption{
    NEE sampling budget on \emph{Kitchen}.
    Emission-plus-direct renders are evaluated against single-bounce ground truth.
    All outputs use the same OptiX denoising.
    Timings include rendering and denoising.
    The bold method label identifies our default sampling budget.
  }
  \label{tab:nee-spp}
\end{table*}

\paragraph{Proxy budget.}
\cref{tab:supp-proxy-budget} extends the proxy-mesh row of the main ablation (\cref{tab:main_ablation}) to a budget sweep.
The composite is budget-robust from 200k triangles upward: doubling to 400k changes PSNR by at most 0.08\,dB but raises the edit time to 0.41--0.49\,s, and halving to 100k lowers PSNR by 0.12--0.44\,dB and worsens FLIP in every condition.

\begin{table*}[t]
  \centering
  \small
  \renewcommand{\arraystretch}{0.95}%
  \edef\padTsep{\the\dimexpr\belowrulesep/2\relax}%
  \edef\padBsep{\the\dimexpr\aboverulesep/2\relax}%
  \setlength{\aboverulesep}{0pt}\setlength{\belowrulesep}{0pt}%
  \def\padT{\rule{0pt}{\dimexpr\arraystretch\ht\strutbox+\padTsep\relax}}%
  \def\padB{\rule[-\dimexpr\arraystretch\dp\strutbox+\padBsep\relax]{0pt}{0pt}}%
  \begin{tabular}{@{}l*{12}{c}@{}}
    \toprule
    \padT & \multicolumn{4}{c}{Relit~1} & \multicolumn{4}{c}{Relit~2} & \multicolumn{4}{c}{Relit~3} \\
    \cmidrule(r){2-5}\cmidrule(lr){6-9}\cmidrule(l){10-13}
    \padB Proxy budget
      & PSNR$\uparrow$ & SSIM$\uparrow$ & FLIP$\downarrow$ & Time (s)$\downarrow$
      & PSNR$\uparrow$ & SSIM$\uparrow$ & FLIP$\downarrow$ & Time (s)$\downarrow$
      & PSNR$\uparrow$ & SSIM$\uparrow$ & FLIP$\downarrow$ & Time (s)$\downarrow$ \\
    \midrule
    \padT 100k triangles
      & 35.59 & 0.982 & 0.153 & 0.170
      & 36.87 & 0.983 & 0.129 & 0.227
      & 36.52 & 0.983 & 0.127 & 0.212 \\
    \textbf{200k triangles}
      & 35.95 & 0.983 & 0.139 & 0.244
      & 36.99 & 0.983 & 0.125 & 0.307
      & 36.96 & 0.984 & 0.117 & 0.287 \\
    \padB 400k triangles
      & 35.95 & 0.983 & 0.139 & 0.411
      & 36.97 & 0.983 & 0.126 & 0.487
      & 36.88 & 0.984 & 0.118 & 0.461 \\
    \bottomrule
  \end{tabular}
  \caption{
  Proxy-budget sweep on \emph{Kitchen} against each condition's full ground truth. The 100k and 200k rows repeat \cref{tab:main_ablation}, and time is edit-to-first-image.
  The bold method label identifies our default proxy mesh setting.
  }
  \label{tab:supp-proxy-budget}
\end{table*}

\paragraph{Form-factor pruning.}
\cref{tab:supp-ff-prune} applies row-mass pruning to $F$ of the 200k \emph{Kitchen} proxy, keeping in each row the fewest largest entries that reach a fraction $f$ of the row's mass and discarding the rest.
At $f=0.98$, 86\% of the nonzeros remain, $F$'s GPU footprint drops from 8.08 to 6.95\,GiB, and
edit-to-first-image time falls by about 11\,ms at a cost of 0.22--0.72\,dB PSNR.
At $f=0.95$, the loss grows to 1.38--2.84\,dB and FLIP worsens in every condition. 
The loss follows the discarded transport: total solved radiosity power falls by 1.4--1.6\% at $f=0.98$ and by 4.7--4.8\% at $f=0.95$, darkening indirect light.

\begin{table*}[t]
\centering
\small
\renewcommand{\arraystretch}{0.95}%
\edef\padTsep{\the\dimexpr\belowrulesep/2\relax}%
\edef\padBsep{\the\dimexpr\aboverulesep/2\relax}%
\setlength{\aboverulesep}{0pt}\setlength{\belowrulesep}{0pt}%
\def\padT{\rule{0pt}{\dimexpr\arraystretch\ht\strutbox+\padTsep\relax}}%
\def\padB{\rule[-\dimexpr\arraystretch\dp\strutbox+\padBsep\relax]{0pt}{0pt}}%
\begin{tabular}{@{}l*{11}{c}@{}}
  \toprule
  \padT & & & \multicolumn{3}{c}{Relit~1} & \multicolumn{3}{c}{Relit~2} & \multicolumn{3}{c}{Relit~3} \\
  \cmidrule(lr){4-6}\cmidrule(lr){7-9}\cmidrule(l){10-12}
  \padB Pruning & nnz & GiB$\downarrow$
    & PSNR$\uparrow$ & SSIM$\uparrow$ & FLIP$\downarrow$
    & PSNR$\uparrow$ & SSIM$\uparrow$ & FLIP$\downarrow$
    & PSNR$\uparrow$ & SSIM$\uparrow$ & FLIP$\downarrow$ \\
  \midrule
  \padT None
    & 100\% & 8.08
    & 35.95 & 0.983 & 0.139
    & 36.99 & 0.983 & 0.125
    & 36.96 & 0.984 & 0.117 \\
  $f=0.98$
    & 86\% & 6.95
    & 35.73 & 0.982 & 0.151
    & 36.43 & 0.982 & 0.140
    & 36.24 & 0.983 & 0.129 \\
  \padB $f=0.95$
    & 76\% & 6.13
    & 34.57 & 0.980 & 0.194
    & 34.71 & 0.980 & 0.181
    & 34.12 & 0.982 & 0.168 \\
  \bottomrule
\end{tabular}
\caption{
Row-mass pruning of the form-factor matrix on \emph{Kitchen} (200k proxy), against each condition's full ground truth.
Only $F$ changes; the proxy, emitter selection, projection operator, and direct-light renders are those of the unpruned row.
\emph{nnz} is the fraction of nonzeros kept, and \emph{GiB} is $F$'s GPU footprint.
}
\label{tab:supp-ff-prune}
\end{table*}

\paragraph{Projection fitting.}
In the composite comparison of \cref{tab:main_ablation}, all three projection operators share the same direct-light renders, which carry most of the image energy, so composite metrics alone do not isolate the projection error.
\cref{tab:supp-projection} therefore evaluates the projected radiosity on its own.
For indirect light projected by each operator and evaluated against indirect ground truth (top), the fitted $P$ improves PSNR by 0.89--1.33\,dB over nearest-centroid assignment and by 0.35--0.80\,dB over the unfitted initialization $P^0$.
When the complete radiosity solution, including emission and direct light, is projected and evaluated against full ground truth without NEE (bottom), the margins grow to 2.45--4.65\,dB and 1.21--3.92\,dB, because sharp emitters and direct-light boundaries are most sensitive to the projection.
The fitted $P$ is best on every reported metric and condition.

\begin{table*}[t]
  \centering
  \small
  \renewcommand{\arraystretch}{0.95}%
  \edef\padTsep{\the\dimexpr\belowrulesep/2\relax}%
  \edef\padBsep{\the\dimexpr\aboverulesep/2\relax}%
  \setlength{\aboverulesep}{0pt}\setlength{\belowrulesep}{0pt}%
  \def\padT{\rule{0pt}{\dimexpr\arraystretch\ht\strutbox+\padTsep\relax}}%
  \def\padB{\rule[-\dimexpr\arraystretch\dp\strutbox+\padBsep\relax]{0pt}{0pt}}%
  \begin{tabular}{@{}l*{9}{c}@{}}
    \toprule
    \padT & \multicolumn{3}{c}{Relit~1} & \multicolumn{3}{c}{Relit~2} & \multicolumn{3}{c}{Relit~3} \\
    \cmidrule(r){2-4}\cmidrule(lr){5-7}\cmidrule(l){8-10}
    \padB & PSNR$\uparrow$ & SSIM$\uparrow$ & FLIP$\downarrow$ & PSNR$\uparrow$ & SSIM$\uparrow$ & FLIP$\downarrow$ & PSNR$\uparrow$ & SSIM$\uparrow$ & FLIP$\downarrow$ \\
    \midrule
    \multicolumn{10}{@{}l}{\padT\textit{vs.\ indirect ground truth}} \\
    \quad \textbf{Fitted $P$} & \textbf{37.04} & \textbf{0.978} & \textbf{0.171} & \textbf{38.19} & \textbf{0.979} & \textbf{0.148} & \textbf{37.07} & \textbf{0.965} & \textbf{0.161} \\
    \quad Nearest-centroid & 36.15 & 0.974 & 0.188 & 37.06 & 0.975 & 0.165 & 35.74 & 0.960 & 0.179 \\
    \quad Unfitted $P^0$ & 36.69 & 0.977 & 0.181 & 37.63 & 0.978 & 0.159 & 36.27 & 0.963 & 0.174 \\
    \multicolumn{10}{@{}l}{\textit{vs.\ full ground truth}} \\
    \quad \textbf{Fitted $P$} & \textbf{32.84} & \textbf{0.973} & \textbf{0.179} & \textbf{35.52} & \textbf{0.979} & \textbf{0.132} & \textbf{35.96} & \textbf{0.983} & \textbf{0.117} \\
    \quad Nearest-centroid & 30.39 & 0.963 & 0.209 & 31.75 & 0.967 & 0.160 & 31.31 & 0.969 & 0.151 \\
    \padB \quad Unfitted $P^0$ & 31.63 & 0.971 & 0.184 & 32.70 & 0.973 & 0.147 & 32.04 & 0.974 & 0.139 \\
    \bottomrule
  \end{tabular}
  \caption{
  Projection-operator ablation on \emph{Kitchen}, projecting radiosity alone (without NEE).
  Top: indirect radiosity against indirect ground truth (full minus single-bounce reference).
  Bottom: the complete radiosity solution against full ground truth; the fitted row equals the radiosity-only row of \cref{tab:main_ablation}.
  Only the projection operator changes. The bold method label identifies our default operator.
  }
  \label{tab:supp-projection}
  \end{table*}

\end{document}